\documentclass{IEEEtaes}

\usepackage{color,array,amsthm}
\usepackage{graphicx}
\usepackage{amsmath}   
\usepackage{amsmath,amssymb} 
\usepackage{mathtools} 
\usepackage{algorithm}
\usepackage{algorithmic}
\jvol{XX}
\jnum{XX}
\jmonth{XXXXX}
\paper{1234567}
\pubyear{2020}
\doiinfo{TAES.2020.Doi Number}

\usepackage{subcaption}

\begin{document}

\title{Flying by Inference: Active Inference World Models for Adaptive UAV Swarms}

\author{Kaleem Arshid}
\member{Member, IEEE}
\affil{Department of Engineering and Naval Architecture (DITEN), University of Genoa, 16145 Genoa,  and  Intelligent Systems Laboratory, Department of Systems Engineering and Automation, Carlos III University of Madrid, 28911 Leganés, Spain} 
\author{Ali Krayani}
\member{Member, IEEE}
\affil{Department of Engineering and Naval Architecture (DITEN), University of Genoa, 16145 Genoa, Italy} 

\author{Lucio Marcenaro}
\member{Senior Member, IEEE}
\affil{Department of Engineering and Naval Architecture (DITEN), University of Genoa, 16145 Genoa, Italy}

 \author{David Martin Gomez}
 \member{Senior Member, IEEE}
 \affil{Intelligent Systems Laboratory, Department of Systems Engineering and Automation, Carlos III University of Madrid, 28911 Leganés, Spain}
\author{Carlo Regazzoni}
\member{Senior Member, IEEE}
\affil{Department of Engineering and Naval Architecture (DITEN), University of Genoa, 16145 Genoa, Italy}

%
\receiveddate{Manuscript received XXXXX 00, 0000; revised XXXXX 00, 0000; accepted XXXXX 00, 0000.\\
A preliminary version of this work appeared in~\cite{11465094}. \\
}

\corresp{The name of the corresponding author appears after the financial information, e.g. {\itshape (Corresponding author: M. Smith)}. Here you may also indicate if authors contributed equally or if there are co-first authors.}

\corresp{(Corresponding author: Kaleem Arshid, e-mail: \href{mailto:kaleem.arshid@edu.unige.it}{kaleem.arshid@edu.unige.it}).}

\authoraddress{Kaleem Arshid is with the Department of Engineering and Naval Architecture (DITEN), University of Genoa, 16145 Genoa, Italy, and the Intelligent Systems Laboratory, Department of Systems Engineering and Automation, Carlos III University of Madrid, 28911 Leganés, Spain (e-mail: \href{mailto:kaleem.arshid@edu.unige.it}{kaleem.arshid@edu.unige.it and 100526589@alumnos.uc3m.es}).
Ali Krayani is with the Department of Engineering and Naval Architecture (DITEN), University of Genoa, 16145 Genoa, Italy (e-mail: \href{mailto:ali.krayan@ieee.org}{ali.krayan@ieee.org}).
Lucio Marcenaro is with the Department of Engineering and Naval Architecture (DITEN), University of Genoa, 16145 Genoa, Italy (e-mail: \href{mailto:lucio.marcenaro@unige.it}{lucio.marcenaro@unige.it}).
David Martin Gomez is with the Intelligent Systems Laboratory, Department of Systems Engineering and Automation, Carlos III University of Madrid, 28911 Leganés, Spain (e-mail: \href{mailto:david.martin@uc3m.es}{david.martin@uc3m.es}).
Carlo Regazzoni is with the Department of Engineering and Naval Architecture (DITEN), University of Genoa, 16145 Genoa, Italy (e-mail: \href{mailto:carlo.regazzoni@unige.it}{carlo.regazzoni@unige.it}).}



\markboth{AUTHOR ET AL.}{SHORT ARTICLE TITLE}
\maketitle

%
\begin{abstract}
This paper presents an expert-guided active-inference-inspired framework for adaptive UAV swarm trajectory planning. The proposed method converts multi-UAV trajectory design from a repeated combinatorial optimization problem into a hierarchical probabilistic inference problem. In the offline phase, a genetic-algorithm planner with repulsive-force collision avoidance (GA--RF) generates expert demonstrations, which are abstracted into Mission, Route, and Motion dictionaries. These dictionaries are used to learn a probabilistic world model that captures how expert mission allocations induce route orders and how route orders induce motion-level behaviors. During online operation, the UAV swarm evaluates candidate actions by forming posterior beliefs over symbolic states and minimizing KL-divergence-based abnormality indicators with respect to expert-derived reference distributions. This enables mission allocation, route insertion, motion adaptation, and collision-aware replanning without rerunning the offline optimizer. Bayesian state estimators, including EKF and PF modules, are integrated at the motion level to improve trajectory correction under uncertainty. Simulation results show that the proposed framework preserves expert-like planning structure while producing smoother and more stable behavior than modified Q-learning. Additional validation using real-flight UAV trajectory data demonstrates that the learned world model can correct symbolic predictions under noisy and non-smooth observations, supporting its applicability to adaptive UAV swarm autonomy.
\end{abstract}


\section{INTRODUCTION}
%
Unmanned aerial vehicle (UAV) swarms are attracting increasing interest as an enabling technology for distributed autonomy, scalability, and cooperative decision-making \cite{Ahmad2025}. Beyond classical surveillance and mapping, swarms are now deployed for coordinated inspection, search and rescue, transportation, and airborne communication support, where multiple vehicles must jointly satisfy stringent constraints on energy, safety, and mission time \cite{Arshid2025UAVSwarm}. The central challenge is to synthesize trajectories that (i) allocate tasks across many vehicles, (ii) avoid collisions and dynamic obstacles, and (iii) adapt online to environment changes, all while respecting global mission objectives and limited on-board computation.

%
Deterministic optimization frameworks provide performance guarantees but typically require accurate a priori models and static environments, which limits their practicality in uncertain or rapidly varying conditions \cite{liu2021car, chai2025trajectory}. Metaheuristic approaches, such as genetic algorithms and particle-swarm optimizers, alleviate modeling assumptions and can cope with high-dimensional mixed-integer formulations, but they are mainly designed for offline planning: any significant change in targets, threats, or vehicle states demands a new optimization run, leading to limited online adaptability \cite{wang2023improved, gad2022particle, manullang2023optimum}. Data-driven schemes based on deep reinforcement learning (DRL) and multi-agent reinforcement learning (MARL) \cite{shi2024deep, li2023computation} improve autonomy by learning policies from interaction data, yet they require large and carefully curated training sets, exhibit non-trivial sample complexity, and often generalize poorly to scenarios that deviate from the training distribution.

%
Recent research has therefore moved toward probabilistic generative models and cognitive inference frameworks that explicitly couple perception, prediction, and control \cite{pezzulo2024active, nozari2022incremental}. Within this line, \emph{active inference} provides a Bayesian formalism for decision-making under uncertainty, in which agents maintain a generative model of the world and select actions that minimize discrepancies between predicted and preferred observations. Unlike canonical formulations that rely on minimizing a full variational free-energy functional, the present work focuses on a set of \emph{abnormality indicators} derived from Kullback–Leibler (KL) divergences between inferred posteriors and expert-derived reference distributions. By minimizing these indicators, the UAV swarm is driven toward sequences of symbolic observations that are statistically consistent with expert behavior.

A second key aspect of this work is the explicit use of expert knowledge to shape the action space. Instead of evaluating a large set of arbitrary candidate actions, as commonly done in traditional active inference, we construct a hierarchical \emph{world model} from trajectories generated by an offline expert planner \cite{Arshid2025}. This world model encodes, at mission, route, and motion levels, how expert solutions behave in scenarios that are well understood. During online operation, new situations are compared against these known patterns via KL-based abnormality measures. The controller then focuses computation on actions that correct the mismatch between the current situation and the set of expert-consistent behaviors, which significantly reduces online planning effort while preserving adaptability to novel conditions.

%
Building on these ideas, this paper introduces an active-inference–based world-model framework for UAV swarm trajectory design. The framework integrates probabilistic reasoning with symbolic hierarchical decision-making, enabling self-consistent adaptation across mission allocation, route ordering, and local motion generation. It combines the robustness and interpretability of model-based structure with the flexibility of learning from expert demonstrations.

%
The main contributions of this study are summarized as follows:
\begin{itemize}
  \item \textbf{Hierarchical World Model:} We construct a symbolic world model of UAV swarm behavior that spans three abstraction levels (mission, route, and motion), learned from trajectories computed by an expert genetic-algorithm–based planner.
  \item \textbf{KL-Based Active Inference:} We propose a probabilistic decision mechanism that evaluates candidate actions through multiple KL-divergence–based abnormality indicators between inferred posteriors and expert reference distributions, thereby steering the swarm toward preferred symbolic observations while avoiding full free-energy optimization.
  \item \textbf{Filter-Assisted Control:} We integrate Bayesian state estimators (e.g., EKF and PF) with the symbolic decision process to achieve smooth trajectory correction, robust collision avoidance, and consistent tracking of the synthesized motion policies.
  \item \textbf{Unified Probabilistic–Symbolic Framework:} We provide a unified architecture for UAV swarm autonomy that merges expert knowledge, probabilistic inference, and hierarchical symbolic control, yielding an interpretable, adaptive, and scalable approach to trajectory design in complex, uncertain environments.
\end{itemize}

%
The remainder of the paper is organized as follows. Section~II reviews related work on multi-UAV task assignment, trajectory planning, and probabilistic decision-making. Section~III presents the system model and problem formulation. Section~IV describes the proposed active-inference–based world-model framework. Section~V reports numerical and experimental results, and Section~VII concludes the paper.

\section{RELATED WORK}
Research on multi-UAV trajectory design spans deterministic optimization, metaheuristic search, learning-based control, and more recent probabilistic inference approaches. This section briefly reviews the most relevant contributions and positions the proposed framework within this landscape.

\subsection{Deterministic and Metaheuristic Multi-UAV Planning}
Early work on UAV path planning focused on deterministic formulations, such as mixed-integer trajectory optimization and multi-traveling-salesman (mTSP) variants. For example, path planning for maximum information collection has been formulated within an mTSP framework and solved using exact and heuristic methods \cite{Ergezer2013TAES}. More recent studies have addressed cooperative formation reconfiguration and multi-UAV task assignment using advanced optimization schemes, such as hybrid offline methods for formation reconfiguration \cite{Li2021TAES} and task-assignment algorithms that explicitly account for parameter and time-sensitive uncertainties \cite{Chen2018TAES}. Dynamic mission-planning frameworks for UAV formations in contested environments have also been proposed, where mission graphs and time-varying constraints are handled through tailored optimization algorithms \cite{Zhang2022TAES, Wu2025SciRep}. Genetic algorithms and related metaheuristics remain popular for handling large-scale discrete search spaces. F or instance, Roberge \emph{et al.} developed a fast GPU-accelerated GA path planner for fixed-wing UAVs \cite{Roberge2018TAES}. These approaches can yield high-quality solutions but are typically executed offline, and their recomputation cost hampers real-time adaptability when mission specifications change.

\subsection{Learning-Based and Swarm-Oriented Methods}
Learning-based methods have been widely explored to enhance autonomy and scalability. DRL and MARL have been applied to cooperative pursuit, coverage, and surveillance scenarios, where agents learn decentralized policies through interaction with simulators or real environments \cite{shi2024deep, li2023computation, Ekechi2025DronesMARL, Chen2025CSURRLUAV}. Recent work has investigated multi-UAV cooperative strategies with attention mechanisms or graph-based representations to capture inter-vehicle dependencies and improve task allocation robustness \cite{10938048}. 
Other studies have combined classical optimal-control or trajectory-optimization methods with learning, for example by training neural networks on trajectories generated via Pontryagin’s Maximum Principle or direct collocation so that optimal guidance commands and cost-to-go information can be approximated in real time \cite{Xu2025Drones,Horn2012JGCD}.
While these techniques can handle large state spaces, they usually require substantial training data and careful reward shaping, and their interpretability and generalization to out-of-distribution missions remain open challenges.

\subsection{Probabilistic Generative Models and Active Inference}
Probabilistic generative modeling offers an alternative viewpoint, where the agent maintains an explicit model of how observations, states, and actions are statistically related. Active inference has recently been investigated for autonomous agents, including UAVs, as a unifying principle for joint state estimation and control \cite{pezzulo2024active, nozari2022incremental}. Active-inference–based schemes have been proposed for resource allocation in UAV-based communication systems \cite{Obite2023AIUAV} and for goal-directed single-UAV trajectory planning, in which policies are treated as words in a generative model and actions are selected by minimizing divergences between predicted and preferred observations \cite{Krayani2023Sensors}. These works demonstrate the potential of active inference to integrate prior knowledge and online adaptation within a single Bayesian formalism.

The framework proposed in this paper differs from the above literature in two main aspects. First, it leverages an offline expert planner to construct a hierarchical symbolic world model of multi-UAV missions, capturing mission division, route ordering, and motion patterns across many expert demonstrations. Second, instead of minimizing a monolithic free-energy functional or exploring a large unconstrained action space, the proposed controller evaluates a restricted set of actions generated by the world model using multiple KL-divergence–based abnormality indicators. This yields an interpretable and computationally efficient decision mechanism suited to real-time UAV swarm trajectory design in dynamic and uncertain environments.

%
\section{System Model and Problem Formulation}

\subsection{System Model and Mission Objective}
We consider a swarm of $Q$ unmanned aerial vehicles (UAVs), $U=\{u_1,\dots ,u_Q\}$, deployed to cooperatively visit a set of $N$ target locations $C=\{c_1,\dots ,c_N\}$ within a bounded area. Each UAV $u_q$ starts and ends at a common depot $L_0=[x_0,y_0,z_0]$ and is described by its position $\mathbf{x}_q(t)=[x_q,y_q,z_q]$, velocity $v_q(t)$, and heading change $\Delta\phi_q(t)$. For notational convenience in the routing formulation, the depot is modeled as a virtual node indexed by $0$, while the target towns are indexed by $\{1,\dots,N\}$. UAVs maintain altitude limits and a minimum inter-UAV distance $d_{\min}$ to ensure safe operation.

Mission planning is performed in the horizontal plane, while UAVs operate at a fixed or bounded altitude. Accordingly, inter-target distances are computed in 2D, whereas dynamic feasibility, collision avoidance, and inter-UAV separation are enforced during trajectory generation.

The swarm must ensure that each target $c_i \in C$ is visited once by exactly one UAV while minimizing overall mission cost (distance, time, or energy).

\subsection{Hierarchical Decision Structure}
The decision process is organized hierarchically:
\\

\textbf{High-level (Mission Allocation):} 
The set of target towns $\mathcal{C}$ is partitioned into $Q$ disjoint subsets $\{\mathcal{C}_1, \mathcal{C}_2, \ldots, \mathcal{C}_Q\}$, 
where $\mathcal{C}_q$ is assigned to UAV $u_q$. 
The allocation aims to balance workload among UAVs and minimize total travel cost, subject to
\begin{equation}
    \bigcup_{q=1}^{Q} \mathcal{C}_q = \mathcal{C}, \qquad 
    \mathcal{C}_i \cap \mathcal{C}_j = \emptyset, \; \forall i \neq j.
\end{equation}

\textbf{Medium-level (Route Sequencing):} 
For each UAV $u_q$, determine the visiting order of its assigned towns,
\begin{equation}
    \pi_q = [c_{q,1}, c_{q,2}, \ldots, c_{q,|\mathcal{C}_q|}],
\end{equation}
which minimizes the local travel distance or time. 
This level is analogous to solving a traveling salesman problem (TSP) for each UAV.

\textbf{Low-level (Trajectory Generation):} 
Given the visiting order $\pi_q$, UAV $u_q$ generates a dynamically feasible and collision-free trajectory $\mathbf{x}_q(t)$ connecting consecutive towns, while respecting dynamic and safety constraints (including altitude bounds and minimum inter-UAV separation $d_{\min}$). Each UAV follows a closed tour that starts and ends at 
the common depot.

\subsection{MTSP Formulation and Constraints Interpretation}
Let $d_{i,j}$ denote the Euclidean distance between two towns $c_i$ and $c_j$, including the depot node $0$, forming the distance matrix $\mathbb{D} = [d_{i,j}]_{(N+1) \times (N+1)}$. 
The optimization problem is formulated as a multi-traveling salesman problem (MTSP) with the following decision variables:

\begin{equation}
X_{i,j}^q =
\begin{cases}
1, & \text{if UAV $u_q$ travels directly from city $i$ to $j$,}\\
0, & \text{otherwise.}
\end{cases}
\end{equation}

The objective is to minimize the total travel cost of all UAVs:
\begin{subequations}\label{eq:opt}
\begin{align}
\min_{X_{i,j}^q}\quad 
& \sum_{q=1}^{Q} \sum_{i=0}^{N} \sum_{j=0}^{N} d_{i,j}\, X_{i,j}^q \label{eq:obj1} \\[0.25em]
\text{s.t.}\quad 
& \sum_{q=1}^{Q} \sum_{i=0}^{N} X_{i,j}^q = 1 
&& \forall\, j \in \{1,\dots,N\} \label{eq:visit} \\[0.25em]
& \sum_{j=1}^{N} X_{i,j}^q - \sum_{j=1}^{N} X_{j,i}^q = 0 
&& \substack{
       \forall\, q \in \{1,\dots,Q\},\\
       \forall\, i \in \{1,\dots,N\}
   } \label{eq:flow} \\[0.25em]
& \sum_{i \in S} \sum_{j \in S} X_{i,j}^q \le |S| - 1 
&& \forall\, q,\ \forall\, S \subset \mathcal{C},\ S\neq\varnothing \label{eq:subtour} \\[0.25em]
& \sum_{q=1}^{Q} \sum_{j=0}^{N} X_{i,j}^q = 1 
&& \forall\, i \in \{1,\dots,N\} \label{eq:outgoing} \\[0.25em]
& \sum_{j=1}^{N} X_{0,j}^q \le 1 
&& \forall\, q \in \{1,\dots,Q\} \label{eq:depot_out} \\[0.25em]
& \sum_{i=1}^{N} X_{i,0}^q \le 1 
&& \forall\, q \in \{1,\dots,Q\} \label{eq:depot_in} \\[0.25em]
& \sum_{j=1}^{N} X_{0,j}^q = \sum_{i=1}^{N} X_{i,0}^q 
&& \forall\, q \in \{1,\dots,Q\}. \label{eq:depot_balance}
\end{align}
\end{subequations}

Constraint \eqref{eq:visit} enforces that each target city is entered exactly once across the entire swarm, i.e., each town is assigned to exactly one UAV. Constraint \eqref{eq:outgoing} complements \eqref{eq:visit} by enforcing that each target city is also departed exactly once across the swarm, preventing inconsistent assignments where a city is entered by one UAV and exited by another. Constraint \eqref{eq:flow} imposes route continuity for each UAV by balancing its incoming and outgoing edges at every visited city. Constraint \eqref{eq:subtour} prevents disconnected subtours, ensuring that each 
UAV route forms a single connected tour rather than multiple isolated loops. Finally, constraints \eqref{eq:depot_out} and \eqref{eq:depot_in} explicitly model the assumption that UAVs start from and return to the common depot, while \eqref{eq:depot_balance} ties departure and return decisions together so that routes remain closed (a UAV that leaves the depot must also return).

The global problem therefore determines, simultaneously, 
(i)~the optimal allocation of towns among UAVs, 
(ii)~the optimal visiting order for each UAV, and 
(iii)~the feasible trajectories connecting the towns. 
Because of its combinatorial nature and high dimensionality, 
this problem is computationally challenging to solve exactly for large $N$ and $Q$. 
To approximate the global optimum, a \emph{genetic optimization algorithm} is employed. 
Each individual (chromosome) in the population encodes a potential solution consisting of 
city allocation, route sequencing, and trajectory parameters. 
The fitness of each individual is evaluated using the global cost function in 
(\ref{eq:obj1}), with penalty terms included for constraint violations such as 
collision risk, insufficient inter-UAV separation, or unvisited targets. 
Through iterative evolution using selection, crossover, and mutation operations, 
the algorithm converges toward near-optimal coordinated plans for the UAV swarm.
Hard constraints in \eqref{eq:opt} are enforced either explicitly or through adaptive penalty terms, allowing efficient exploration of feasible and near-feasible solutions.

\begin{figure*}[t!]
    \centering
    \includegraphics[width=\textwidth]{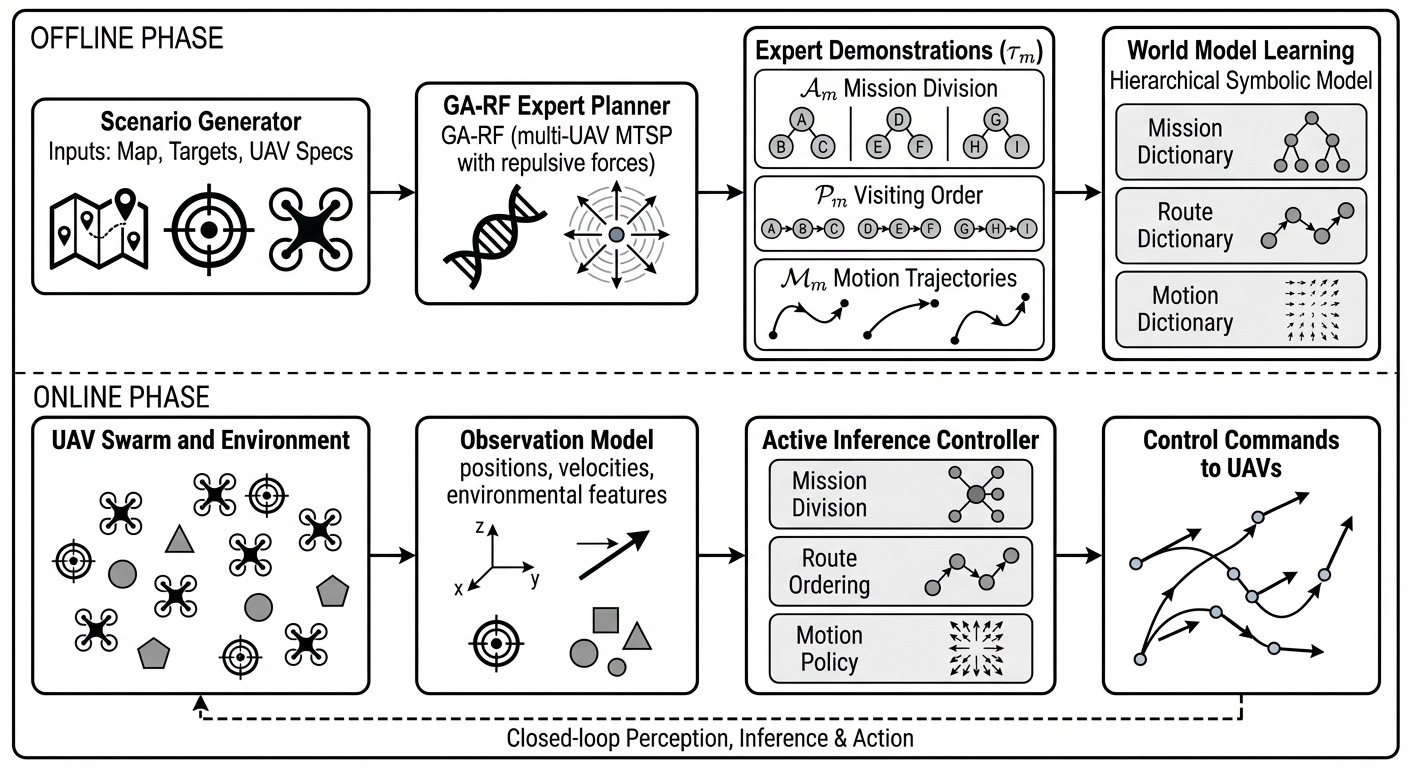}
    \caption{Workflow of the proposed active inference--driven world modeling framework for adaptive UAV swarm trajectory design.}
    \label{fig:my_AIModel}
\end{figure*}

\section{Proposed Active Inference-Based Framework}
\label{sec:proposed_framework}

This section presents the proposed active-inference-inspired framework for
multi-UAV trajectory design. The objective is to enable a UAV swarm to perform
mission allocation, route ordering, and local motion generation in an adaptive
and computationally efficient manner. Instead of repeatedly solving the full
multi-traveling salesman problem (MTSP) during online operation, the proposed
framework first extracts structured knowledge from expert demonstrations and
then uses this knowledge as a probabilistic world model for online inference.

The method is motivated by the observation that high-quality multi-UAV
solutions exhibit recurring hierarchical regularities. At the highest level,
the swarm divides the mission among UAVs. At the intermediate level, each UAV
orders the cities assigned to it. At the lowest level, continuous trajectories
are generated through local motion behaviors such as target attraction,
obstacle avoidance, and inter-UAV collision avoidance. The proposed framework
explicitly captures these regularities through three coupled symbolic
dictionaries: a Mission Dictionary, a Route Dictionary, and a Motion Dictionary.

The overall methodology consists of two phases. In the offline phase, a
genetic-algorithm planner with repulsive-force collision avoidance, denoted as
GA--RF, generates expert demonstrations for multiple mission instances. These
demonstrations are decomposed into mission-level, route-level, and motion-level
symbolic patterns. The extracted symbolic patterns are then used to learn a
hierarchical probabilistic world model that represents how expert mission
decisions induce route decisions, and how route decisions induce motion
behaviors.

In the online phase, the UAV swarm uses this learned world model to evaluate
candidate decisions under new observations. For each candidate action, the
framework forms a posterior belief over symbolic states and compares it with
the expert-derived reference distribution using a Kullback--Leibler (KL)
divergence. This divergence is interpreted as an abnormality indicator: low
abnormality indicates that the candidate action is consistent with the learned
expert behavior, while high abnormality indicates a mismatch between the
current situation and the expert-derived world model. Online decision-making
is therefore performed by selecting the candidate action that minimizes
abnormality across the mission, route, and motion levels.

The proposed framework can be interpreted as an active-inference-inspired
decision architecture. In classical active inference, agents maintain a
generative model and select actions that reduce expected disagreement between
predicted and preferred observations. In this work, the preferred observations
are represented by expert-derived symbolic reference distributions, and the
action-selection mechanism is implemented through KL-based abnormality
minimization. This provides an interpretable and computationally tractable
alternative to direct online optimization over the full MTSP solution space.

\subsection{Expert Demonstration Generation Using GA--RF}
\label{subsec:garf_demonstrations}

Let
\begin{equation}
\mathcal{E}=\{\tau_m\}_{m=1}^{M}
\label{eq:expert_dataset}
\end{equation}
denote the set of \(M\) expert demonstrations generated during the offline
learning phase. For the \(m\)-th mission instance, the input is defined as
\begin{equation}
\mathcal{I}_m =
\left\{
\mathcal{C}_m,\mathcal{U}_m,c_0,\mathcal{O}_m
\right\},
\label{eq:mission_instance_input}
\end{equation}
where
\(\mathcal{C}_m=\{c_1,\ldots,c_{N_m}\}\) is the set of target cities,
\(\mathcal{U}_m=\{u_1,\ldots,u_{Q_m}\}\) is the set of available UAVs,
\(c_0\) is the common depot, and \(\mathcal{O}_m\) denotes obstacles or
restricted regions.

The GA--RF planner produces an expert demonstration
\begin{equation}
\tau_m =
\left\{
\mathcal{C}_m,
\mathcal{U}_m,
\mathcal{A}_m,
\mathcal{P}_m,
\mathcal{M}_m
\right\},
\label{eq:expert_demonstration}
\end{equation}
where \(\mathcal{A}_m\) is the mission allocation, \(\mathcal{P}_m\) is the
set of ordered UAV routes, and \(\mathcal{M}_m\) is the set of continuous
motion trajectories.

The mission allocation is given by
\begin{equation}
\mathcal{A}_m =
\left\{
\mathcal{C}_{m,1},
\mathcal{C}_{m,2},
\ldots,
\mathcal{C}_{m,Q_m}
\right\},
\label{eq:allocation_set}
\end{equation}
where \(\mathcal{C}_{m,q}\) is the subset of cities assigned to UAV \(u_q\).
The allocation satisfies
\begin{equation}
\bigcup_{q=1}^{Q_m}\mathcal{C}_{m,q}=\mathcal{C}_m,
\qquad
\mathcal{C}_{m,i}\cap\mathcal{C}_{m,j}=\emptyset,\quad i\neq j.
\label{eq:allocation_constraints}
\end{equation}
Therefore, every target city is assigned to one and only one UAV.

For each UAV \(u_q\), the ordered route is
\begin{equation}
\pi_{m,q} =
\left[
c_0,
c_{q,1},
c_{q,2},
\ldots,
c_{q,|\mathcal{C}_{m,q}|},
c_0
\right],
\label{eq:route_definition}
\end{equation}
where the route starts and ends at the depot. The set of all UAV routes in
mission \(m\) is
\begin{equation}
\mathcal{P}_m =
\left\{
\pi_{m,1},
\pi_{m,2},
\ldots,
\pi_{m,Q_m}
\right\}.
\label{eq:route_set}
\end{equation}

The continuous trajectory of UAV \(u_q\) is denoted by
\begin{equation}
\mathbf{x}_{m,q}(t)
=
[x_{m,q}(t),y_{m,q}(t),z_{m,q}(t)]^{T},
\qquad
t\in[0,T_{m,q}],
\label{eq:continuous_trajectory}
\end{equation}
and the set of all trajectories is
\begin{equation}
\mathcal{M}_m =
\left\{
\mathbf{x}_{m,1}(t),
\ldots,
\mathbf{x}_{m,Q_m}(t)
\right\}.
\label{eq:motion_set}
\end{equation}

The GA--RF expert planner optimizes mission allocation, route ordering, and
trajectory feasibility jointly. A representative objective minimized by the
expert planner is
\begin{equation}
J_{\mathrm{GA-RF}}
=
J_{\mathrm{dist}}
+
\mu_{\mathrm{bal}}J_{\mathrm{bal}}
+
\mu_{\mathrm{obs}}J_{\mathrm{obs}}
+
\mu_{\mathrm{uav}}J_{\mathrm{uav}},
\label{eq:garf_objective}
\end{equation}
where \(J_{\mathrm{dist}}\) is the total travel cost,
\(J_{\mathrm{bal}}\) penalizes workload imbalance among UAVs,
\(J_{\mathrm{obs}}\) penalizes proximity to obstacles, and
\(J_{\mathrm{uav}}\) penalizes unsafe inter-UAV distances. The parameters
\(\mu_{\mathrm{bal}}\), \(\mu_{\mathrm{obs}}\), and \(\mu_{\mathrm{uav}}\)
control the relative importance of mission balancing, obstacle avoidance, and
inter-UAV collision avoidance.

The total route distance is
\begin{equation}
J_{\mathrm{dist}}
=
\sum_{q=1}^{Q_m}
\sum_{i=0}^{|\mathcal{C}_{m,q}|}
d(c_{q,i},c_{q,i+1}),
\label{eq:distance_cost}
\end{equation}
where \(c_{q,0}=c_0\) and
\(c_{q,|\mathcal{C}_{m,q}|+1}=c_0\). The workload-balancing term can be
defined as
\begin{equation}
J_{\mathrm{bal}}
=
\sum_{q=1}^{Q_m}
\left(
L_{m,q}-\bar{L}_{m}
\right)^2,
\qquad
\bar{L}_{m}
=
\frac{1}{Q_m}
\sum_{q=1}^{Q_m}L_{m,q},
\label{eq:balance_cost}
\end{equation}
where \(L_{m,q}\) is the length of the route assigned to UAV \(u_q\).

The inter-UAV safety penalty is activated when two UAVs approach below the
minimum separation distance \(d_{\min}\):
\begin{equation}
\small
J_{\mathrm{uav}}
=
\sum_{q=1}^{Q_m}
\sum_{\substack{r=1\\r\neq q}}^{Q_m}
\int_{0}^{T_m}
\left[
\max
\left(
0,
d_{\min}
-
\left\|
\mathbf{x}_{m,q}(t)-\mathbf{x}_{m,r}(t)
\right\|
\right)
\right]^2
dt.
\label{eq:uav_penalty}
\end{equation}
Similarly, the obstacle penalty is
\begin{equation}
J_{\mathrm{obs}}
=
\sum_{q=1}^{Q_m}
\sum_{o\in\mathcal{O}_m}
\int_{0}^{T_m}
\left[
\max
\left(
0,
d_{\min}^{\mathrm{obs}}
-
d_o(\mathbf{x}_{m,q}(t))
\right)
\right]^2
dt,
\label{eq:obstacle_penalty}
\end{equation}
where \(d_o(\mathbf{x}_{m,q}(t))\) is the distance from UAV \(u_q\) to obstacle
\(o\), and \(d_{\min}^{\mathrm{obs}}\) is the obstacle safety distance.

The expert demonstrations therefore provide structured examples of how to
solve the multi-UAV planning problem under route-efficiency, workload-balance,
and safety constraints. These demonstrations are not used to directly imitate
continuous trajectories. Instead, they are decomposed into symbolic
representations and used to learn the hierarchical world model described next.

\subsection{Hierarchical Symbolic Abstraction}
\label{subsec:symbolic_abstraction}

Each expert demonstration \(\tau_m\) is converted into a symbolic
representation
\begin{equation}
\zeta_m =
\left(
\mathbf{W}^{(\mathrm{Msn})}_m,
\mathbf{W}^{(\mathrm{Rte})}_m,
\mathbf{W}^{(\mathrm{Mot})}_m
\right),
\label{eq:symbolic_triplet}
\end{equation}
where \(\mathbf{W}^{(\mathrm{Msn})}_m\) is the mission-level symbolic pattern,
\(\mathbf{W}^{(\mathrm{Rte})}_m\) is the route-level symbolic pattern, and
\(\mathbf{W}^{(\mathrm{Mot})}_m\) is the motion-level symbolic pattern.

The symbolic abstraction follows a hierarchical language interpretation. At
the mission level, the letters correspond to target cities and the words
correspond to city subsets assigned to individual UAVs. At the route level,
the words correspond to ordered city sequences. At the motion level, the
letters correspond to local motion primitives and the words correspond to
sequences of such primitives along a route. This abstraction allows the
framework to compare, store, and reuse expert behaviors without relying on
exact geometric repetition of previous trajectories.

\subsubsection{Mission Dictionary}

At the mission level, the purpose is to encode how the global target set is
partitioned among the UAVs. For demonstration \(m\), the mission phrase is
defined as
\begin{equation}
\mathbf{W}^{(\mathrm{Msn})}_m
=
\left[
\mathcal{W}^{(\mathrm{Msn})}_{m,1},
\ldots,
\mathcal{W}^{(\mathrm{Msn})}_{m,Q_m}
\right],
\label{eq:mission_phrase}
\end{equation}
where
\begin{equation}
\mathcal{W}^{(\mathrm{Msn})}_{m,q}
=
\left\{
c \;|\; c\in\mathcal{C}_{m,q}
\right\}
\label{eq:mission_word}
\end{equation}
is the Mission Word associated with UAV \(u_q\).

The Mission Dictionary is the set of unique mission-level symbolic patterns
extracted from the expert dataset:
\begin{equation}
\mathcal{D}_{\mathrm{Msn}}
=
\left\{
w^{(\mathrm{Msn})}_1,
w^{(\mathrm{Msn})}_2,
\ldots,
w^{(\mathrm{Msn})}_{K_M}
\right\},
\label{eq:mission_dictionary}
\end{equation}
where \(K_M\) is the number of mission-level symbols. This dictionary captures
regularities in expert task allocation, such as balanced workload division,
spatial grouping of nearby cities, and allocation patterns that reduce
inter-UAV conflicts.

\subsubsection{Route Dictionary}

At the route level, each Mission Word is refined into an ordered sequence of
visits. For UAV \(u_q\), the Route Word is
\begin{equation}
\mathcal{W}^{(\mathrm{Rte})}_{m,q}
=
\left[
c_{q,1},
c_{q,2},
\ldots,
c_{q,|\mathcal{C}_{m,q}|}
\right].
\label{eq:route_word}
\end{equation}
The route phrase for the whole mission is
\begin{equation}
\mathbf{W}^{(\mathrm{Rte})}_m
=
\left[
\mathcal{W}^{(\mathrm{Rte})}_{m,1},
\ldots,
\mathcal{W}^{(\mathrm{Rte})}_{m,Q_m}
\right].
\label{eq:route_phrase}
\end{equation}

The Route Dictionary is defined as
\begin{equation}
\mathcal{D}_{\mathrm{Rte}}
=
\left\{
w^{(\mathrm{Rte})}_1,
w^{(\mathrm{Rte})}_2,
\ldots,
w^{(\mathrm{Rte})}_{K_R}
\right\},
\label{eq:route_dictionary}
\end{equation}
where \(K_R\) is the number of route-level symbols. This dictionary captures
how expert allocations are converted into feasible and efficient visiting
orders.

\subsubsection{Motion Dictionary}

At the motion level, each route segment
\((c_{q,i}\rightarrow c_{q,i+1})\) is associated with a continuous trajectory
segment
\begin{equation}
\gamma_{m,q,i}(t)
=
\mathbf{x}_{m,q}(t),
\qquad
t\in[t_i,t_{i+1}].
\label{eq:trajectory_segment}
\end{equation}

Each segment is mapped to a motion-feature vector
\begin{equation}
\boldsymbol{\phi}(\gamma_{m,q,i})
=
\left[
v,
\dot{\psi},
\kappa,
\rho,
d_{\min}^{\mathrm{obs}},
d_{\min}^{\mathrm{uav}}
\right],
\label{eq:motion_feature_vector}
\end{equation}
where \(v\) is the segment speed, \(\dot{\psi}\) is the heading rate,
\(\kappa\) is the trajectory curvature, \(\rho\) is the repulsive-energy
ratio, \(d_{\min}^{\mathrm{obs}}\) is the minimum obstacle distance along the
segment, and \(d_{\min}^{\mathrm{uav}}\) is the minimum distance to neighboring
UAVs along the segment.

The feature vectors are clustered to obtain a finite alphabet of Motion
Letters:
\begin{equation}
\mathcal{L}_{\mathrm{Mot}}
=
\left\{
\ell_1,\ell_2,\ldots,\ell_{K_L}
\right\},
\label{eq:motion_letters}
\end{equation}
where \(K_L\) is the number of motion-letter clusters. A Motion Word is then a
sequence of Motion Letters observed along a route:
\begin{equation}
\mathcal{W}^{(\mathrm{Mot})}_{m,q}
=
\left[
\ell_{m,q,1},
\ell_{m,q,2},
\ldots,
\ell_{m,q,H_q}
\right],
\label{eq:motion_word}
\end{equation}
where \(H_q\) is the number of motion segments for UAV \(u_q\).

The Motion Dictionary is
\begin{equation}
\mathcal{D}_{\mathrm{Mot}}
=
\left\{
w^{(\mathrm{Mot})}_1,
w^{(\mathrm{Mot})}_2,
\ldots,
w^{(\mathrm{Mot})}_{K_O}
\right\},
\label{eq:motion_dictionary}
\end{equation}
where \(K_O\) is the number of motion-level symbols.

The motion features are physically grounded through attractive and repulsive
potential-field quantities. The attractive potential driving UAV \(u_q\)
toward target position \(\mathbf{p}\) is
\begin{equation}
U_{\mathrm{att}}(\mathbf{x};\mathbf{p})
=
\frac{1}{2}
k_{\mathrm{att}}
\left\|
\mathbf{x}-\mathbf{p}
\right\|^2,
\label{eq:att_potential}
\end{equation}
where \(k_{\mathrm{att}}>0\) is the attractive gain.

The repulsive potential is
\begin{multline}
U_{\mathrm{rep}}(\mathbf{x})
=
\sum_{o\in\mathcal{O}}
\frac{1}{2}
k_{\mathrm{rep}}^{(o)}
\left[
\max
\left(
0,
\frac{1}{d_o(\mathbf{x})}
-
\frac{1}{d_0}
\right)
\right]^2
\\
+
\sum_{\substack{r=1\\r\neq q}}^{Q}
\frac{1}{2}
k_{\mathrm{rep}}^{(\mathrm{uav})}
\left[
\max
\left(
0,
\frac{1}{d_r(\mathbf{x})}
-
\frac{1}{d_0}
\right)
\right]^2,
\label{eq:rep_potential}
\end{multline}
where \(d_o(\mathbf{x})\) is the distance to obstacle \(o\), \(d_r(\mathbf{x})\)
is the distance to UAV \(u_r\), \(d_0\) is the influence distance beyond which
the repulsive term vanishes, \(k_{\mathrm{rep}}^{(o)}\) is the
obstacle-repulsion gain, and \(k_{\mathrm{rep}}^{(\mathrm{uav})}\) is the
inter-UAV repulsion gain.

The nominal continuous motion induced by the potential field is
\begin{equation}
\dot{\mathbf{x}}
=
-K
\nabla
\left[
U_{\mathrm{att}}(\mathbf{x})
+
U_{\mathrm{rep}}(\mathbf{x})
\right],
\label{eq:gradient_motion}
\end{equation}
where \(K\) is a positive control gain.

The repulsive contribution along a segment is quantified by
\begin{equation}
\rho
=
\frac{
\int_{\gamma}
U_{\mathrm{rep}}(\mathbf{x}(t))dt
}{
\int_{\gamma}
\left[
U_{\mathrm{att}}(\mathbf{x}(t))
+
U_{\mathrm{rep}}(\mathbf{x}(t))
\right]dt
}.
\label{eq:repulsive_ratio}
\end{equation}
Segments with small \(\rho\) correspond mainly to attractive, target-seeking
motion, whereas segments with large \(\rho\) correspond to avoidance-dominated
motion. In this way, the Motion Dictionary encodes both geometric trajectory
structure and the physical cause of local motion behavior.

\subsection{Probabilistic World Model Learning}
\label{subsec:probabilistic_world_model}

The symbolic dictionaries define the support of the world model, but online
inference requires probabilistic relationships between symbols. Therefore, the
proposed framework estimates reference distributions and transition matrices
from the expert demonstrations.

The mission-level reference distribution is estimated as
\begin{equation}
p_{\mathrm{ref}}^{(\mathrm{Msn})}
\left(
w_i^{(\mathrm{Msn})}
\right)
=
\frac{
n_i^{(\mathrm{Msn})}+\alpha
}{
\sum_{r=1}^{K_M}
n_r^{(\mathrm{Msn})}
+
\alpha K_M
},
\label{eq:mission_reference}
\end{equation}
where \(n_i^{(\mathrm{Msn})}\) is the number of times mission symbol
\(w_i^{(\mathrm{Msn})}\) appears in the expert dataset, and \(\alpha>0\) is a
smoothing constant that prevents zero-probability symbols.

The transition probability from mission symbols to route symbols is
\begin{equation}
T_{\mathrm{Msn}\rightarrow\mathrm{Rte}}(i,j)
=
p
\left(
w_j^{(\mathrm{Rte})}
\mid
w_i^{(\mathrm{Msn})}
\right),
\label{eq:mission_to_route_def}
\end{equation}
and is estimated by
\begin{equation}
T_{\mathrm{Msn}\rightarrow\mathrm{Rte}}(i,j)
=
\frac{
n_{ij}^{(\mathrm{Msn},\mathrm{Rte})}+\alpha
}{
\sum_{r=1}^{K_R}
n_{ir}^{(\mathrm{Msn},\mathrm{Rte})}
+
\alpha K_R
},
\label{eq:mission_to_route}
\end{equation}
where \(n_{ij}^{(\mathrm{Msn},\mathrm{Rte})}\) counts how often route symbol
\(w_j^{(\mathrm{Rte})}\) is observed together with mission symbol
\(w_i^{(\mathrm{Msn})}\).

Similarly, the transition probability from route symbols to motion symbols is
\begin{equation}
T_{\mathrm{Rte}\rightarrow\mathrm{Mot}}(j,k)
=
p
\left(
w_k^{(\mathrm{Mot})}
\mid
w_j^{(\mathrm{Rte})}
\right),
\label{eq:route_to_motion_def}
\end{equation}
with empirical estimate
\begin{equation}
T_{\mathrm{Rte}\rightarrow\mathrm{Mot}}(j,k)
=
\frac{
n_{jk}^{(\mathrm{Rte},\mathrm{Mot})}+\alpha
}{
\sum_{r=1}^{K_O}
n_{jr}^{(\mathrm{Rte},\mathrm{Mot})}
+
\alpha K_O
}.
\label{eq:route_to_motion}
\end{equation}

The resulting hierarchical probabilistic world model factorizes as
\begin{multline}
p
\left(
\mathbf{W}^{(\mathrm{Msn})},
\mathbf{W}^{(\mathrm{Rte})},
\mathbf{W}^{(\mathrm{Mot})}
\mid
\mathcal{E}
\right)
=
p_{\mathrm{ref}}^{(\mathrm{Msn})}
\left(
\mathbf{W}^{(\mathrm{Msn})}
\right)
\\
\times
p
\left(
\mathbf{W}^{(\mathrm{Rte})}
\mid
\mathbf{W}^{(\mathrm{Msn})}
\right)
p
\left(
\mathbf{W}^{(\mathrm{Mot})}
\mid
\mathbf{W}^{(\mathrm{Rte})}
\right).
\label{eq:hierarchical_factorization}
\end{multline}

This factorization is the probabilistic core of the proposed method. It states
that expert swarm behavior is generated hierarchically: a mission allocation
pattern is selected first, then a route-ordering pattern is selected
conditioned on the mission allocation, and finally a motion behavior is
selected conditioned on the route. Hence, the learned world model does not
only store expert trajectories; it encodes conditional dependencies that
connect high-level mission decisions to low-level motion execution.

For compact notation, the reference distribution at level
\(\ell\in\{\mathrm{Msn},\mathrm{Rte},\mathrm{Mot}\}\) is denoted as
\begin{equation}
p_{\mathrm{ref}}^{(\ell)}
=
p
\left(
\mathbf{W}^{(\ell)}
\mid
\mathrm{Pa}(\mathbf{W}^{(\ell)})
\right),
\label{eq:reference_general}
\end{equation}
where \(\mathrm{Pa}(\mathbf{W}^{(\ell)})\) denotes the parent symbolic variable
in the hierarchy. For the mission level, the parent is the mission context; for
the route level, the parent is the selected mission word; and for the motion
level, the parent is the selected route word.

\subsection{Online Belief Updating and Abnormality Minimization}
\label{subsec:belief_abnormality}

During online operation, the UAV swarm observes a possibly changing
environment. The online observation at time \(t\) is
\begin{equation}
o_t =
\left\{
\mathcal{C}_t,
\mathbf{X}_t,
\mathbf{Z}_t
\right\},
\label{eq:online_observation}
\end{equation}
where \(\mathcal{C}_t\) is the set of currently observed target cities,
\(\mathbf{X}_t=\{\mathbf{x}_c\}_{c\in\mathcal{C}_t}\) contains their
coordinates, and \(\mathbf{Z}_t\) contains motion-level observations such as
UAV positions, velocities, and obstacle measurements.

The hidden symbolic state is
\begin{equation}
s_t =
\left\{
\mathbf{W}^{(\mathrm{Msn})}_t,
\mathbf{W}^{(\mathrm{Rte})}_t,
\mathbf{W}^{(\mathrm{Mot})}_t
\right\}.
\label{eq:hidden_symbolic_state}
\end{equation}

Given an online observation \(o_t\) and a candidate action \(a^{(\ell)}\) at
level \(\ell\), the controller forms a posterior belief over symbolic states:
\begin{equation}
q_t
\left(
\mathbf{W}^{(\ell)}
\mid
o_t,
a^{(\ell)}
\right)
=
\eta_\ell
p
\left(
o_t
\mid
\mathbf{W}^{(\ell)},a^{(\ell)}
\right)
p_{\mathrm{ref}}^{(\ell)}
\left(
\mathbf{W}^{(\ell)}
\right),
\label{eq:belief_update}
\end{equation}
where \(\eta_\ell\) is a normalization constant and
\(p(o_t|\mathbf{W}^{(\ell)},a^{(\ell)})\) is the likelihood of the observation
under the symbolic state induced by the candidate action.

The likelihood is computed from a level-specific mismatch cost:
\begin{equation}
p
\left(
o_t
\mid
\mathbf{W}^{(\ell)},a^{(\ell)}
\right)
=
\frac{
\exp
\left[
-\beta_\ell
J_\ell
\left(
o_t,
\mathbf{W}^{(\ell)},
a^{(\ell)}
\right)
\right]
}{
\sum_{\bar{w}\in\mathcal{D}_{\ell}}
\exp
\left[
-\beta_\ell
J_\ell
\left(
o_t,
\bar{w},
a^{(\ell)}
\right)
\right]
},
\label{eq:likelihood_model}
\end{equation}
where \(\mathcal{D}_{\ell}\) is the dictionary at level \(\ell\),
\(J_\ell(\cdot)\) is the mismatch cost, and \(\beta_\ell>0\) controls the
sharpness of the likelihood. A larger \(\beta_\ell\) makes the posterior more
sensitive to differences in candidate costs, while a smaller value produces a
smoother belief distribution.

The abnormality associated with action \(a^{(\ell)}\) is defined as
\begin{equation}
\Delta_{\ell}
\left(
a^{(\ell)}
\right)
=
D_{\mathrm{KL}}
\left(
q_t
\left(
\mathbf{W}^{(\ell)}
\mid
o_t,
a^{(\ell)}
\right)
\middle\|
p_{\mathrm{ref}}^{(\ell)}
\right),
\label{eq:abnormality}
\end{equation}
or equivalently,
\begin{equation}
\Delta_{\ell}
\left(
a^{(\ell)}
\right)
=
\sum_{w\in\mathcal{D}_{\ell}}
q_t(w|o_t,a^{(\ell)})
\log
\frac{
q_t(w|o_t,a^{(\ell)})
}{
p_{\mathrm{ref}}^{(\ell)}(w)
}.
\label{eq:kl_expanded}
\end{equation}

A small value of \(\Delta_{\ell}\) means that
the candidate action leads to a posterior belief consistent with expert
behavior. A large value means that the candidate action produces a symbolic
state that deviates from the world model.

The total abnormality of a hierarchical decision is
\begin{equation}
\Delta_{\mathrm{total}}
=
\lambda_{\mathrm{Msn}}
\Delta_{\mathrm{Msn}}
+
\lambda_{\mathrm{Rte}}
\Delta_{\mathrm{Rte}}
+
\lambda_{\mathrm{Mot}}
\Delta_{\mathrm{Mot}},
\label{eq:total_abnormality}
\end{equation}
where \(\lambda_{\mathrm{Msn}}\), \(\lambda_{\mathrm{Rte}}\), and
\(\lambda_{\mathrm{Mot}}\) are non-negative weights controlling the relative
importance of the three abstraction levels.

The selected hierarchical action is
\begin{equation}
a_t^{\star}
=
\arg\min_{a_t}
\Delta_{\mathrm{total}}(a_t),
\quad
a_t
=
\left\{
a^{(\mathrm{Msn})},
a^{(\mathrm{Rte})},
a^{(\mathrm{Mot})}
\right\}.
\label{eq:hierarchical_action}
\end{equation}

\subsection{Mission-Level Decision}
\label{subsec:mission_decision}

At the mission level, the controller decides how the observed cities should be
assigned to the available UAVs. A mission-level action is a partitioning
function
\begin{equation}
a^{(\mathrm{Msn})}
:
\mathcal{C}_t
\rightarrow
\left\{
\mathcal{C}_{t,1},
\ldots,
\mathcal{C}_{t,Q}
\right\},
\label{eq:mission_action}
\end{equation}
where \(\mathcal{C}_{t,q}\) is the subset of cities assigned to UAV \(u_q\) at
time \(t\).

The feasible mission-action set is
\begin{equation}
\mathcal{A}_{t}^{(\mathrm{Msn})}
=
\left\{
a^{(\mathrm{Msn})}
:
\bigcup_{q=1}^{Q}\mathcal{C}_{t,q}=\mathcal{C}_t,
\;
\mathcal{C}_{t,i}\cap\mathcal{C}_{t,j}=\emptyset
\right\}.
\label{eq:mission_action_set}
\end{equation}

For each candidate allocation, the mission-level mismatch cost is
\begin{equation}
J_{\mathrm{Msn}}
=
\omega_d J_{\mathrm{dist}}
+
\omega_b J_{\mathrm{bal}}
+
\omega_s J_{\mathrm{safe}},
\label{eq:mission_cost}
\end{equation}
where \(J_{\mathrm{dist}}\) penalizes long assignment distances,
\(J_{\mathrm{bal}}\) penalizes unbalanced workloads, and \(J_{\mathrm{safe}}\)
penalizes allocations likely to create unsafe inter-UAV interactions.

The selected mission-level action is
\begin{equation}
a^{(\mathrm{Msn})\star}
=
\arg\min_{a^{(\mathrm{Msn})}\in\mathcal{A}_{t}^{(\mathrm{Msn})}}
\Delta_{\mathrm{Msn}}
\left(
a^{(\mathrm{Msn})}
\right).
\label{eq:mission_selection}
\end{equation}

If a new city \(c^\star\) appears during execution, the system generates
candidate assignment actions
\begin{equation}
a_q^{(\mathrm{add})}(c^\star),
\qquad q=1,\ldots,Q,
\end{equation}
where \(a_q^{(\mathrm{add})}(c^\star)\) denotes assigning the new city to UAV
\(u_q\). The selected UAV is
\begin{equation}
q^\star
=
\arg\min_{q\in\{1,\ldots,Q\}}
\Delta_{\mathrm{Msn}}
\left(
a_q^{(\mathrm{add})}(c^\star)
\right).
\label{eq:new_city_assignment}
\end{equation}

Therefore, when a new target appears, the framework does not rerun the full
GA--RF optimization. Instead, it evaluates how adding the new target to each
UAV changes the mission-level belief and selects the assignment that produces
the smallest abnormality.

\subsection{Route-Level Decision}
\label{subsec:route_decision}

After mission allocation, the route level determines the visiting order for
the cities assigned to each UAV. For UAV \(u_q\), a route-level action defines
a permutation
\begin{equation}
a_q^{(\mathrm{Rte})}
:
\mathcal{C}_{t,q}
\rightarrow
\pi_{t,q}
=
\left[
c_{q,1},
c_{q,2},
\ldots,
c_{q,|\mathcal{C}_{t,q}|}
\right].
\label{eq:route_action}
\end{equation}

The feasible route-action set is
\begin{equation}
\mathcal{A}_{t}^{(\mathrm{Rte})}
=
\left\{
a^{(\mathrm{Rte})}
:
a^{(\mathrm{Rte})}
\text{ is feasible conditioned on }
a^{(\mathrm{Msn})\star}
\right\}.
\label{eq:route_action_set}
\end{equation}

The route-level mismatch cost is
\begin{equation}
J_{\mathrm{Rte}}
=
\omega_L L(\pi_{t,q})
+
\omega_T J_{\mathrm{turn}}
+
\omega_C J_{\mathrm{cross}},
\label{eq:route_cost}
\end{equation}
where \(L(\pi_{t,q})\) is the route length, \(J_{\mathrm{turn}}\) penalizes
unnecessary sharp turns, and \(J_{\mathrm{cross}}\) penalizes route structures
that increase the probability of path crossings or inter-UAV conflicts.

The selected route-level action is
\begin{equation}
a^{(\mathrm{Rte})\star}
=
\arg\min_{a^{(\mathrm{Rte})}\in\mathcal{A}_{t}^{(\mathrm{Rte})}}
\Delta_{\mathrm{Rte}}
\left(
a^{(\mathrm{Rte})}
\right).
\label{eq:route_selection}
\end{equation}

If a new city \(c^\star\) has already been assigned to UAV \(u_{q^\star}\),
the route-level controller evaluates all possible insertion positions:
\begin{equation}
a_j^{(\mathrm{ins})}(c^\star),
\qquad
j=1,\ldots,|\mathcal{C}_{t,q^\star}|+1.
\end{equation}
The selected insertion index is
\begin{equation}
j^\star
=
\arg\min_{j}
\Delta_{\mathrm{Rte}}
\left(
a_j^{(\mathrm{ins})}(c^\star)
\right).
\label{eq:route_insertion}
\end{equation}

Thus, environmental changes are handled hierarchically: the mission level
decides which UAV should serve the new target, while the route level decides
where that target should be inserted in the selected UAV's route.

\subsection{Motion-Level Decision}
\label{subsec:motion_decision}

At the motion level, the selected route is translated into continuous motion
commands. A motion-level action corresponds to a Motion Word or a short
sequence of Motion Letters selected from the Motion Dictionary:
\begin{equation}
a^{(\mathrm{Mot})}
\in
\mathcal{A}_{t}^{(\mathrm{Mot})}
\subseteq
\mathcal{D}_{\mathrm{Mot}}.
\label{eq:motion_action}
\end{equation}

The motion-level mismatch cost is
\begin{equation}
J_{\mathrm{Mot}}
=
\omega_x
\left\|
\mathbf{x}_{q,t+1}^{\mathrm{pred}}
-
\mathbf{x}_{q,t+1}^{\mathrm{ref}}
\right\|^2
+
\omega_o J_{\mathrm{obs}}
+
\omega_u J_{\mathrm{uav}},
\label{eq:motion_cost}
\end{equation}
where \(\mathbf{x}_{q,t+1}^{\mathrm{pred}}\) is the predicted next position of
UAV \(u_q\), \(\mathbf{x}_{q,t+1}^{\mathrm{ref}}\) is the reference position
implied by the selected motion word, \(J_{\mathrm{obs}}\) penalizes obstacle
proximity, and \(J_{\mathrm{uav}}\) penalizes violation of the minimum
inter-UAV distance.

The selected motion-level action is
\begin{equation}
a^{(\mathrm{Mot})\star}
=
\arg\min_{a^{(\mathrm{Mot})}\in\mathcal{A}_{t}^{(\mathrm{Mot})}}
\Delta_{\mathrm{Mot}}
\left(
a^{(\mathrm{Mot})}
\right).
\label{eq:motion_selection}
\end{equation}

When the predicted distance to an obstacle or another UAV becomes small, the
likelihood of an attractive-only motion word decreases. Consequently, the
posterior belief shifts toward motion words containing repulsive Motion
Letters. This increases the abnormality of the previous motion behavior and
causes the controller to switch toward avoidance-oriented behavior. Therefore,
collision avoidance is treated as a belief-correction mechanism within the
world model rather than as an isolated reactive rule.

\subsection{Filter-Assisted Motion-Level Belief Correction}
\label{subsec:filter_assisted_correction}

The symbolic motion-level decision determines the desired behavioral mode, but
continuous execution requires state estimation under noisy observations.
Therefore, Bayesian filters are integrated into the motion-level loop.

The continuous state of UAV \(u_q\) is defined as
\begin{equation}
\mathbf{s}_{q,t}
=
\left[
\mathbf{x}_{q,t}^{T},
\mathbf{v}_{q,t}^{T}
\right]^T,
\label{eq:continuous_state}
\end{equation}
where \(\mathbf{x}_{q,t}\) and \(\mathbf{v}_{q,t}\) are the position and
velocity of UAV \(u_q\). The state evolves according to
\begin{equation}
\mathbf{s}_{q,t+1}
=
f
\left(
\mathbf{s}_{q,t},
\mathbf{u}_{q,t}
\right)
+
\mathbf{w}_{q,t},
\label{eq:state_transition}
\end{equation}
where \(\mathbf{u}_{q,t}\) is the control input associated with the selected
motion word, and \(\mathbf{w}_{q,t}\) is process noise. The observation model
is
\begin{equation}
\mathbf{y}_{q,t}
=
h
\left(
\mathbf{s}_{q,t}
\right)
+
\mathbf{v}_{q,t}^{\mathrm{obs}},
\label{eq:observation_model}
\end{equation}
where \(\mathbf{y}_{q,t}\) is the sensor measurement and
\(\mathbf{v}_{q,t}^{\mathrm{obs}}\) is observation noise.

For computationally efficient state correction, the Extended Kalman Filter
(EKF) prediction step is
\begin{equation}
\hat{\mathbf{s}}_{q,t+1|t}
=
f
\left(
\hat{\mathbf{s}}_{q,t|t},
\mathbf{u}_{q,t}
\right),
\label{eq:ekf_prediction_state}
\end{equation}
\begin{equation}
\mathbf{P}_{q,t+1|t}
=
\mathbf{F}_{q,t}
\mathbf{P}_{q,t|t}
\mathbf{F}_{q,t}^{T}
+
\mathbf{Q}_{q,t},
\label{eq:ekf_prediction_cov}
\end{equation}
where \(\mathbf{F}_{q,t}\) is the Jacobian of the transition model,
\(\mathbf{P}_{q,t|t}\) is the state covariance, and \(\mathbf{Q}_{q,t}\) is
the process-noise covariance.

The EKF correction step is
\begin{equation}
\mathbf{K}_{q,t+1}
=
\mathbf{P}_{q,t+1|t}
\mathbf{H}_{q,t+1}^{T}
\left(
\mathbf{H}_{q,t+1}
\mathbf{P}_{q,t+1|t}
\mathbf{H}_{q,t+1}^{T}
+
\mathbf{R}_{q,t+1}
\right)^{-1},
\label{eq:ekf_gain}
\end{equation}
\begin{equation}
\hat{\mathbf{s}}_{q,t+1|t+1}
=
\hat{\mathbf{s}}_{q,t+1|t}
+
\mathbf{K}_{q,t+1}
\left[
\mathbf{y}_{q,t+1}
-
h
\left(
\hat{\mathbf{s}}_{q,t+1|t}
\right)
\right],
\label{eq:ekf_update_state}
\end{equation}
\begin{equation}
\mathbf{P}_{q,t+1|t+1}
=
\left(
\mathbf{I}
-
\mathbf{K}_{q,t+1}
\mathbf{H}_{q,t+1}
\right)
\mathbf{P}_{q,t+1|t}.
\label{eq:ekf_update_cov}
\end{equation}

When the dynamics are strongly nonlinear or the uncertainty is non-Gaussian, a
Particle Filter (PF) can be used. The PF approximates the belief over the
continuous state as
\begin{equation}
p(\mathbf{s}_{q,t}|\mathbf{y}_{1:t})
\approx
\sum_{i=1}^{N_p}
w_{q,t}^{(i)}
\delta
\left(
\mathbf{s}_{q,t}
-
\mathbf{s}_{q,t}^{(i)}
\right),
\label{eq:pf_representation}
\end{equation}
where \(N_p\) is the number of particles,
\(\mathbf{s}_{q,t}^{(i)}\) is the \(i\)-th particle, and \(w_{q,t}^{(i)}\) is
its normalized weight.

The predicted continuous state is used to evaluate the motion-level likelihood
and safety terms. If
\begin{equation}
\left\|
\hat{\mathbf{x}}_{q,t+1|t}
-
\hat{\mathbf{x}}_{r,t+1|t}
\right\|
<
d_{\min},
\qquad r\neq q,
\label{eq:predicted_collision}
\end{equation}
then the mismatch cost of the current motion word increases, which reduces its
posterior probability and increases its abnormality. This causes the
controller to select a safer motion word, typically one with stronger
repulsive behavior.

Thus, the filter and the symbolic world model are coupled. The filter predicts
and corrects the continuous UAV state, while the symbolic inference layer
selects the motion behavior that best explains and corrects the predicted
state. This coupling connects high-level probabilistic decision-making with
executable UAV trajectories.

\subsection{Online Hierarchical Inference Algorithm}
\label{subsec:online_algorithm}

The complete online decision procedure is summarized in
Algorithm~\ref{alg:online_hierarchical_abnormality}. At each time step, the UAV
swarm observes the environment, evaluates mission-level candidates,
route-level candidates, and motion-level candidates, and selects the
hierarchical action that minimizes the total abnormality.

\begin{algorithm}[t]
\caption{Online Hierarchical Abnormality-Minimization}
\label{alg:online_hierarchical_abnormality}
\begin{algorithmic}[1]
\REQUIRE Learned dictionaries
\(\mathcal{D}_{\mathrm{Msn}}\),
\(\mathcal{D}_{\mathrm{Rte}}\),
\(\mathcal{D}_{\mathrm{Mot}}\);
transition matrices
\(T_{\mathrm{Msn}\rightarrow\mathrm{Rte}}\),
\(T_{\mathrm{Rte}\rightarrow\mathrm{Mot}}\);
online observation \(o_t\)
\ENSURE Selected hierarchical action
\(a_t^\star=
\{a^{(\mathrm{Msn})\star},
a^{(\mathrm{Rte})\star},
a^{(\mathrm{Mot})\star}\}\)

\STATE Observe current targets, UAV states, and obstacles:
\(o_t=\{\mathcal{C}_t,\mathbf{X}_t,\mathbf{Z}_t\}\)

\STATE Generate feasible mission-level candidate allocations
\(\mathcal{A}^{(\mathrm{Msn})}_t\)

\FOR{each \(a^{(\mathrm{Msn})}\in\mathcal{A}^{(\mathrm{Msn})}_t\)}
    \STATE Compute mission-level mismatch \(J_{\mathrm{Msn}}\)
    \STATE Compute posterior belief
    \(q_t(\mathbf{W}^{(\mathrm{Msn})}|o_t,a^{(\mathrm{Msn})})\)
    \STATE Compute abnormality
    \(\Delta_{\mathrm{Msn}}(a^{(\mathrm{Msn})})\)
\ENDFOR

\STATE Select
\(a^{(\mathrm{Msn})\star}
=
\arg\min
\Delta_{\mathrm{Msn}}\)

\STATE Generate feasible route-level candidates conditioned on
\(a^{(\mathrm{Msn})\star}\)

\FOR{each \(a^{(\mathrm{Rte})}\in\mathcal{A}^{(\mathrm{Rte})}_t\)}
    \STATE Compute route-level mismatch \(J_{\mathrm{Rte}}\)
    \STATE Compute posterior belief
    \(q_t(\mathbf{W}^{(\mathrm{Rte})}|o_t,a^{(\mathrm{Rte})})\)
    \STATE Compute abnormality
    \(\Delta_{\mathrm{Rte}}(a^{(\mathrm{Rte})})\)
\ENDFOR

\STATE Select
\(a^{(\mathrm{Rte})\star}
=
\arg\min
\Delta_{\mathrm{Rte}}\)

\STATE Generate feasible motion-level candidates conditioned on
\(a^{(\mathrm{Rte})\star}\)

\FOR{each \(a^{(\mathrm{Mot})}\in\mathcal{A}^{(\mathrm{Mot})}_t\)}
    \STATE Predict UAV continuous state using EKF or PF
    \STATE Compute motion-level mismatch \(J_{\mathrm{Mot}}\)
    \STATE Compute posterior belief
    \(q_t(\mathbf{W}^{(\mathrm{Mot})}|o_t,a^{(\mathrm{Mot})})\)
    \STATE Compute abnormality
    \(\Delta_{\mathrm{Mot}}(a^{(\mathrm{Mot})})\)
\ENDFOR

\STATE Select
\(a^{(\mathrm{Mot})\star}
=
\arg\min
\Delta_{\mathrm{Mot}}\)

\STATE Return selected hierarchical action:
\[
a_t^\star=
\{a^{(\mathrm{Msn})\star},
a^{(\mathrm{Rte})\star},
a^{(\mathrm{Mot})\star}\}
\]

\STATE Execute the selected motion action and update observations at \(t+1\)
\end{algorithmic}
\end{algorithm}

The proposed online procedure avoids enumerating the complete MTSP solution
space. Instead, it evaluates a restricted set of candidate symbolic actions at
each abstraction level. The online computational effort is therefore governed
by the number of candidates evaluated at the three levels:
\begin{equation}
\mathcal{O}_{\mathrm{online}}
=
\mathcal{O}
\left(
|\mathcal{A}_{t}^{(\mathrm{Msn})}|
+
|\mathcal{A}_{t}^{(\mathrm{Rte})}|
+
|\mathcal{A}_{t}^{(\mathrm{Mot})}|
\right),
\label{eq:online_complexity}
\end{equation}
rather than by the full combinatorial number of possible multi-UAV tours. This
is the main mechanism that enables online replanning when new targets,
obstacles, or collision risks appear.

\section{Results and Analysis}
\label{sec:results}

This section evaluates the proposed active-inference-inspired world-model framework for multi-UAV trajectory design. The objective of the evaluation is not only to compare trajectory quality with baseline methods, but also to verify whether the proposed hierarchical inference mechanism behaves consistently with the methodology introduced in Section~\ref{sec:proposed_framework}. In particular, the experiments are designed to assess five aspects: 
(i) whether GA--RF demonstrations can be transformed into meaningful hierarchical transition structures; 
(ii) whether online action selection follows the proposed KL-based abnormality-minimization rule; 
(iii) whether symbolic mission and route decisions can be translated into feasible and collision-aware trajectories; 
(iv) whether the framework can adapt to environmental changes without retraining or rerunning the offline optimizer; and 
(v) whether the learned world model remains effective when evaluated using real-flight trajectories affected by sensing noise and human-pilot variability.

The evaluation is organized into two complementary parts. First, the framework is assessed using fully simulated multi-UAV missions, allowing controlled analysis of the learned world model, online inference, adaptive replanning, filtering-assisted correction, and comparison with baseline methods. Second, real-flight trajectory data collected in indoor UAV experiments are incorporated into the validation pipeline to examine robustness under non-ideal motion patterns and realistic observation uncertainty.

\subsection{Experimental Setup and Evaluation Protocol}
\label{subsec:experimental_setup}

The proposed framework is evaluated in a fully online multi-UAV trajectory design setting. Simulated experiments are conducted in a bounded two-dimensional environment of size \(1000 \times 1000\)~m, with all UAVs operating at a constant altitude of 200~m. A central depot is used as both the initial and terminal location for all UAVs. Target locations, referred to as towns, are randomly distributed within the operational area, and each target must be visited exactly once by a single UAV.

During the offline phase, \(M=5000\) mission instances are generated using the GA--RF expert planner. Each instance provides a complete expert solution consisting of mission division, route ordering, and dynamically feasible motion trajectories. These expert solutions are converted into symbolic Mission, Route, and Motion Words and are used to learn the hierarchical world model, including the corresponding dictionaries and transition matrices. No information from the online testing scenarios is used during training.

During online testing, new target configurations are generated and treated as unseen mission instances. The UAV swarm relies exclusively on the learned world model and the proposed KL-based abnormality-minimization mechanism to perform mission allocation, route selection, and motion execution. The GA--RF planner is used as the expert reference, while a modified Q-learning baseline trained on the same expert data is used for comparative evaluation. The comparison therefore evaluates whether the proposed framework can preserve expert-like planning structure while avoiding repeated execution of the offline optimizer and improving stability relative to reward-driven learning. Unless otherwise stated, each quantitative curve is averaged over $N_{\mathrm{test}}=1000$ unseen mission instances.

\subsection{Evaluation Metrics}
\label{subsec:evaluation_metrics}

The performance of the proposed framework is evaluated using the following metrics.

\begin{itemize}
\item \textit{Mission Completion Time}: total time required for all UAVs to visit their assigned targets and return to the depot.

\item \textit{Total Traveled Distance}: cumulative distance traveled by all UAVs during mission execution, used as a proxy for energy consumption.

\item \textit{Minimum Inter-UAV Distance}: minimum distance observed between any pair of UAVs during execution, used to assess collision avoidance and safety.

\item \textit{Trajectory Similarity}: similarity between the trajectories or symbolic decisions generated by the proposed framework and those produced by the GA--RF expert planner. This metric is evaluated at the mission-division and route-ordering levels.

\item \textit{State Estimation Error}: root-mean-square error (RMSE) between estimated and ground-truth states when EKF and PF modules are employed.

\item \textit{Abnormality Indicator}: KL divergence between the inferred symbolic posterior distribution and the expert-derived reference distribution at the mission, route, and motion levels.
\end{itemize}

Together, these metrics evaluate both trajectory-level performance and the internal consistency of the proposed inference mechanism. Mission completion time and total distance quantify efficiency, minimum inter-UAV distance evaluates safety, trajectory similarity measures consistency with expert planning structure, RMSE quantifies filter-assisted state estimation accuracy, and the abnormality indicator verifies the proposed decision rule.

\subsection{Learning the Hierarchical World Model from Simulated Demonstrations}
\label{subsec:simulated_world_model}

\begin{figure}[ht!]
\centering
    \begin{center}
      \begin{subfigure}[t]{0.5\linewidth}
        \centering
        \includegraphics[width=5.1cm]{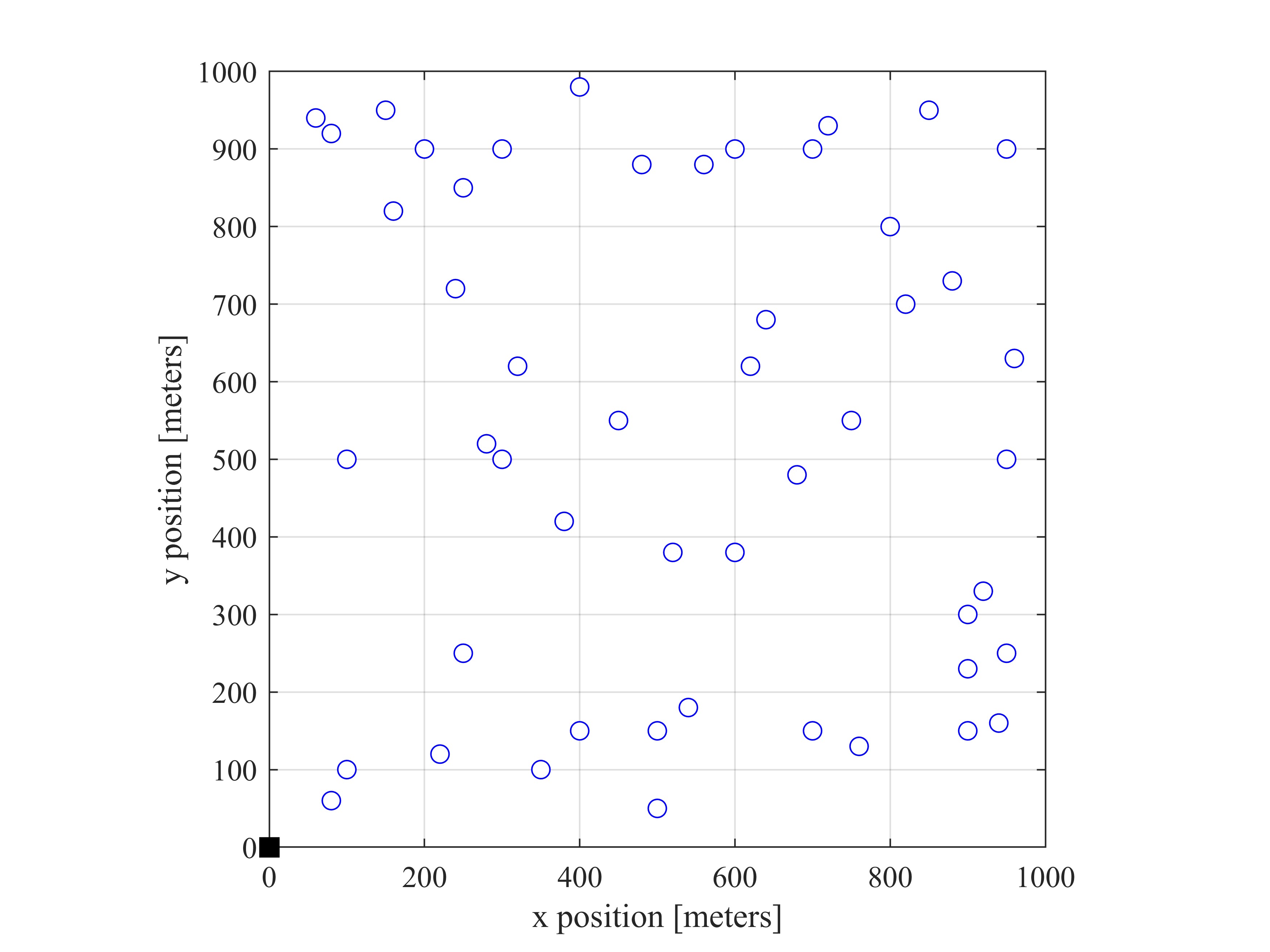}
        \subcaption{}
      \end{subfigure}\hfill
      \begin{subfigure}[t]{0.5\linewidth}
        \centering
        \includegraphics[width=5.1cm]{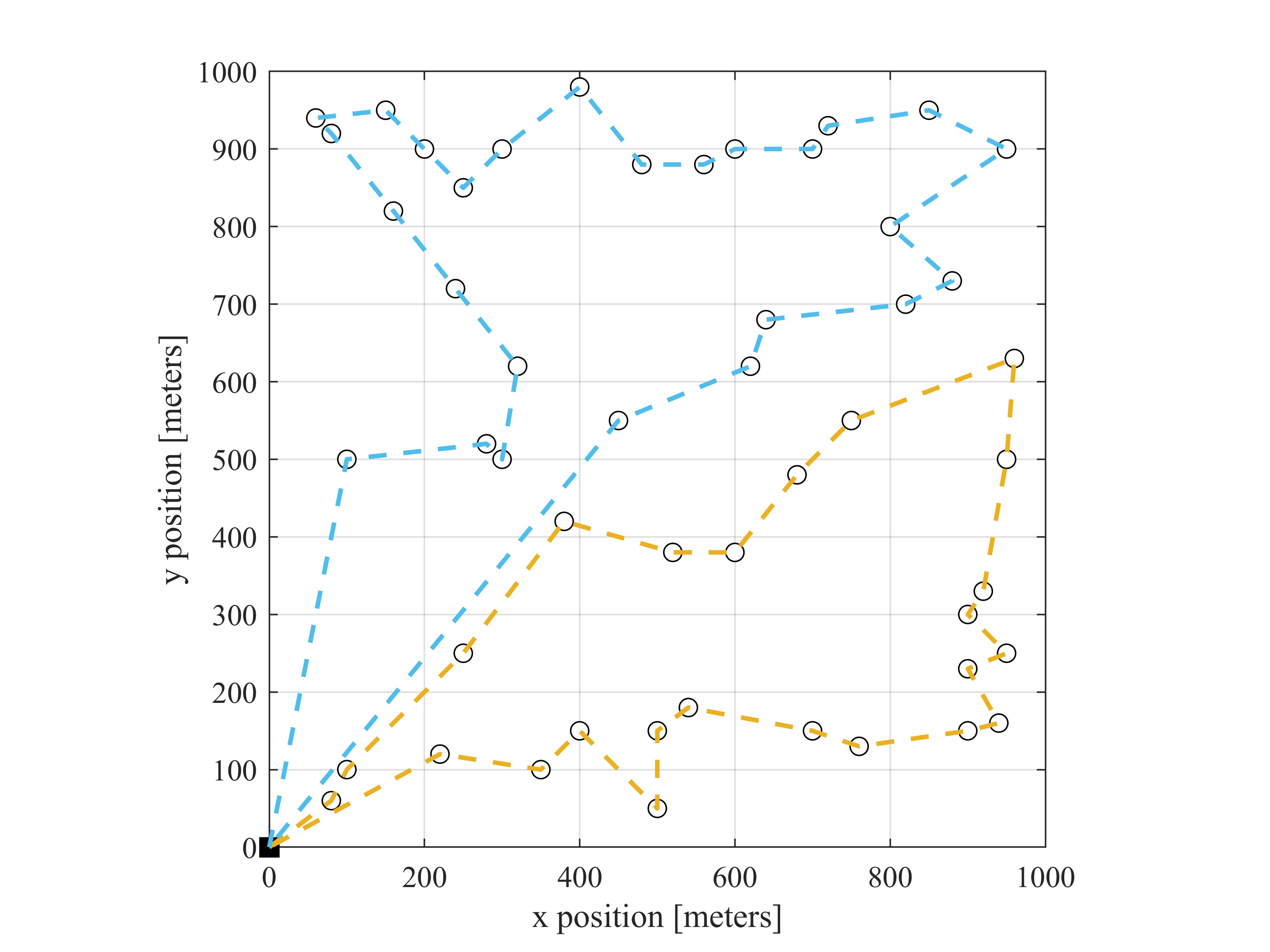}
        \subcaption{}
      \end{subfigure}
    \caption{Representative simulated mission instance. 
    (a) Random spatial distribution of 50 target locations within the operational area. 
    (b) Corresponding expert solution generated by the GA--RF optimizer, illustrating coordinated collision-free trajectories for two UAVs, where each target is visited exactly once.}
    \label{fig:2324}
    \end{center}
\end{figure}

Fig.~\ref{fig:2324} illustrates a representative simulated mission instance used to generate expert demonstrations. Fig.~\ref{fig:2324}(a) shows a random spatial distribution of 50 target locations within the operational area, together with the central depot. Fig.~\ref{fig:2324}(b) shows the corresponding GA--RF expert solution, where the targets are assigned to two UAVs and visited through coordinated collision-free trajectories.

This example illustrates the first stage of the proposed methodology. The GA--RF trajectory is not used as a continuous path to be copied directly during online operation. Instead, it is decomposed into mission-allocation, route-ordering, and motion-level symbolic descriptions. These symbolic descriptions become training samples for the hierarchical world model.

\begin{figure}[t!]
    \centering   
    \begin{minipage}{0.22\textwidth}
        \centering
        \includegraphics[width=\linewidth]{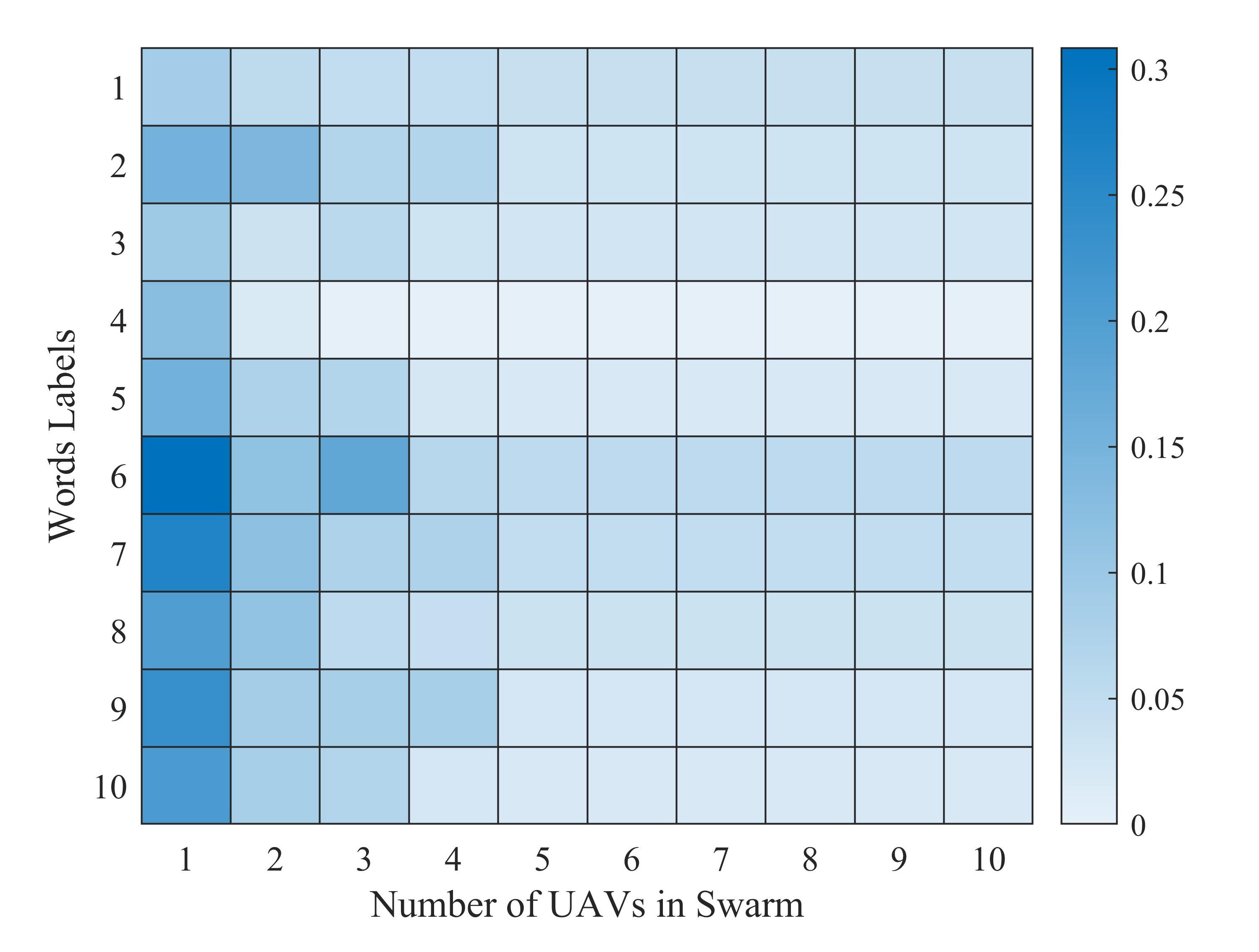}
        \subcaption{}
    \end{minipage}
     \hfill
    \begin{minipage}{0.22\textwidth}
        \centering
        \includegraphics[width=\linewidth]{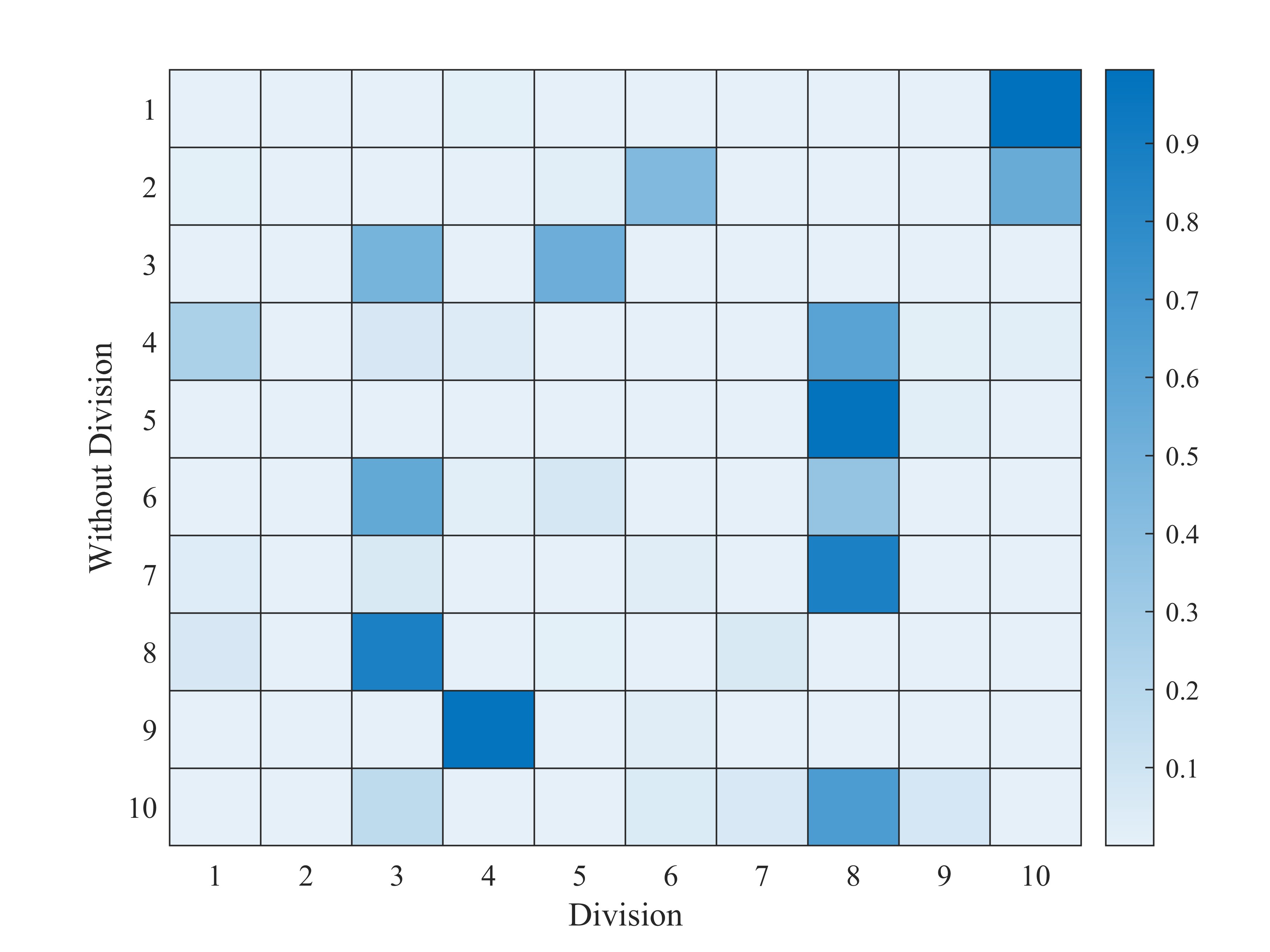}
        \subcaption{}
    \end{minipage}
    \hfill
    \begin{minipage}{0.22\textwidth}
        \centering
        \includegraphics[width=\linewidth]{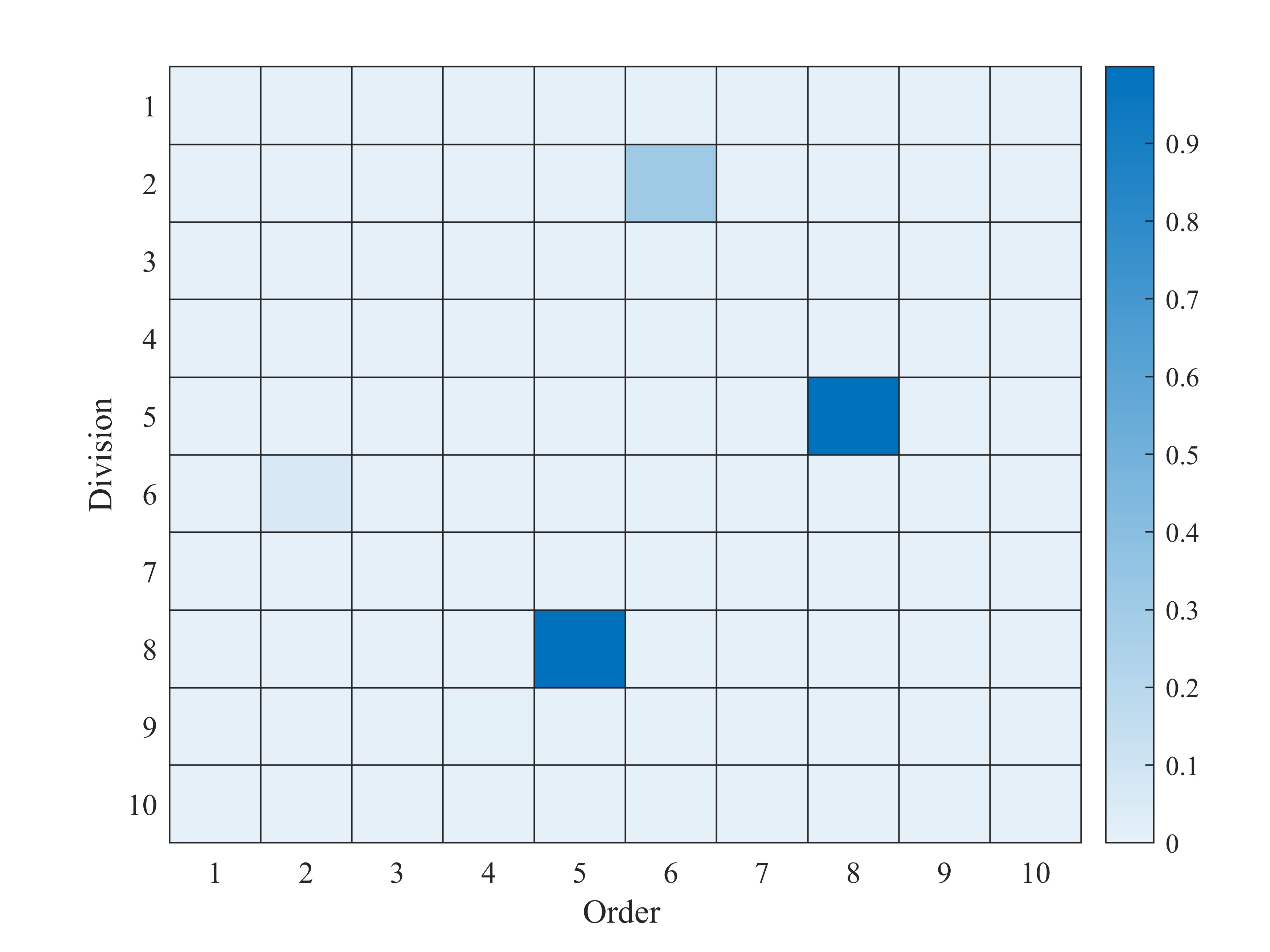}
        \subcaption{}
    \end{minipage}
    \hfill
    \begin{minipage}{0.22\textwidth}
        \centering
        \includegraphics[width=\linewidth]{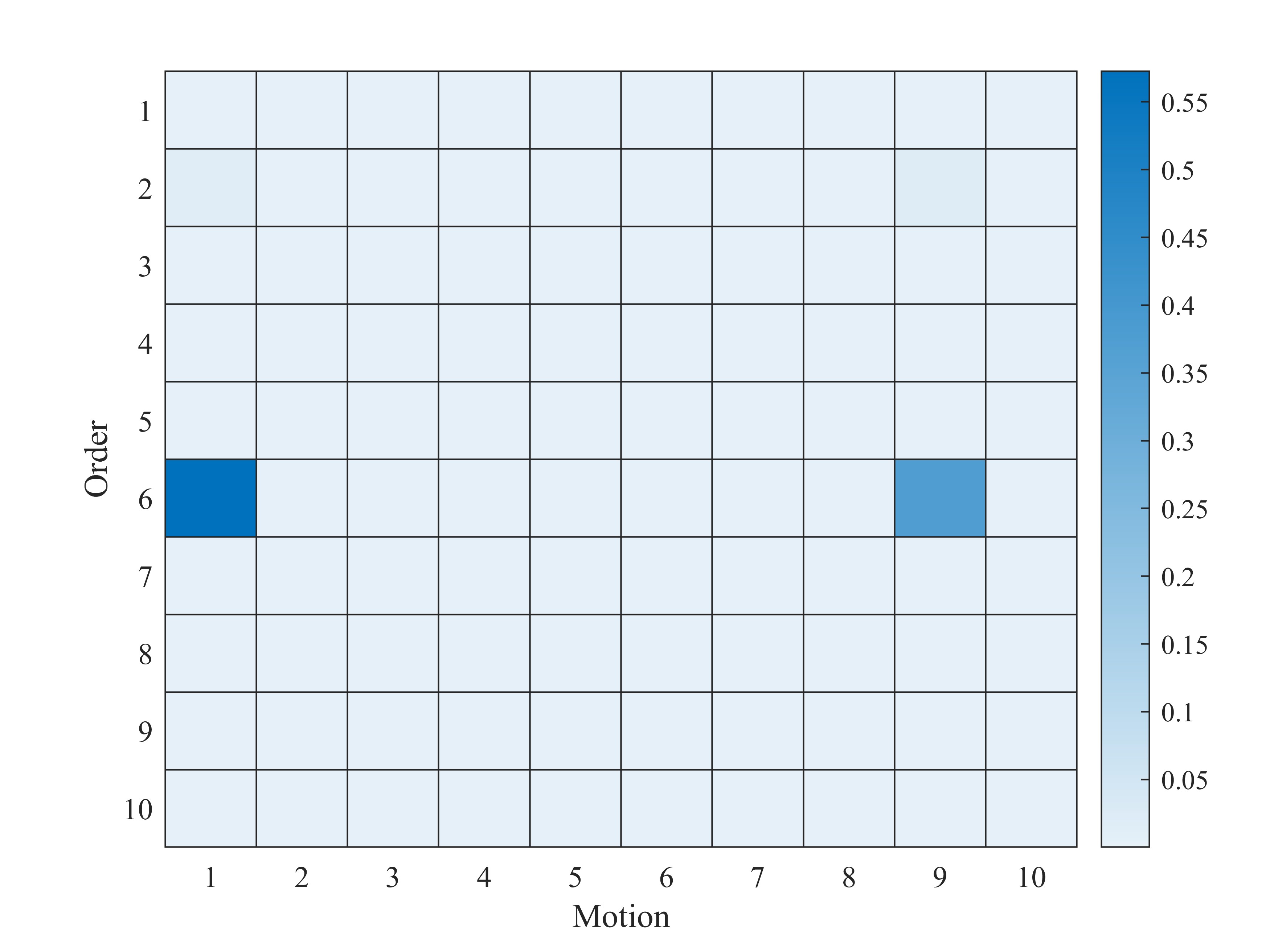}
        \subcaption{}
    \end{minipage}
    \caption{Transition matrices learned from GA--RF expert demonstrations.
    (a) Mapping between the number of targets and the inferred swarm size.
    (b) Transition probabilities from the mission context to mission-division symbols.
    (c) Transition probabilities from mission-division symbols to route-ordering symbols.
    (d) Transition probabilities from route-ordering symbols to motion-level primitives.
    These matrices constitute the empirical probabilistic structure of the learned hierarchical world model.}
  \label{fig:my_Trans}
\end{figure}

The hierarchical transition matrices learned from the expert demonstrations are shown in Fig.~\ref{fig:my_Trans}. These matrices represent the empirical probabilistic structure of the learned world model. Fig.~\ref{fig:my_Trans}(a) captures the relationship between the number of targets and the inferred swarm size. Fig.~\ref{fig:my_Trans}(b) represents the transition from the input mission context to mission-division symbols. Fig.~\ref{fig:my_Trans}(c) corresponds to the learned mission-to-route transition structure, while Fig.~\ref{fig:my_Trans}(d) corresponds to the route-to-motion transition structure.

The sparse and concentrated patterns observed in the matrices indicate that expert behavior is not uniformly distributed over all symbolic possibilities. Instead, only a subset of mission, route, and motion patterns receives high probability. This supports the main assumption of the proposed framework: online planning can be restricted to expert-consistent symbolic candidates rather than exploring the complete combinatorial MTSP solution space.

\subsection{Online Abnormality-Minimization in Unseen Scenarios}
\label{subsec:online_abnormality}

\begin{figure}[t!]
    \centering
    
    \begin{minipage}{0.22\textwidth}
        \centering
        \includegraphics[width=\linewidth]{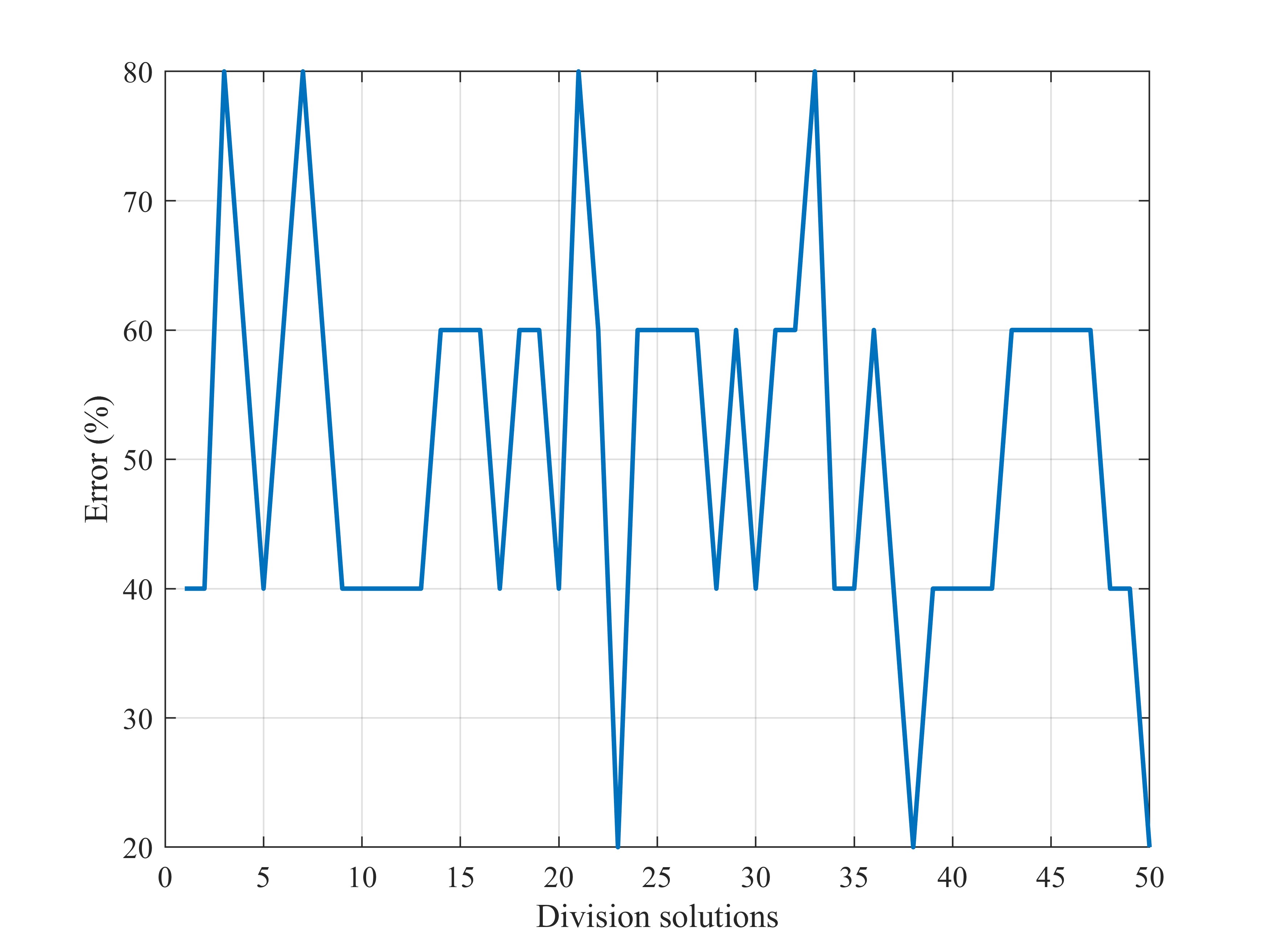}
        \subcaption{}
    \end{minipage}
    \hfill
    \begin{minipage}{0.22\textwidth}
        \centering
        \includegraphics[width=\linewidth]{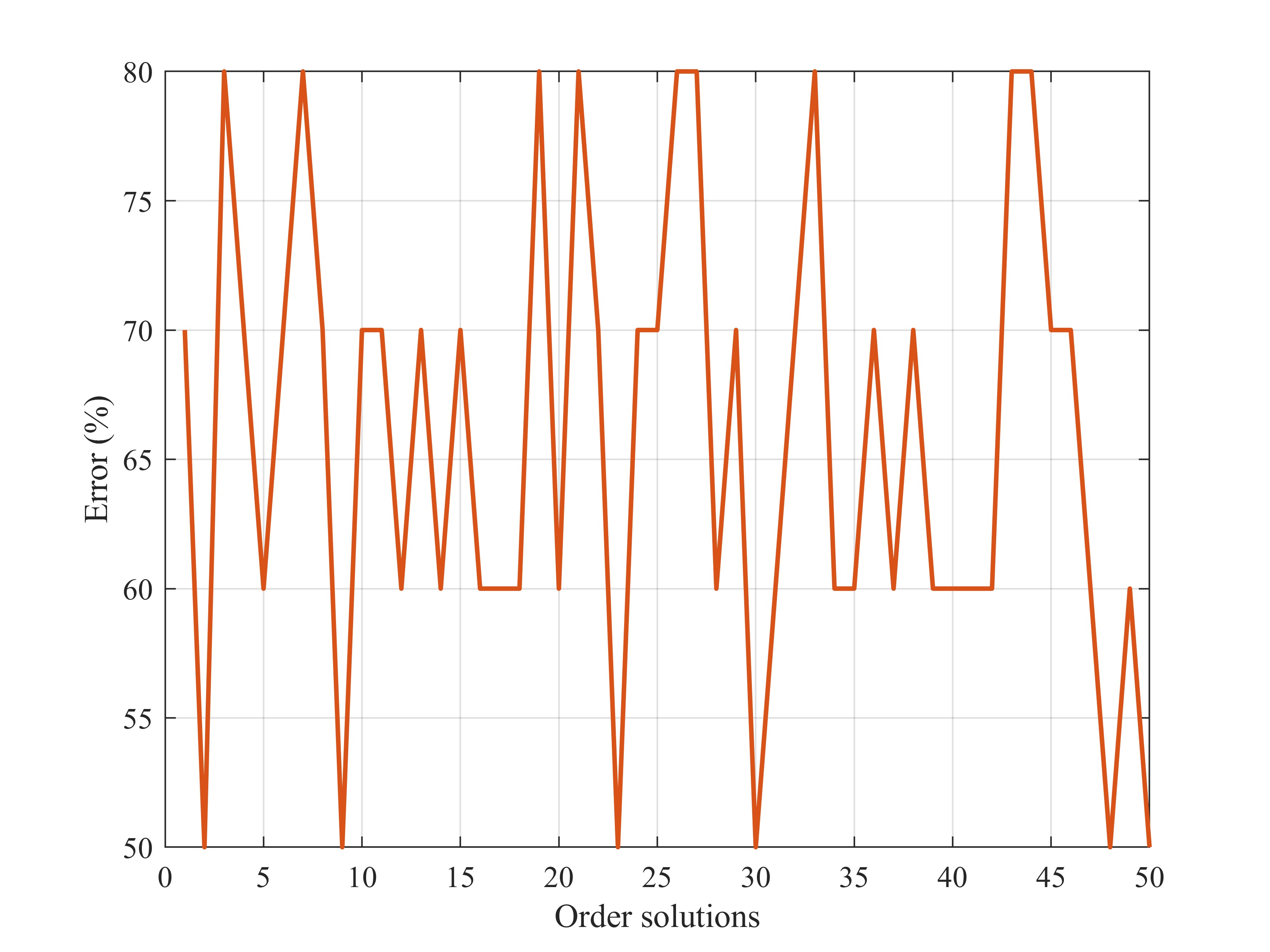}
        \subcaption{}
    \end{minipage}
    \hfill
    \begin{minipage}{0.22\textwidth}
        \centering
        \includegraphics[width=\linewidth]{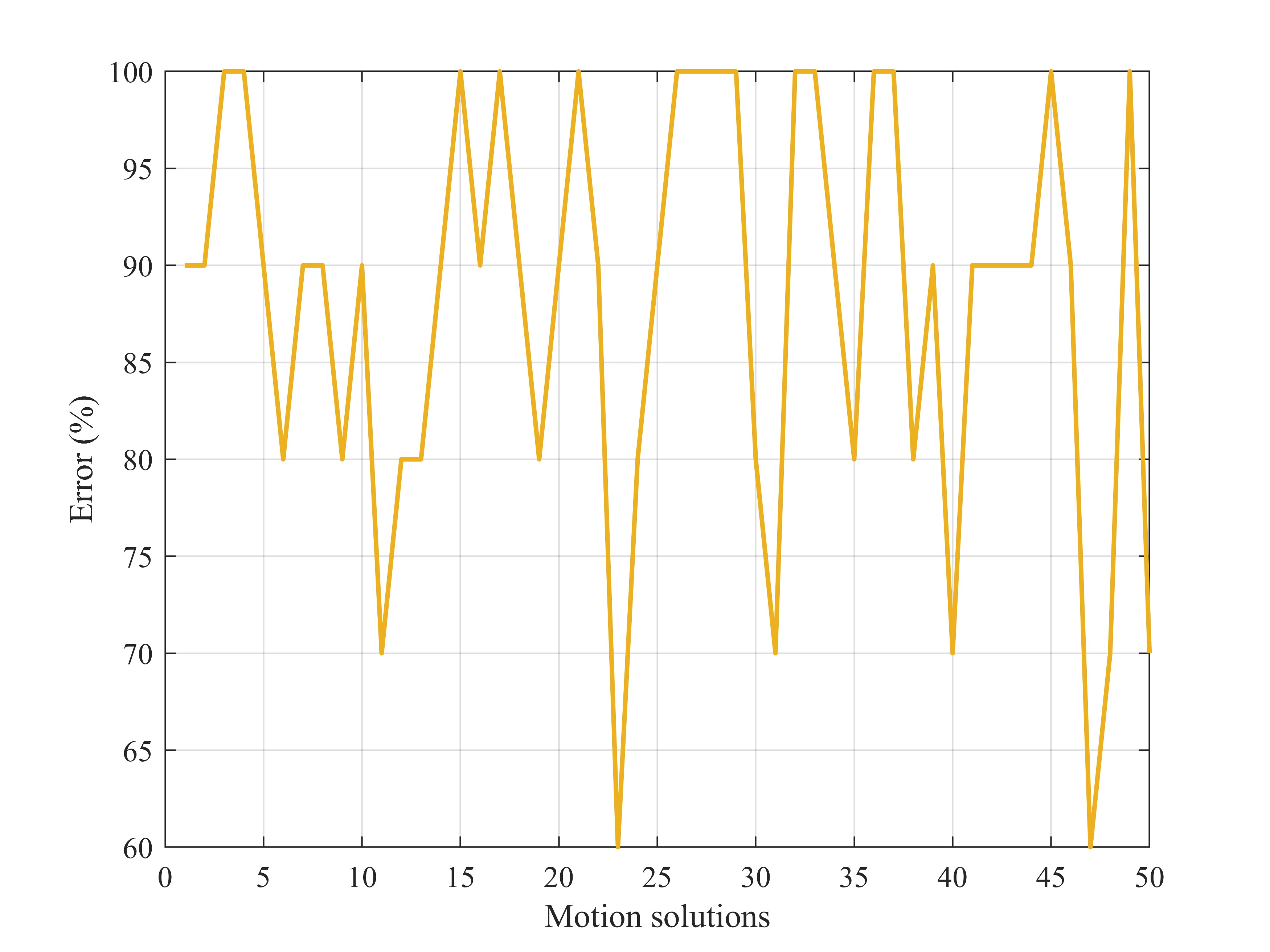}
        \subcaption{}
    \end{minipage}
\caption{KL-based abnormality indicators for candidate actions in an unseen scenario.
(a) Mission-division level.
(b) Route-ordering level.
(c) Motion-execution level.
At each level, the selected action corresponds to the minimum abnormality relative to the expert-derived reference distribution, confirming consistent hierarchical inference.}
  \label{fig:my_Trans1}
\end{figure}

During online testing, the learned world model is used to evaluate candidate decisions at the mission, route, and motion levels. Fig.~\ref{fig:my_Trans1} reports the abnormality indicators associated with candidate actions in an unseen scenario. The three panels correspond to the mission-division, route-ordering, and motion-execution levels.

At each level, the selected candidate corresponds to the minimum KL-based abnormality. This directly validates the online decision rule introduced in the methodology: the controller selects the action whose posterior symbolic belief is closest to the expert-derived reference distribution. Therefore, online decision-making is not performed by rerunning the GA--RF optimizer. Instead, the swarm restores consistency between the current observation and the learned world model through abnormality minimization.

%

\subsection{Hierarchical Trajectory Execution and Adaptive Replanning}
\label{subsec:hierarchical_execution}

\begin{figure}[ht]
\centering
\includegraphics[width=0.40\textwidth]{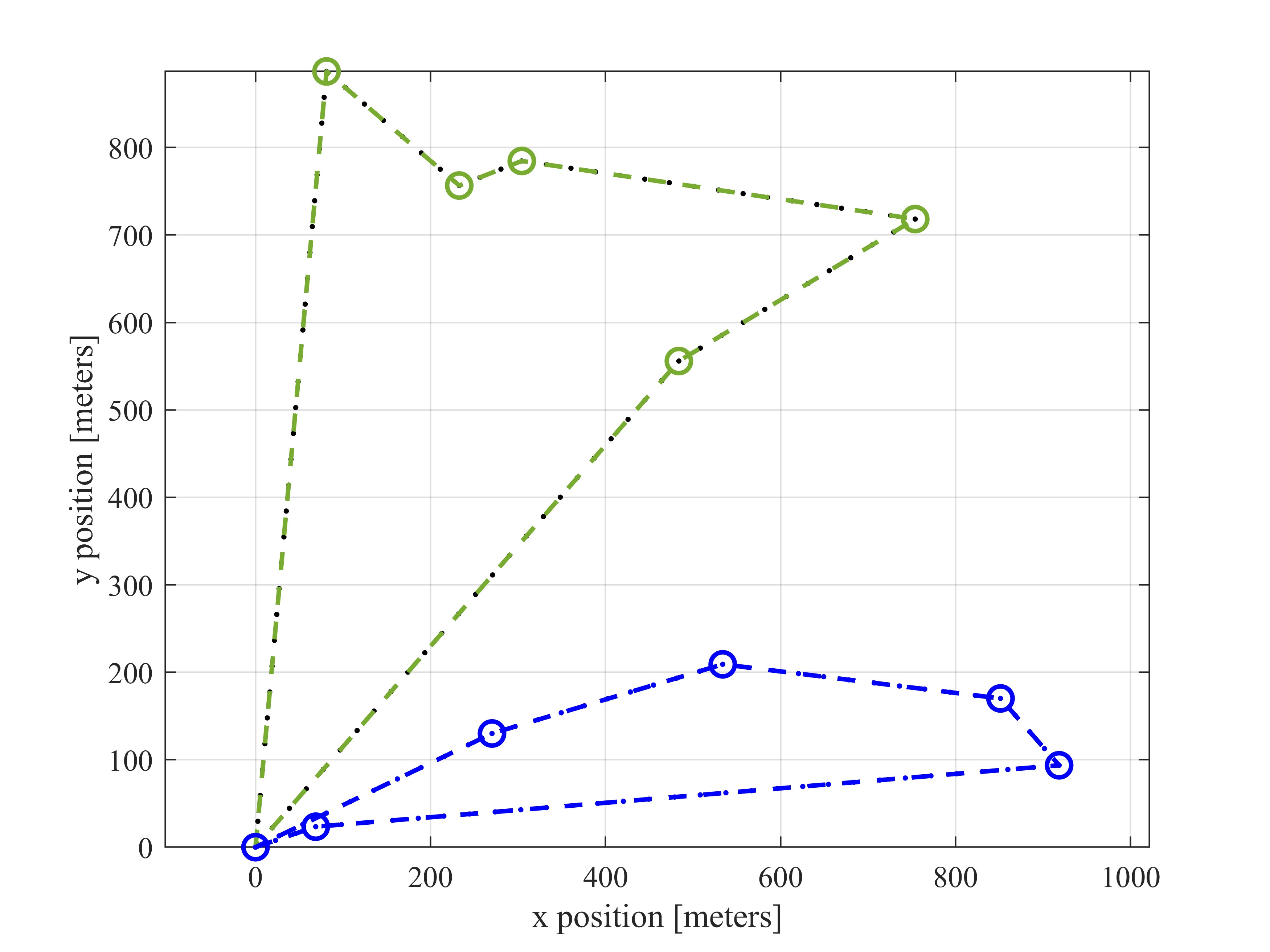} 
\caption{Hierarchical trajectory generation at the mission and route levels.
The inferred task allocation and visiting order are consistent with symbolic patterns learned from expert demonstrations, producing coordinated and redundancy-free swarm behavior.}
\label{fig:ain-levels-a} 
\end{figure}

\begin{figure}[ht]
\centering
\includegraphics[width=0.40\textwidth]{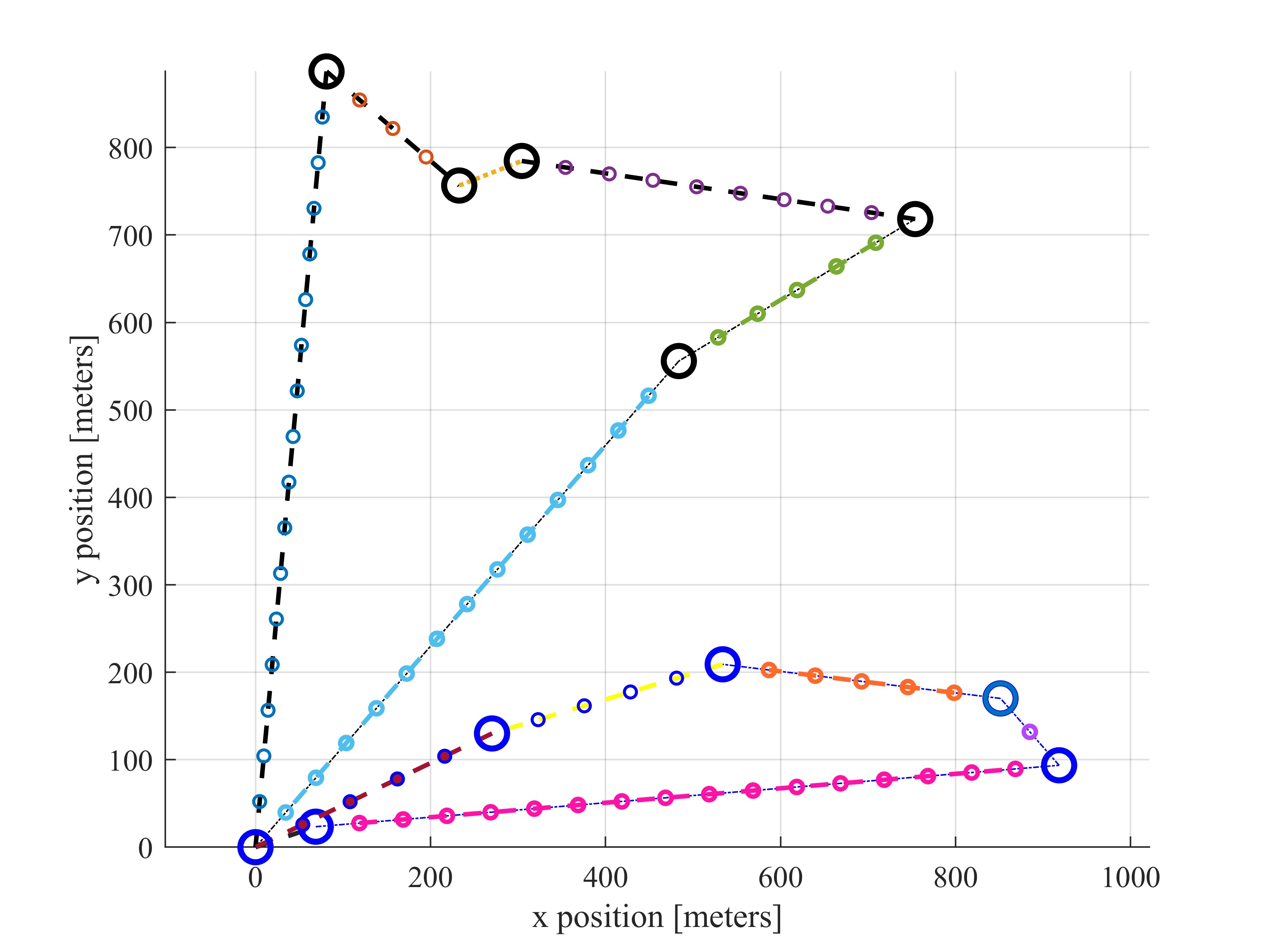} 
\caption{Motion-level execution using learned symbolic motion primitives.
The UAVs generate dynamically feasible trajectories composed of attractive and repulsive behaviors, translating high-level decisions into smooth collision-aware motion.}
\label{fig:ain-levels-b} 
\end{figure}

Figs.~\ref{fig:ain-levels-a} and \ref{fig:ain-levels-b} illustrate how the selected symbolic decisions are translated into executable trajectories. Fig.~\ref{fig:ain-levels-a} shows the mission- and route-level outputs, where the inferred allocation and visiting order generate coordinated routes for the UAV swarm. Fig.~\ref{fig:ain-levels-b} shows the corresponding motion-level execution, where the UAVs follow dynamically feasible trajectories generated from learned motion primitives.

These results demonstrate that the proposed hierarchy produces consistent decisions across abstraction levels. The mission layer assigns targets without redundancy, the route layer orders the assigned targets, and the motion layer converts symbolic motion words into continuous trajectories. The resulting behavior confirms that the learned world model is not only useful for symbolic classification, but also supports executable multi-UAV trajectory generation.


\begin{figure}[ht]
\centering
\includegraphics[width=0.40\textwidth]{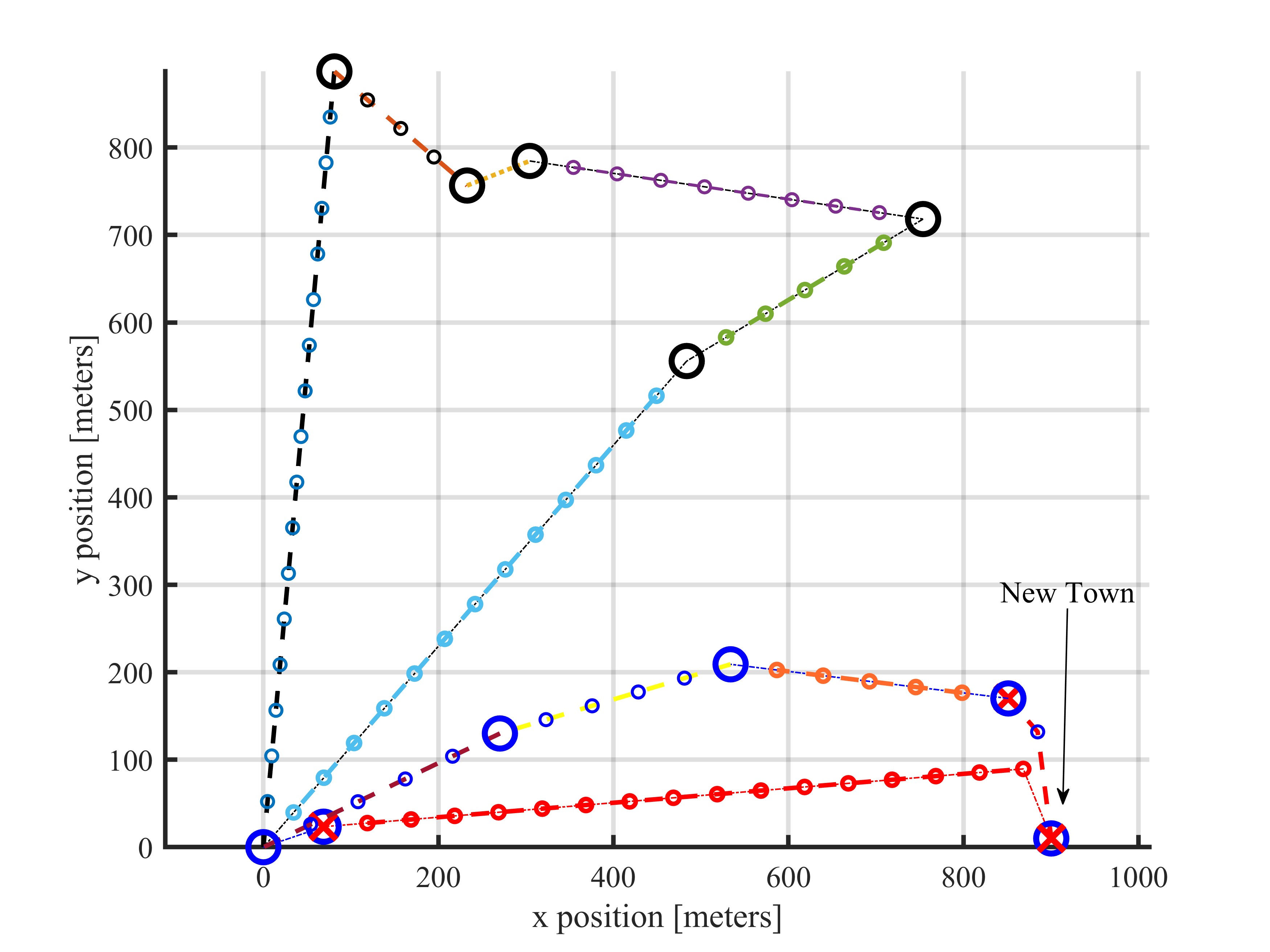} 
\caption{Detection of environmental change during online execution.
The unexpected appearance of a new target increases the symbolic abnormality and triggers belief revision within the proposed inference framework.}
\label{fig:surprise-a} 
\end{figure}

\begin{figure}[ht]
\centering
\includegraphics[width=0.40\textwidth]{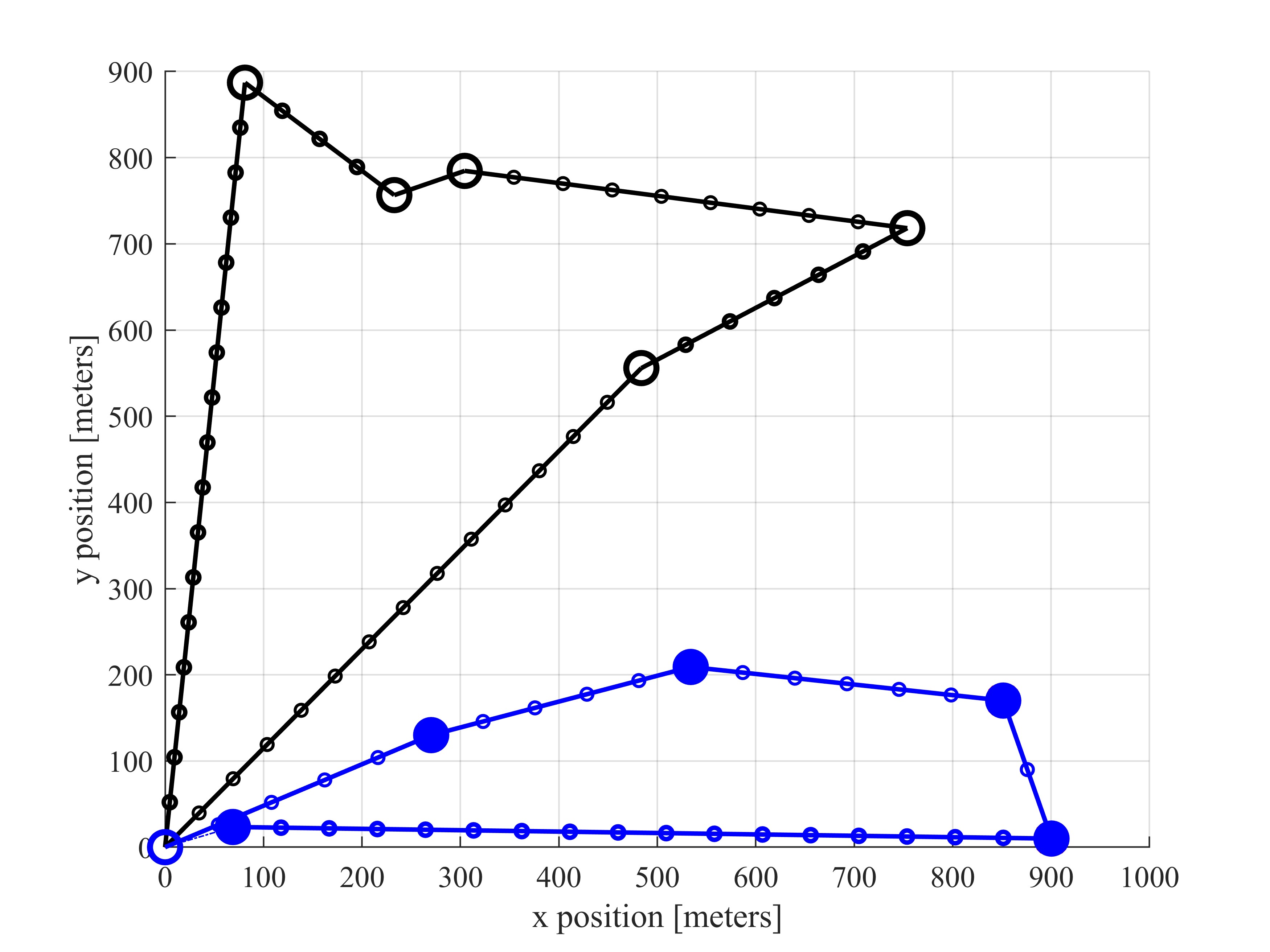} 
\caption{Adaptive trajectory correction following belief updating.
The UAV swarm replans its trajectories to restore consistency with the learned world model while preserving safety constraints.}
\label{fig:surprise-b} 
\end{figure}

Adaptive replanning under environmental changes is shown in Figs.~\ref{fig:surprise-a} and \ref{fig:surprise-b}. In Fig.~\ref{fig:surprise-a}, a new target appears during online execution. This event changes the current observation and increases the abnormality of the previously selected symbolic plan. The proposed framework then performs belief revision by evaluating candidate mission assignments and route insertions for the new target. Fig.~\ref{fig:surprise-b} shows the corrected trajectory after the belief update.

This result is central to the proposed framework. The new target is handled hierarchically: first, the mission level identifies the UAV assignment that minimizes mission abnormality; second, the route level selects the insertion position that minimizes route abnormality; and third, the motion level updates the local trajectory while preserving safety. Thus, the framework adapts to environmental changes without retraining the world model and without executing a new global GA--RF optimization.

%

\subsection{Filter-Assisted State Estimation and Collision Avoidance}
\label{subsec:filter_results}

The motion-level decision provides the selected symbolic behavior, but continuous trajectory execution requires robust state estimation under noisy observations and obstacle uncertainty. For this reason, EKF and PF modules are integrated into the motion-level loop. The filters predict the continuous UAV state, and the predicted state is then used to evaluate motion-level likelihoods and safety costs.

\begin{figure}[ht!]
  \centering
  \begin{subfigure}[t]{0.50\linewidth}
    \centering
    \includegraphics[width=\linewidth]{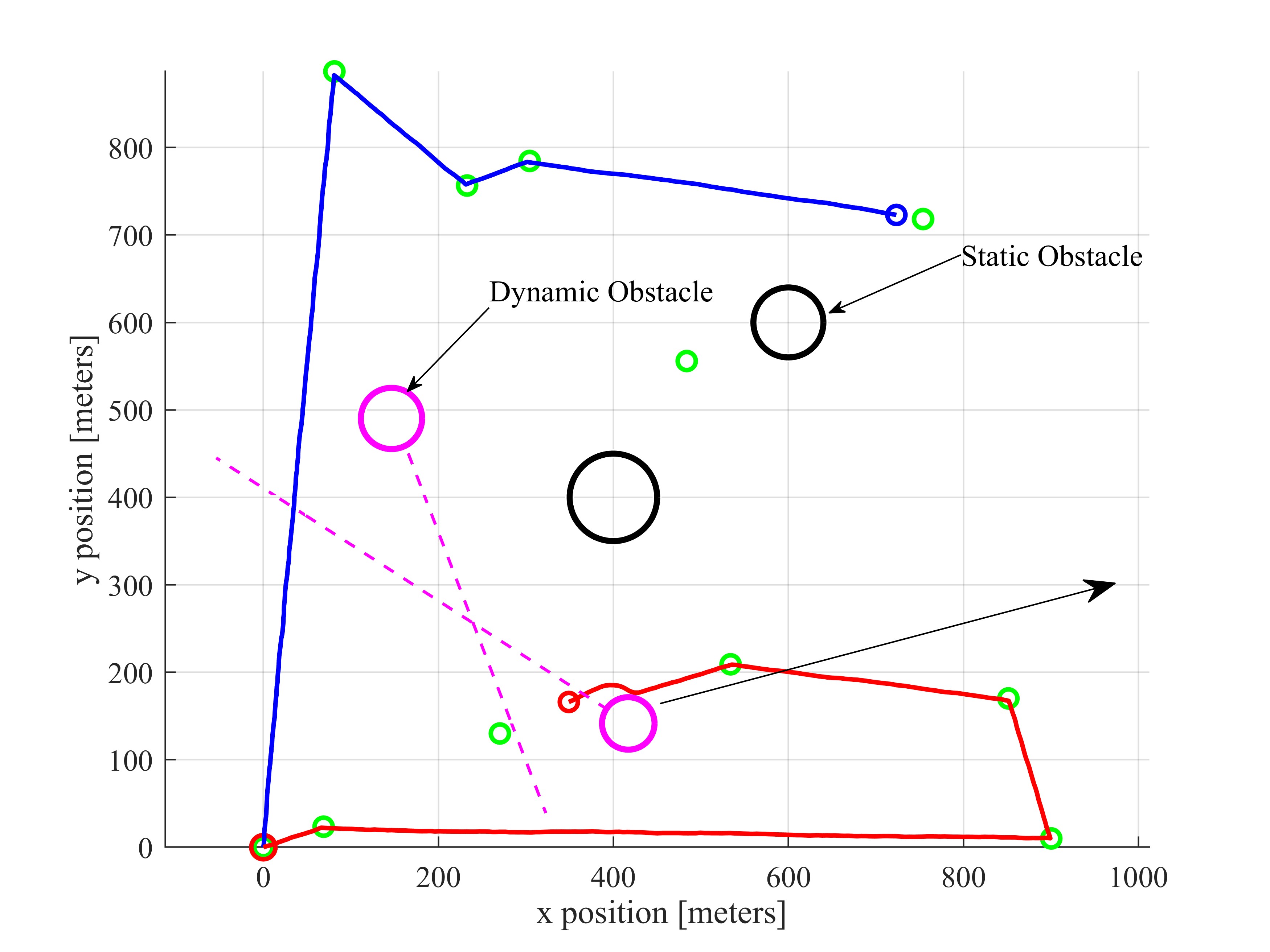}
    \subcaption{}
  \end{subfigure}\hfill
  \begin{subfigure}[t]{0.50\linewidth}
    \centering
    \includegraphics[width=\linewidth]{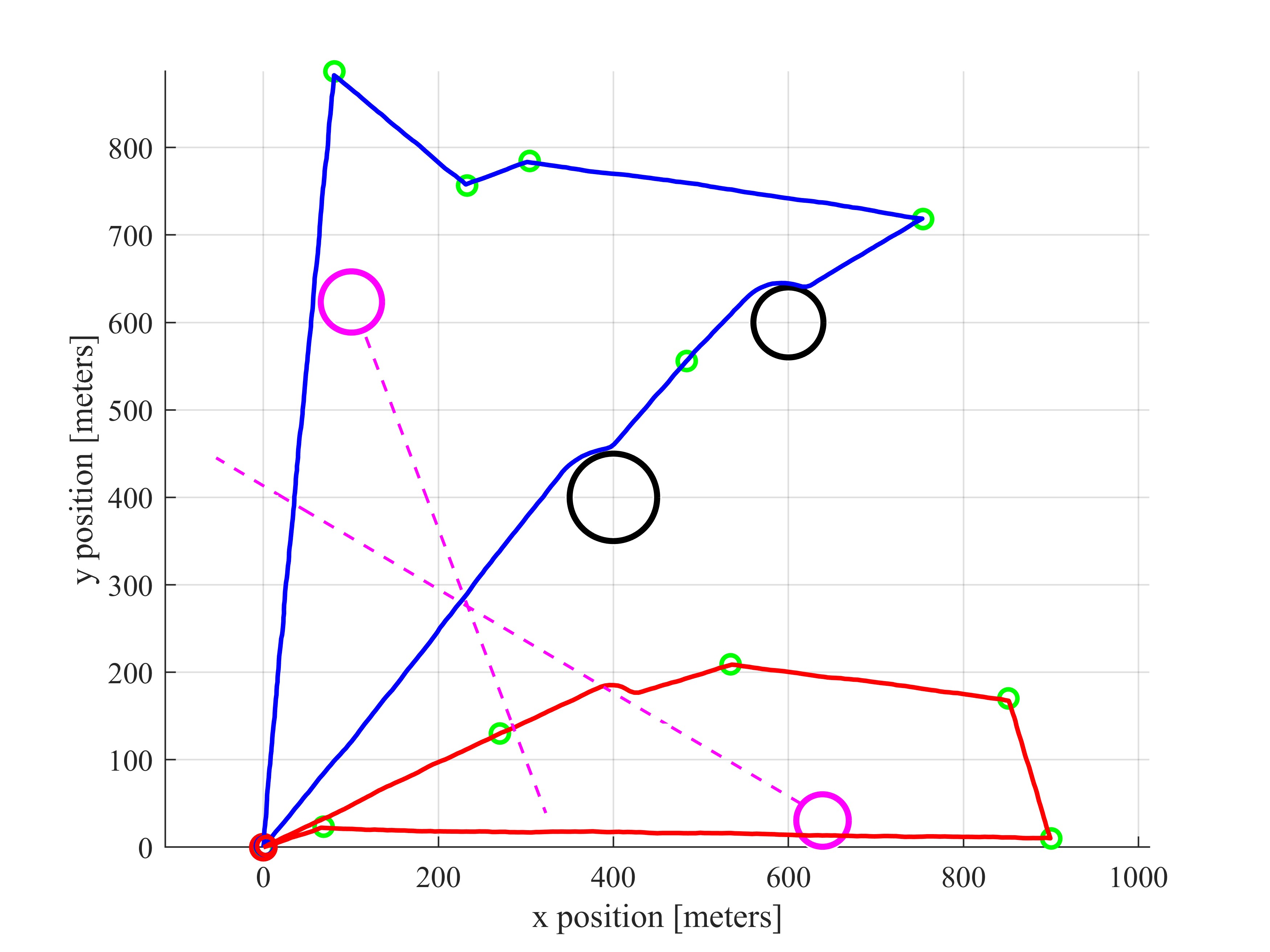}
    \subcaption{}
  \end{subfigure}
  \caption{EKF-assisted state estimation and collision avoidance.
    (a) Real-time trajectory prediction and correction using EKF-based belief updates.
    (b) Avoidance of static and dynamic obstacles through sensor fusion and predictive motion modeling.}
  \label{fig:ekf-obstacles}
\end{figure}

\begin{figure}[ht!]
  \centering
  \begin{subfigure}[t]{0.50\linewidth}
    \centering
    \includegraphics[width=\linewidth]{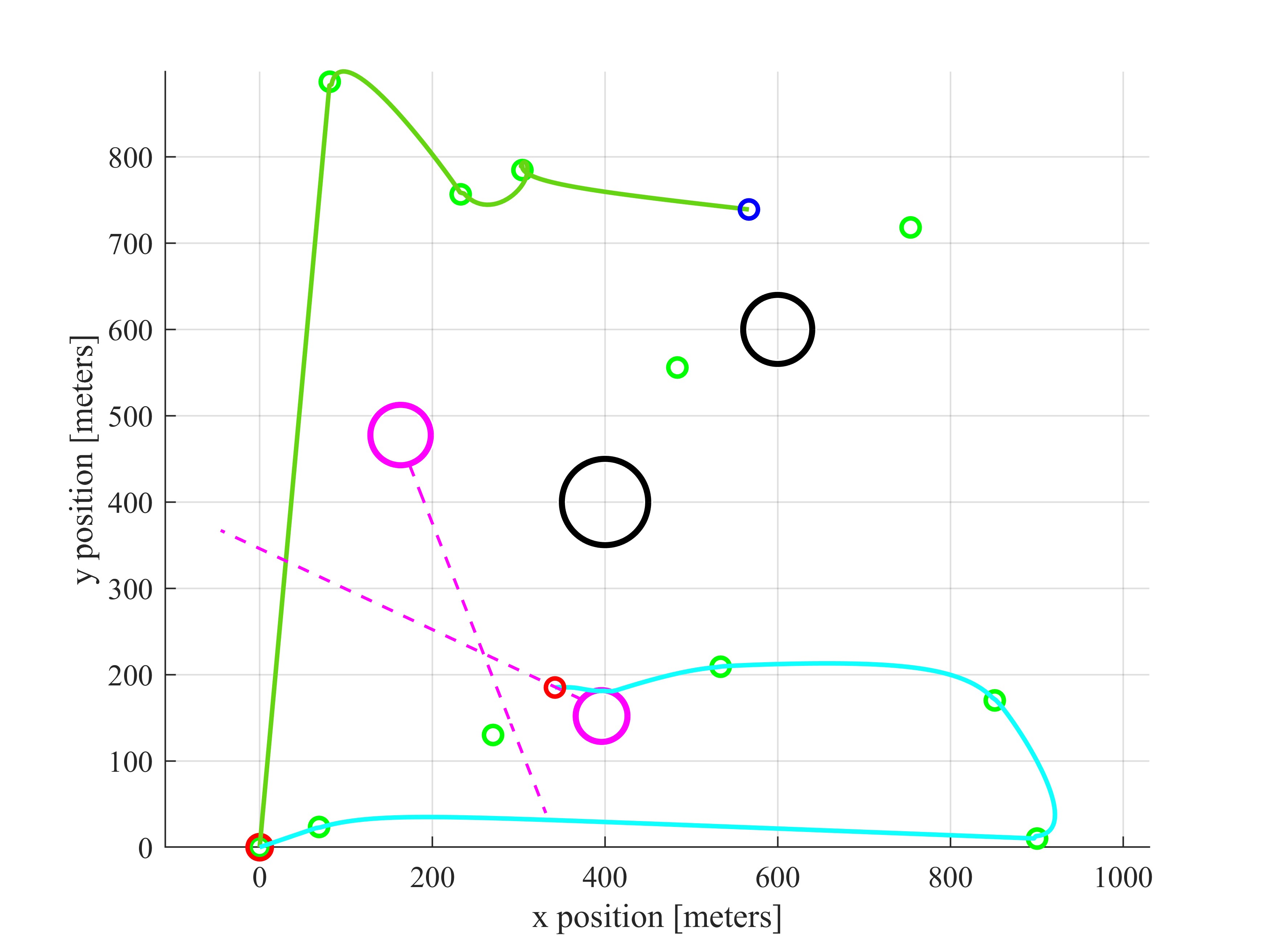}
    \subcaption{}
  \end{subfigure}\hfill
  \begin{subfigure}[t]{0.50\linewidth}
    \centering
    \includegraphics[width=\linewidth]{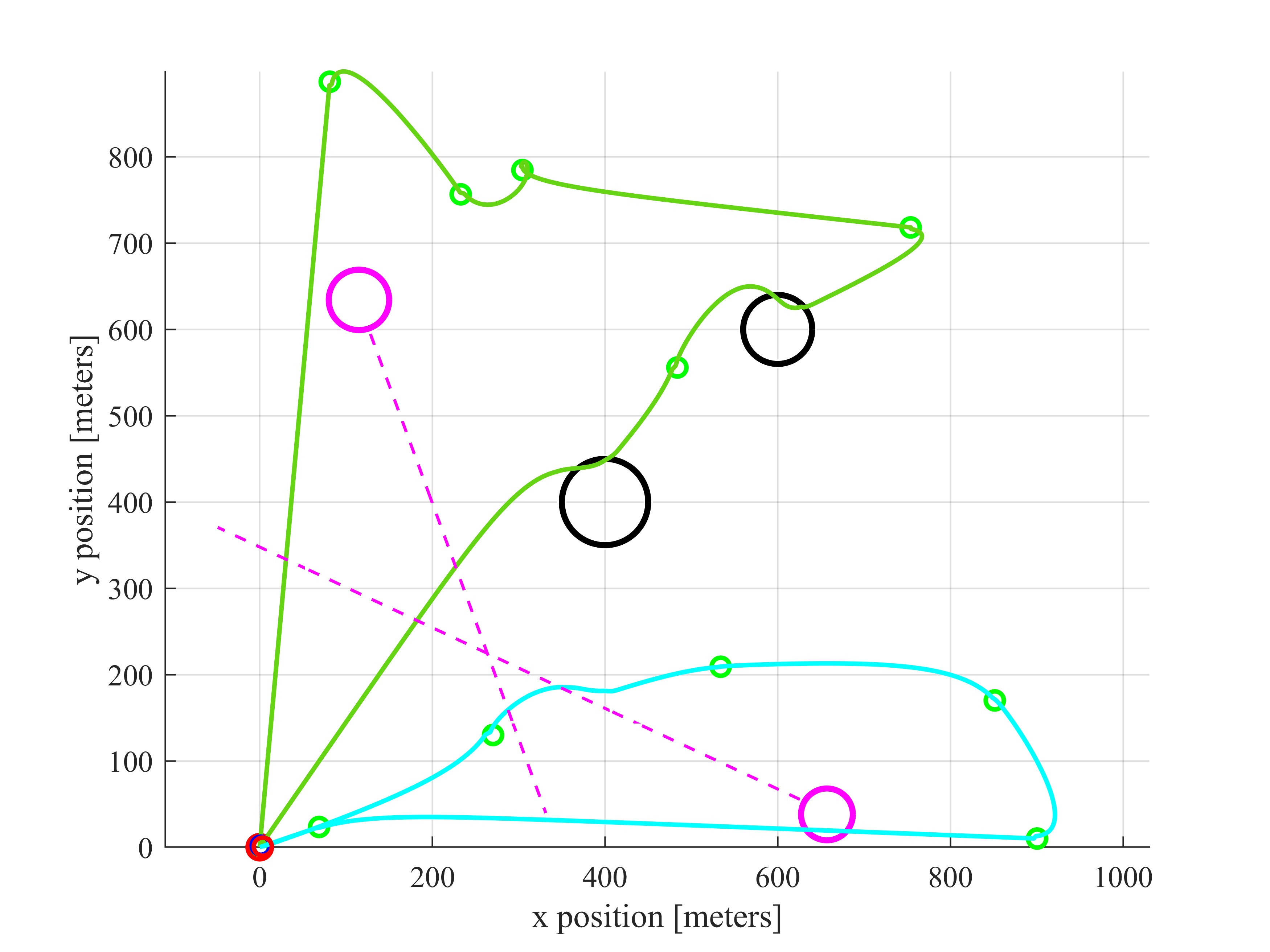}
    \subcaption{}
  \end{subfigure}
  \caption{PF-assisted state estimation and collision avoidance.
(a) Reactive trajectory correction under nonlinear dynamics.
(b) Robust avoidance of static and dynamic obstacles, demonstrating improved accuracy in uncertain environments.}
  \label{fig:Pf-obstacles}
\end{figure}

Figs.~\ref{fig:ekf-obstacles} and \ref{fig:Pf-obstacles} illustrate EKF- and PF-assisted trajectory execution and collision avoidance, respectively. The EKF provides efficient local prediction and correction when the dynamics are locally smooth, while the PF provides stronger robustness under nonlinear and uncertain conditions. In both cases, predicted proximity to obstacles or other UAVs increases the motion-level mismatch cost, causing the inference layer to favor motion words with stronger repulsive behavior.

\begin{figure}[ht]
\centering
\includegraphics[width=0.30\textwidth]{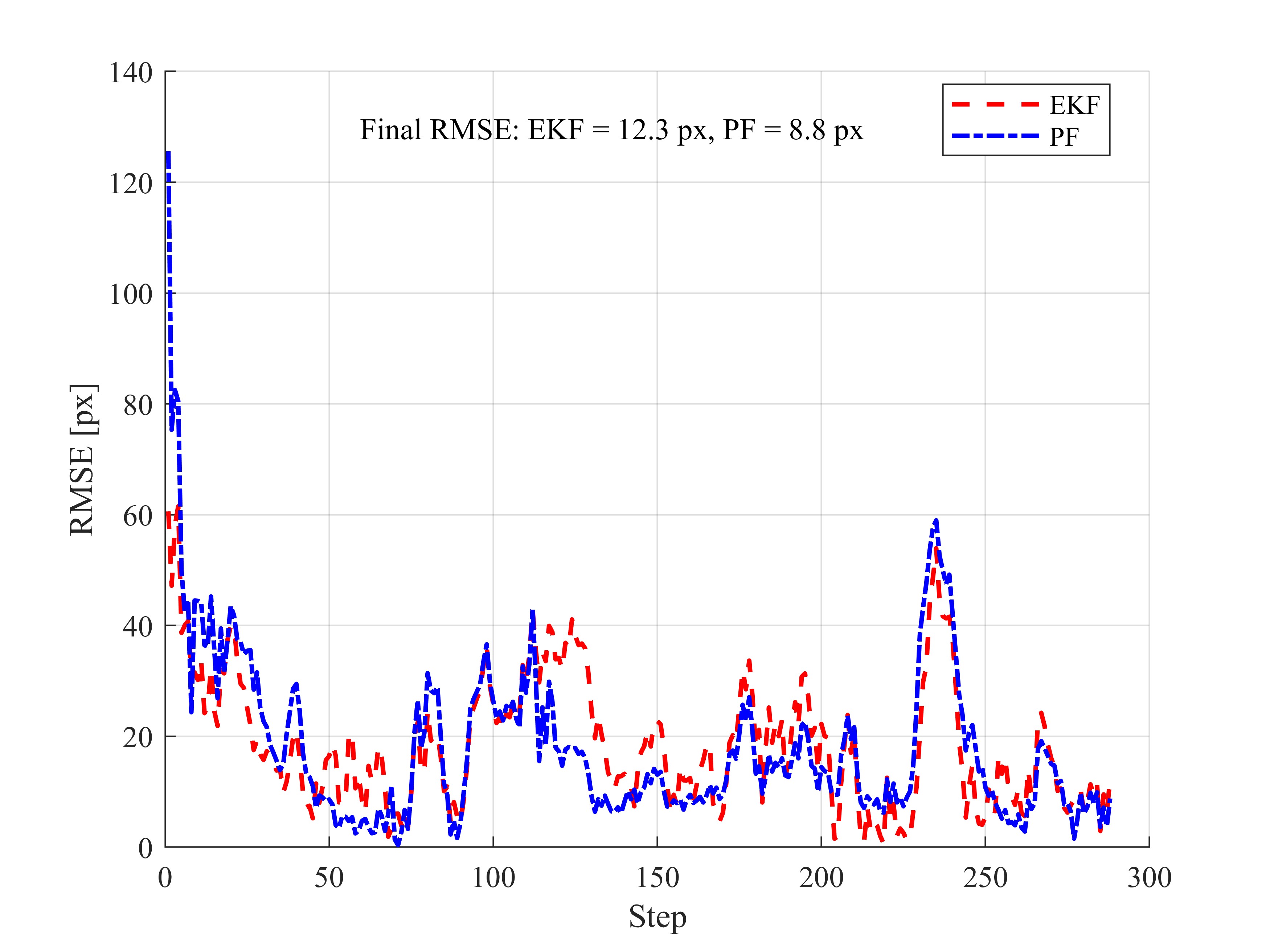} 
\caption{Quantitative comparison of state estimation accuracy.
Root-mean-square error (RMSE) of EKF and PF estimates over time, showing lower PF error under nonlinear conditions and a favorable accuracy--complexity trade-off for EKF.}
\label{fig:Fig19} 
\end{figure}

Fig.~\ref{fig:Fig19} compares EKF and PF estimation accuracy in terms of RMSE. The PF achieves lower error in nonlinear conditions, whereas the EKF remains attractive for computationally efficient online implementation. This result confirms the role of the filter-assisted module as a continuous-state correction mechanism coupled to the symbolic motion-level inference process.

%

\subsection{Comparative Evaluation with Baseline Methods}
\label{subsec:baseline_comparison}

The proposed framework is compared with the GA--RF expert planner and a modified Q-learning baseline. GA--RF represents the expert offline optimizer used to generate demonstrations, while modified Q-learning represents a reward-driven learning baseline trained using the same expert data. The purpose of this comparison is to evaluate whether the proposed world model preserves expert-like structure while improving online stability and reducing the need for repeated global optimization.

\begin{figure}[ht!]
  \centering
  \begin{subfigure}[t]{0.50\linewidth}
    \centering
    \includegraphics[width=\linewidth]{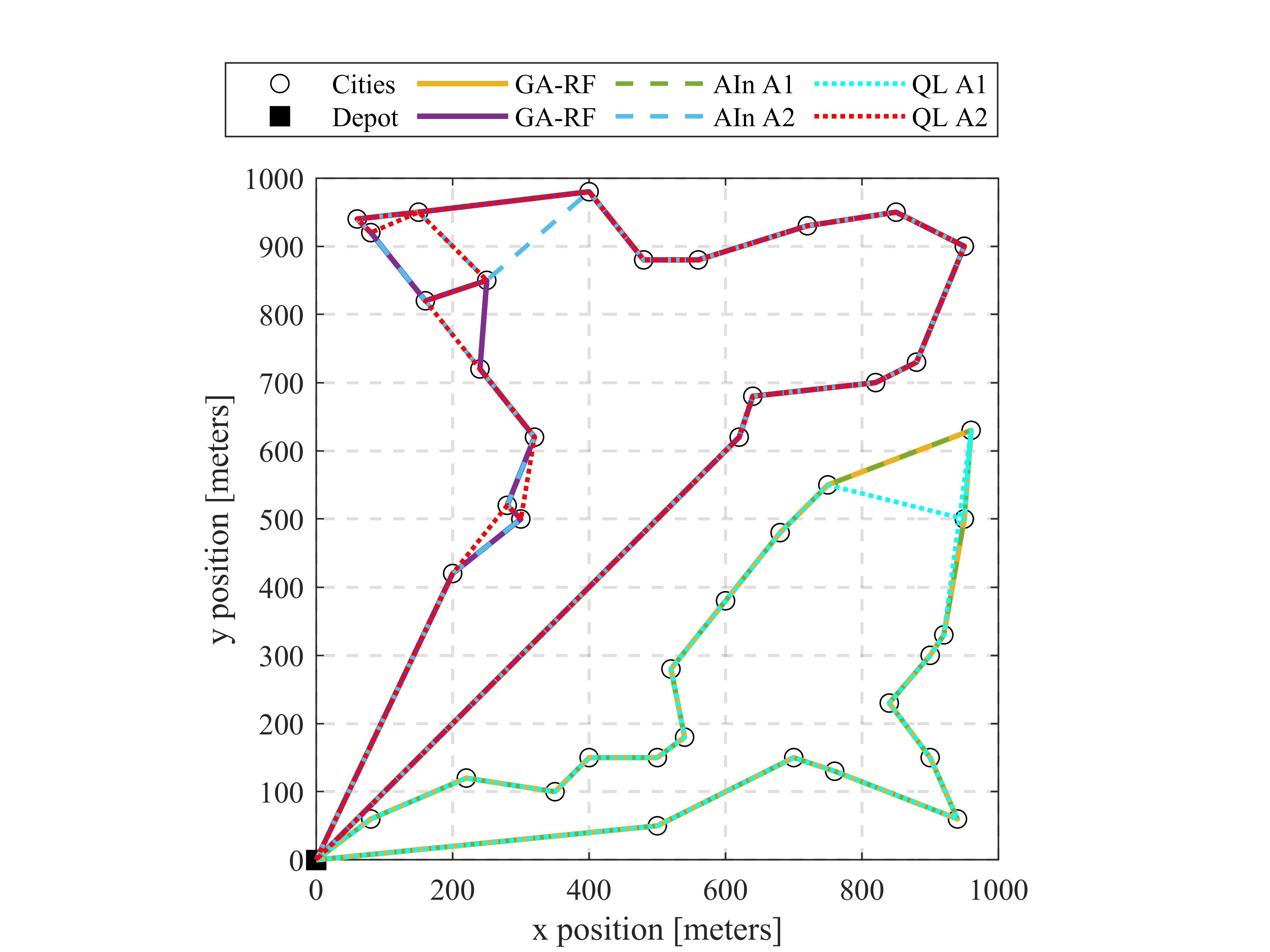}
    \subcaption{40 towns.}
  \end{subfigure}\hfill
  \begin{subfigure}[t]{0.50\linewidth}
    \centering
    \includegraphics[width=\linewidth]{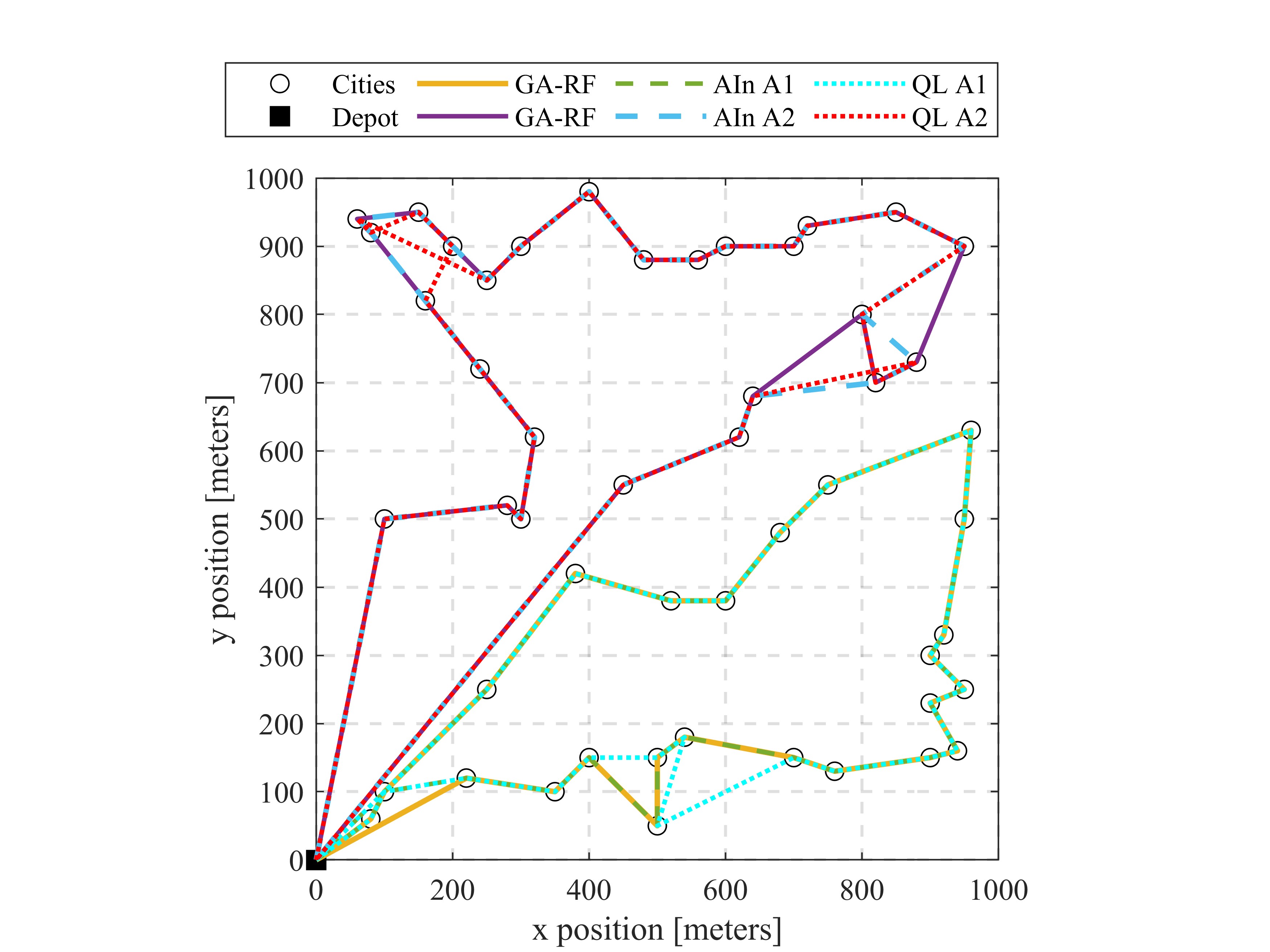}
    \subcaption{50 towns.}
  \end{subfigure}
  \caption{Qualitative comparison between the proposed inference-driven framework and modified Q-learning.
Results for missions with 40 and 50 targets show that the proposed method maintains smoother and more coherent trajectories, while modified Q-learning exhibits oscillatory behavior at the motion level.}
  \label{fig:comparison-traj}
\end{figure}

Fig.~\ref{fig:comparison-traj} qualitatively compares the proposed framework with modified Q-learning for missions with 40 and 50 targets. The proposed method produces smoother trajectories and maintains stronger belief--action consistency. In contrast, modified Q-learning exhibits more oscillatory motion-level behavior, which is expected when action selection is driven by learned values without an explicit hierarchical expert prior.

\begin{figure}[ht!]
  \centering
  \begin{subfigure}[t]{0.48\linewidth}
    \centering
    \includegraphics[width=\linewidth]{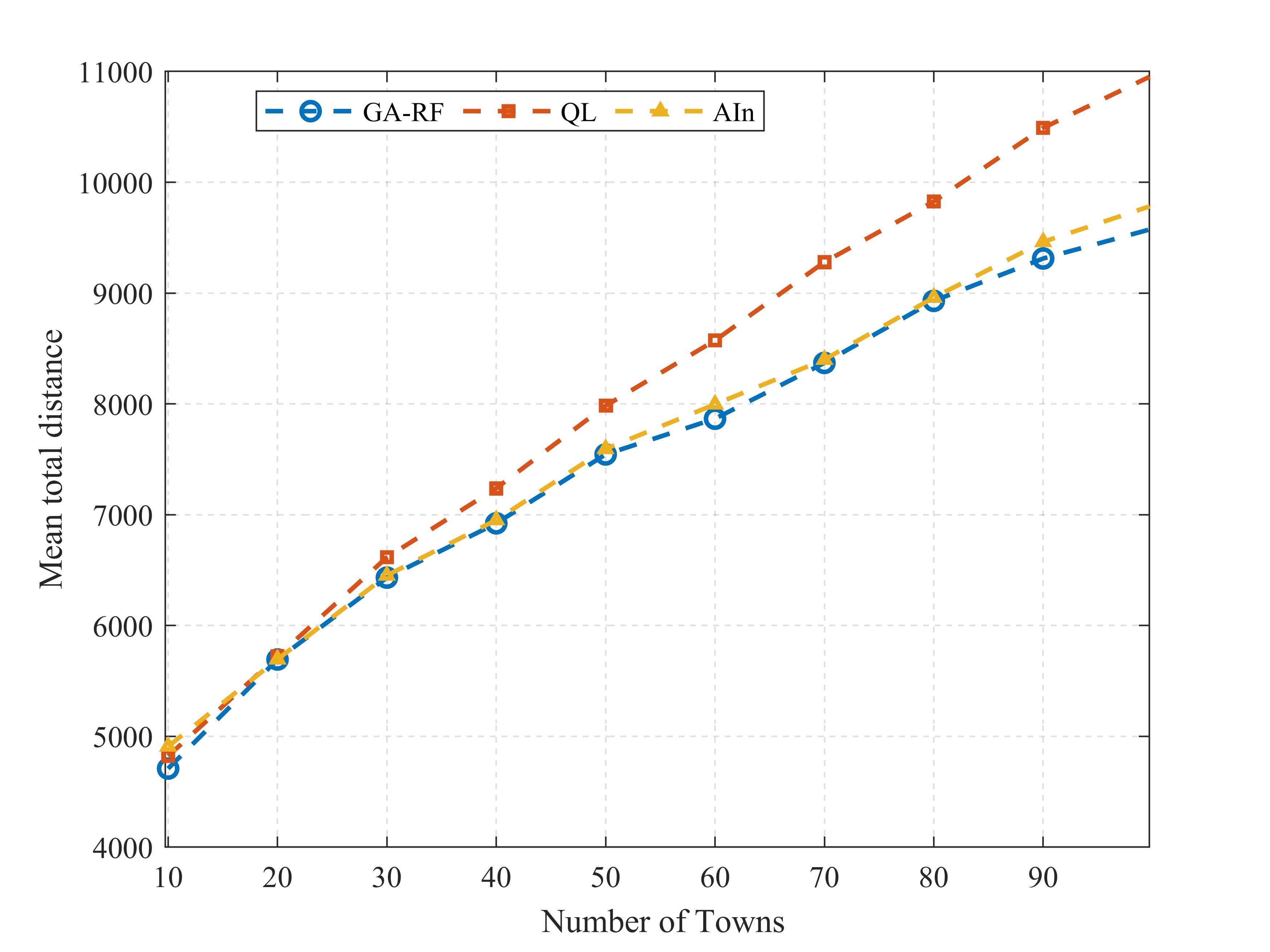}
    \subcaption{Mission completion time.}
  \end{subfigure}
  \hfill
  \begin{subfigure}[t]{0.48\linewidth}
    \centering
    \includegraphics[width=\linewidth]{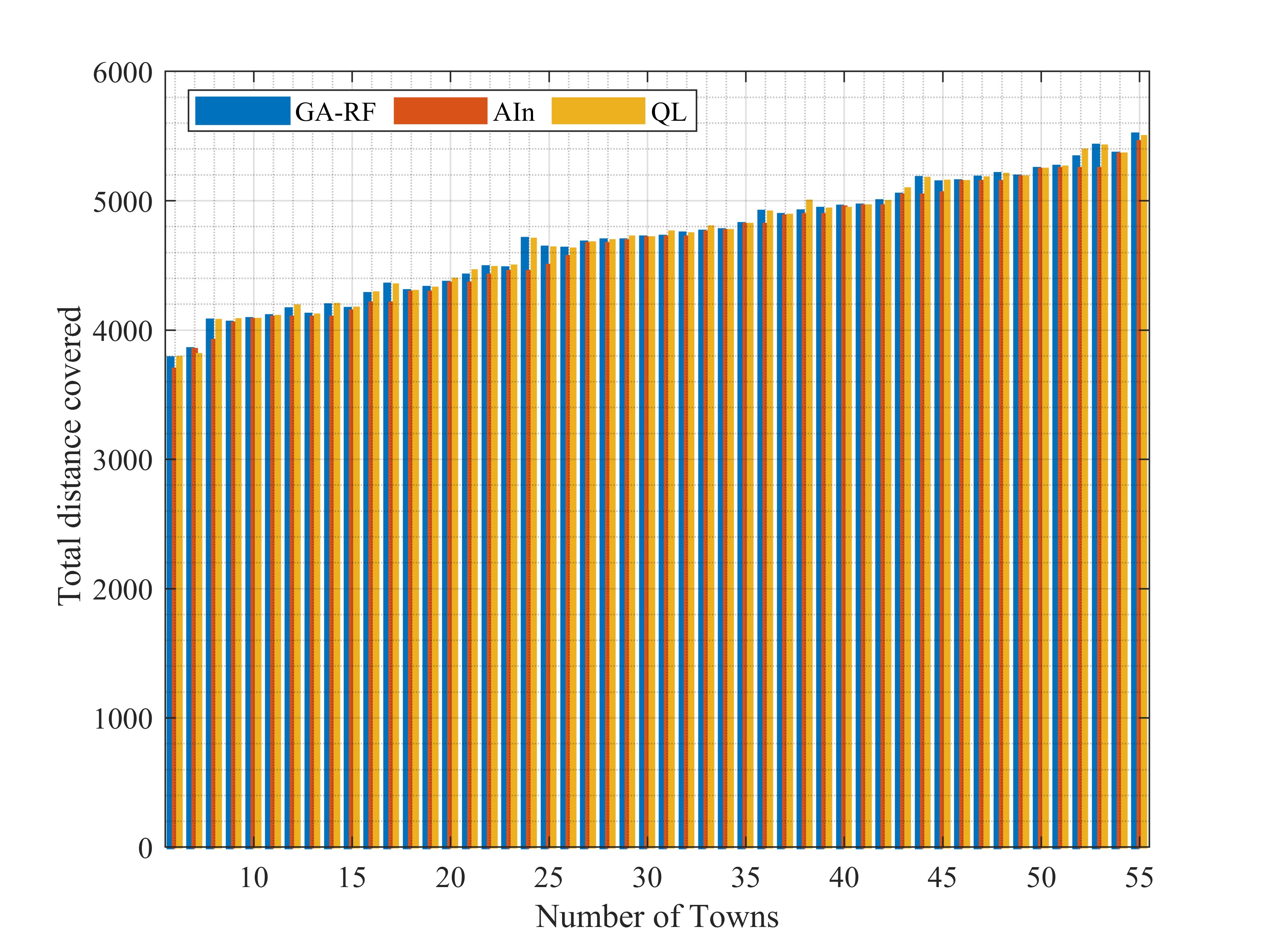}
    \subcaption{Total traveled distance.}
  \end{subfigure}
  \caption{Quantitative comparison with baseline methods.
(a) Mission completion time.
(b) Total traveled distance.
The proposed framework preserves expert-consistent mission efficiency while improving online stability relative to modified Q-learning.}
  \label{fig:comparison-performance}
\end{figure}

Fig.~\ref{fig:comparison-performance} reports mission completion time and total traveled distance for different mission sizes. The proposed framework achieves performance close to the GA--RF expert while enabling online inference from the learned world model. Compared with modified Q-learning, the proposed method provides lower completion time and more efficient trajectories, indicating that the expert-derived probabilistic priors improve online decision quality.

It is important to interpret the comparison with GA--RF carefully. GA--RF is an offline expert planner and serves as the source of demonstrations. The advantage of the proposed framework is not that it replaces GA--RF as a global optimizer, but that it reuses the structure learned from GA--RF to perform fast online inference and replanning under new conditions.

\begin{figure}[ht!]
  \centering
  \begin{subfigure}[t]{0.48\linewidth}
    \centering
    \includegraphics[width=\linewidth]{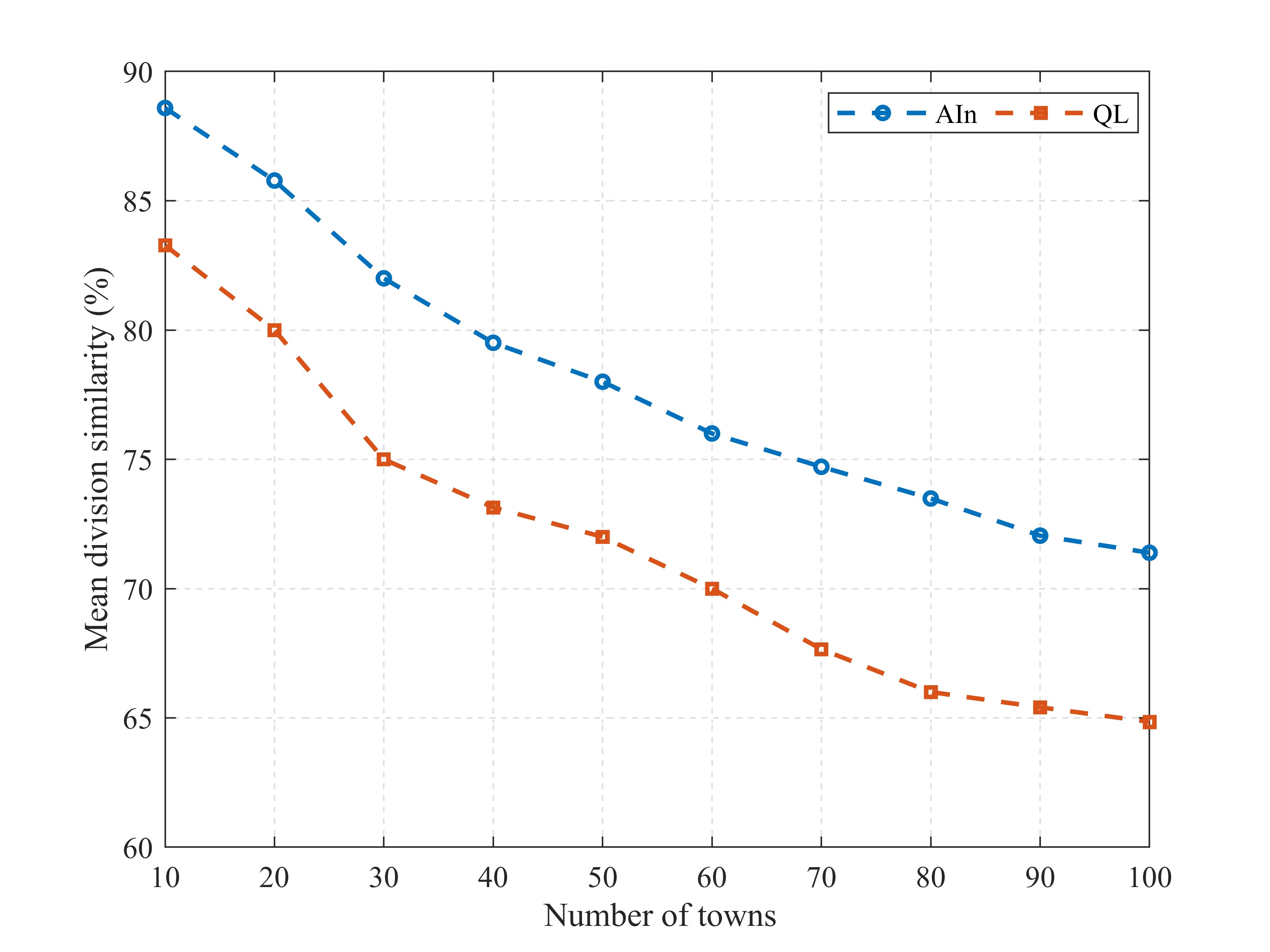}
    \subcaption{Division similarity.}
    \label{fig:similarity-division}
  \end{subfigure}
  \hfill
  \begin{subfigure}[t]{0.48\linewidth}
    \centering
    \includegraphics[width=\linewidth]{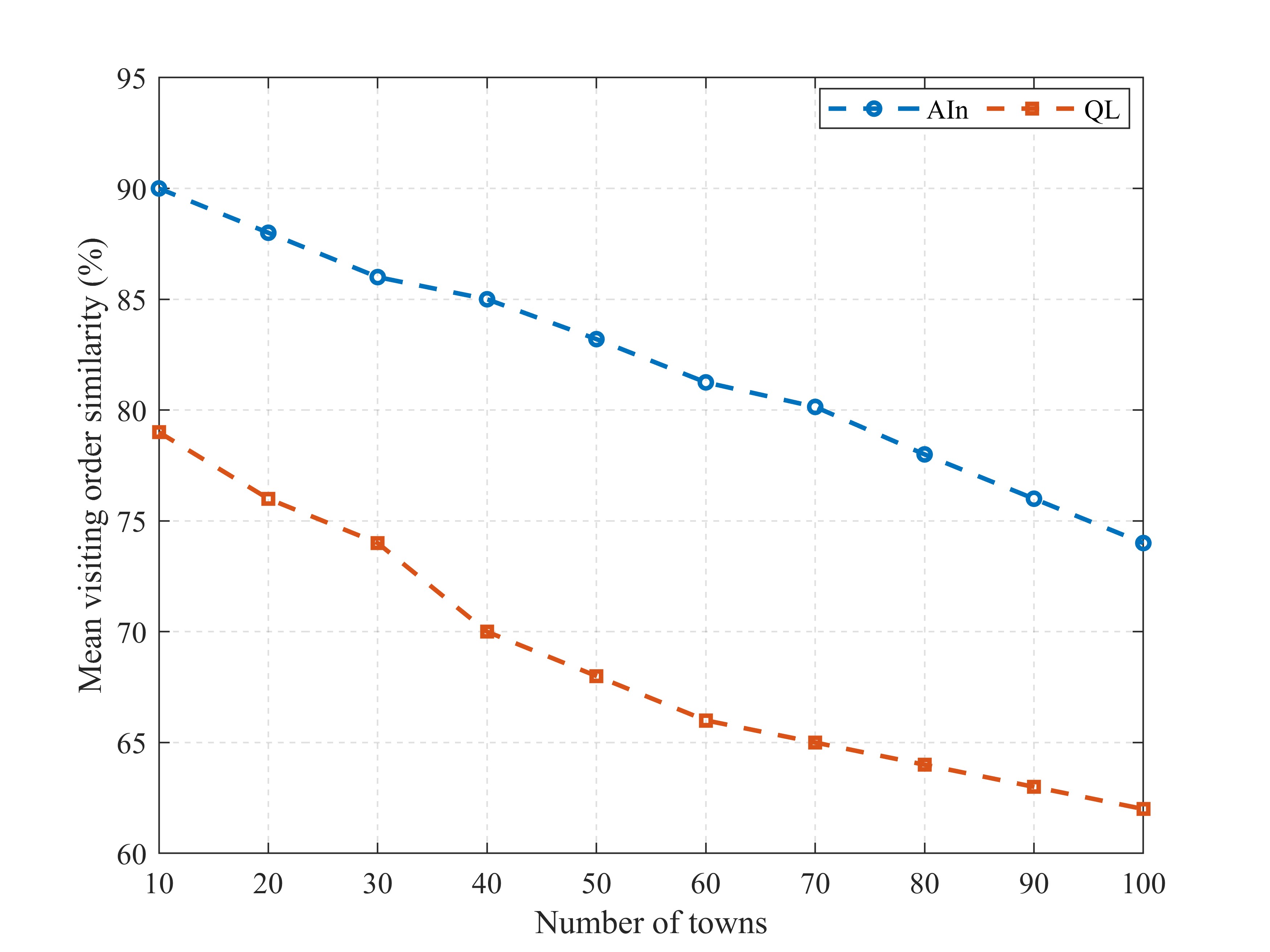}
    \subcaption{Visiting-order similarity.}
    \label{fig:similarity-order}
  \end{subfigure}
  \caption{Similarity with GA--RF expert demonstrations.
(a) Mission-division similarity.
(b) Visiting-order similarity.
The proposed framework better preserves expert allocation and ordering structure than modified Q-learning.}
  \label{fig:divison-order}
\end{figure}

Fig.~\ref{fig:divison-order} evaluates similarity with the GA--RF expert at the mission-division and visiting-order levels. The proposed framework remains closer to the expert strategy than modified Q-learning, indicating that the learned world model preserves structural regularities of expert planning. This does not imply simple memorization of demonstrated trajectories; rather, it shows that the probabilistic-symbolic representation captures reusable allocation and ordering patterns that generalize to unseen scenarios.

\begin{figure}[ht]
\centering
\includegraphics[width=0.30\textwidth]{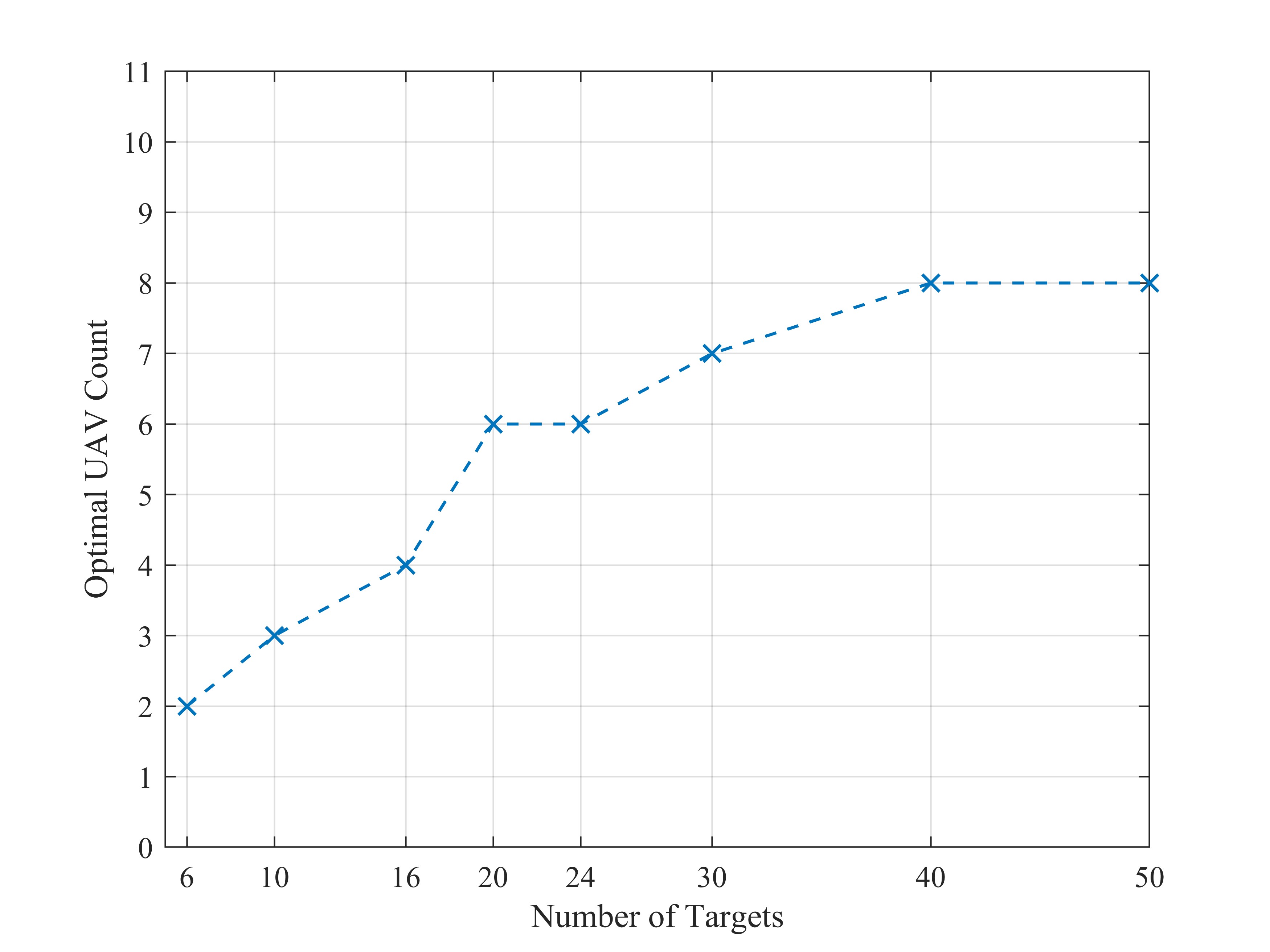}
  \caption{Inferred swarm size as a function of the number of targets.
The learned transition model increases the selected swarm size as task complexity grows, maintaining coverage efficiency while limiting mission uncertainty.}
  \label{fig:swarmsize1}
\end{figure}

Fig.~\ref{fig:swarmsize1} shows the inferred swarm size as a function of the number of targets. As the number of targets increases, the learned transition model selects a larger number of UAVs, reflecting the expert tendency to increase swarm capacity when mission complexity grows. This result further confirms that the world model captures high-level mission-organization patterns rather than only local trajectory shapes.

%

\subsection{Validation Using Real-Flight Trajectory Data}
\label{subsec:real_flight_validation}

To assess robustness under realistic sensing noise and non-ideal motion dynamics, real-flight data collected from indoor experiments at Universidad Carlos III de Madrid (UC3M) are incorporated into the validation pipeline. Two DJI Air~2S UAVs are operated by human pilots via Wi-Fi and flown between predefined target points inside an indoor laboratory.

The UAV platform is equipped with a front-facing RGB camera with \(3840\times2160\) resolution at 30~Hz, dual IMU units including accelerometer, gyroscope, and magnetometer, and a visual--inertial odometry (VIO) system for position and orientation estimation. Ten fixed target points are defined, together with a central depot serving as the start and end location for all flights. Each target is visited by only one UAV to avoid path interference, and the flight altitude is maintained between 10 and 15~feet.

The goal of this part is not to claim full outdoor autonomous deployment. Instead, it evaluates whether the symbolic world model and belief-correction mechanism remain valid when the input trajectories contain human-pilot variability, sensing noise, and non-smooth motion patterns.

\begin{figure}[ht]
\centering
\includegraphics[width=0.50\textwidth]{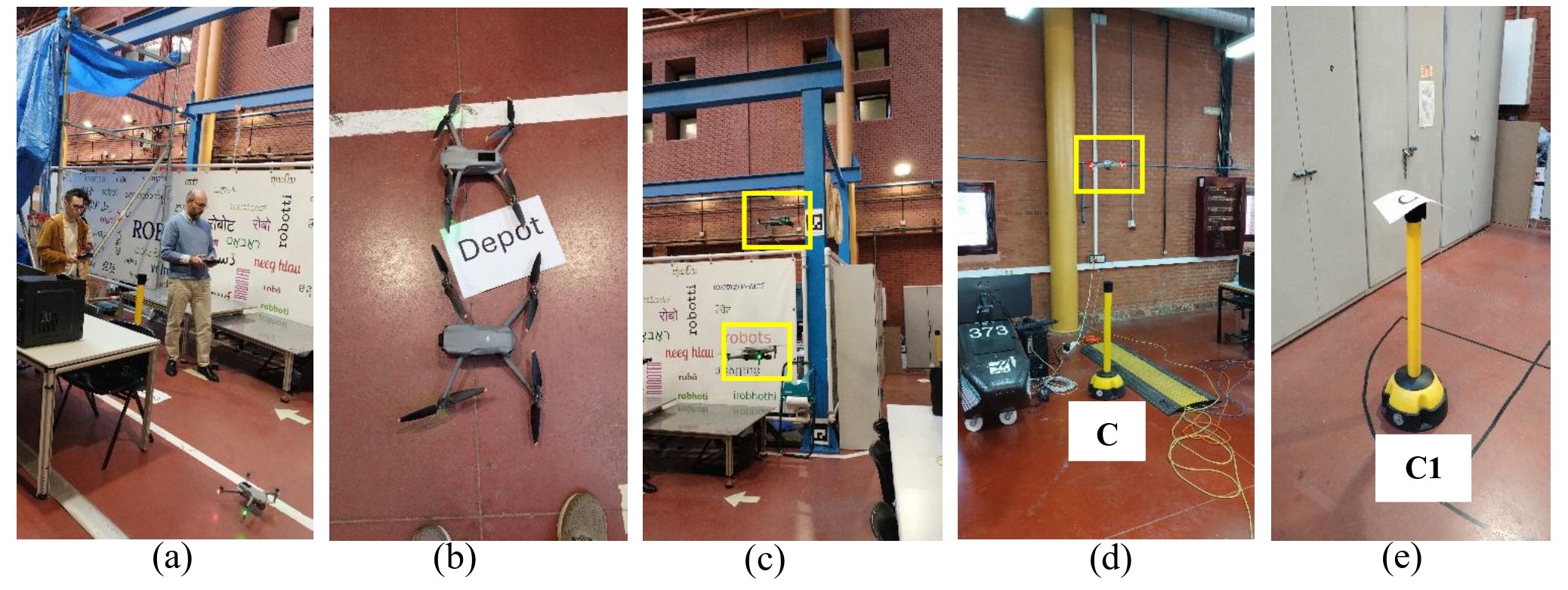} 
\caption{Indoor data-collection process for real-flight experiments.
(a) UAV control via Wi-Fi by a human operator.
(b) Depot initialization.
(c) Mission start.
(d) Target traversal.
(e) Example target location used for data extraction.}
\label{fig:exp1} 
\end{figure}

\begin{figure}[ht]
\centering
\includegraphics[width=0.30\textwidth]{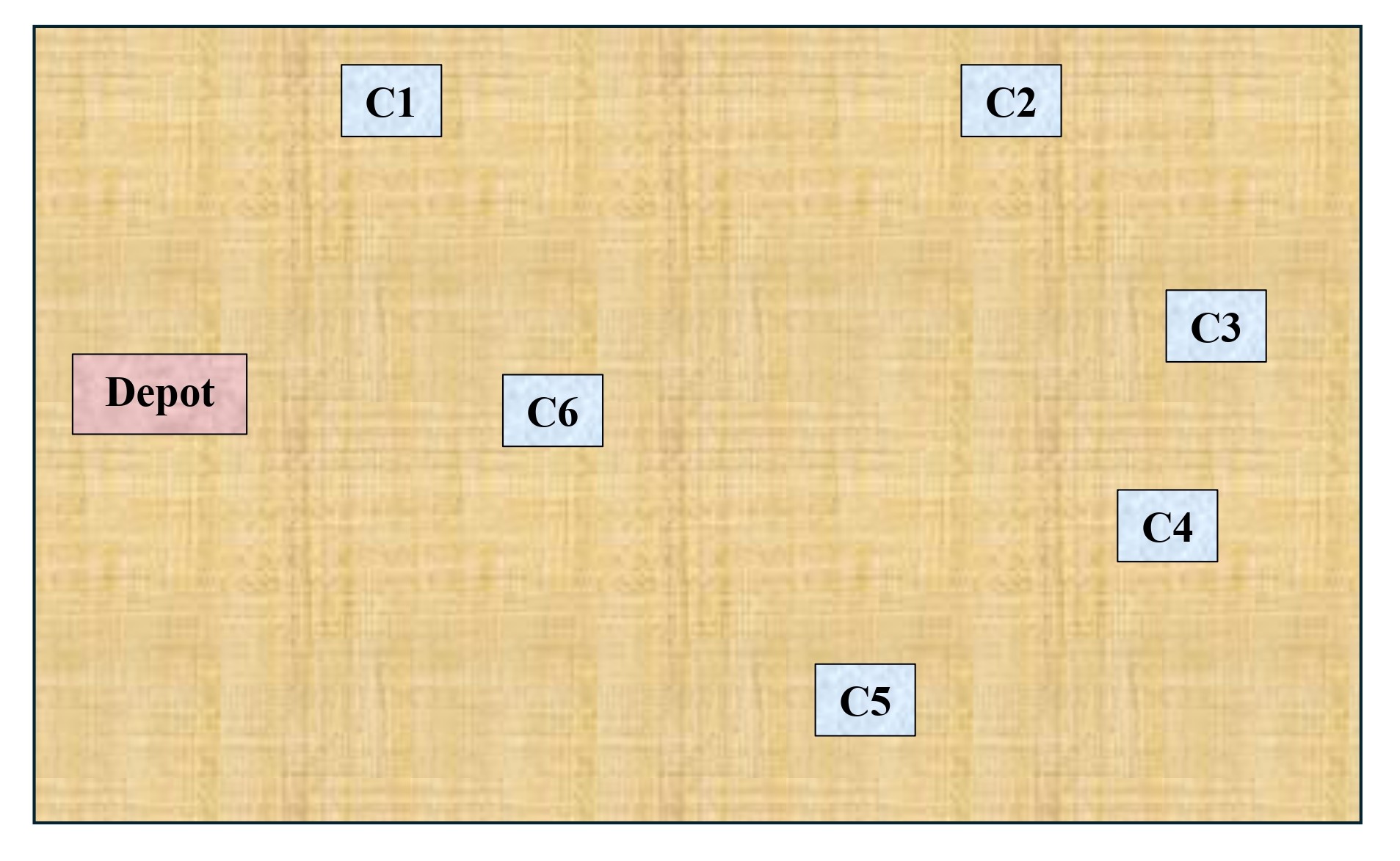} 
\caption{Abstracted representation of target locations used for simulation-based validation.
The abstraction preserves inter-target relationships while enabling flexible spatial arrangements for testing.}
\label{fig:exp2} 
\end{figure}

Fig.~\ref{fig:exp1} illustrates the indoor data-collection process, including UAV control, depot initialization, mission start, target traversal, and an example target point. Fig.~\ref{fig:exp2} shows the abstracted target representation used for validation. This abstraction preserves the relative inter-target structure while allowing the experimental trajectories to be incorporated into the simulation-based world-model evaluation.


\begin{figure}[ht!]
  \centering
  \begin{subfigure}[t]{0.48\linewidth}
    \centering
    \includegraphics[width=\linewidth]{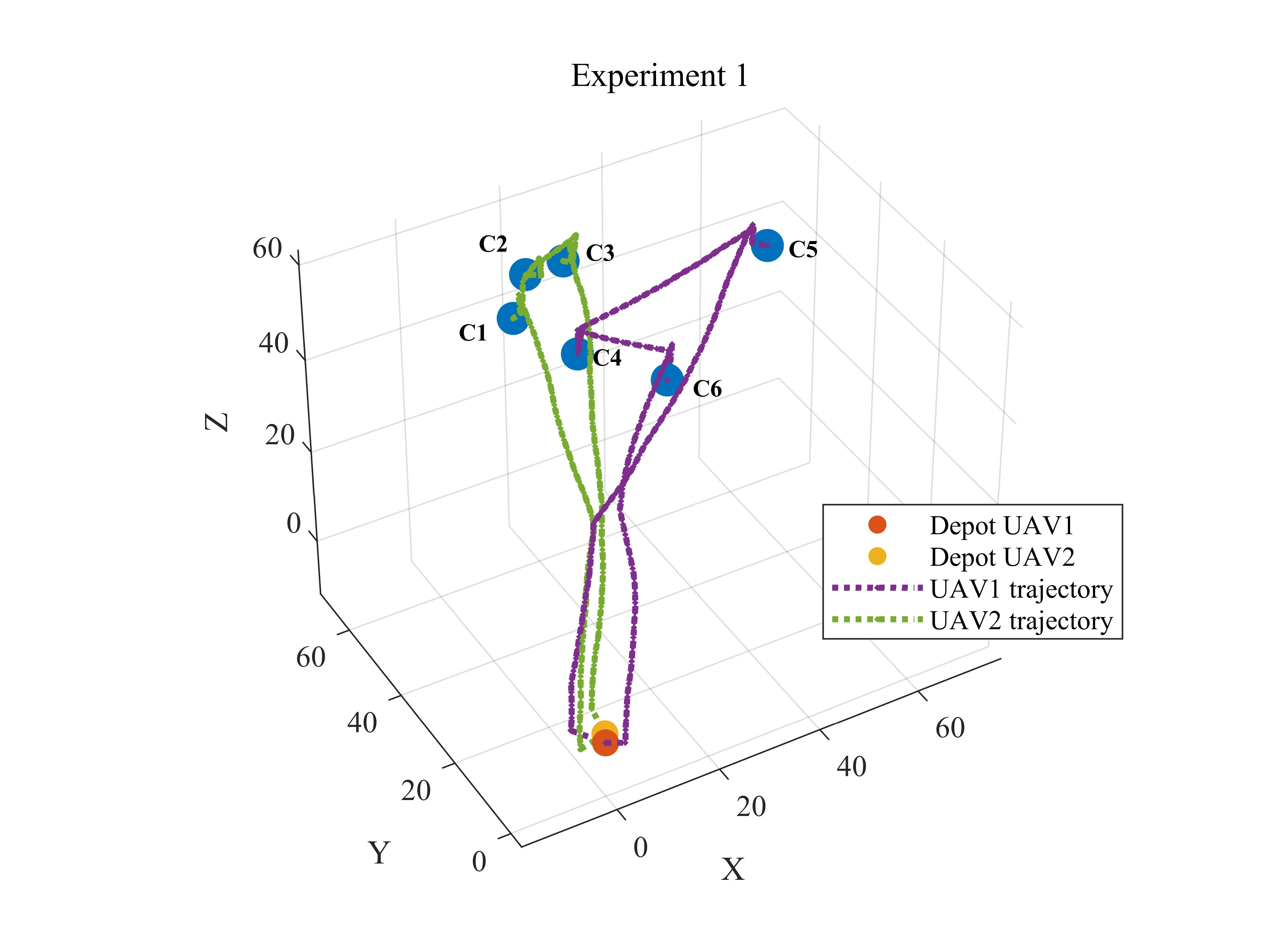}
    \subcaption{}
    \label{fig:Exp1}
  \end{subfigure}
  \hfill
  \begin{subfigure}[t]{0.48\linewidth}
    \centering
\includegraphics[width=\linewidth]{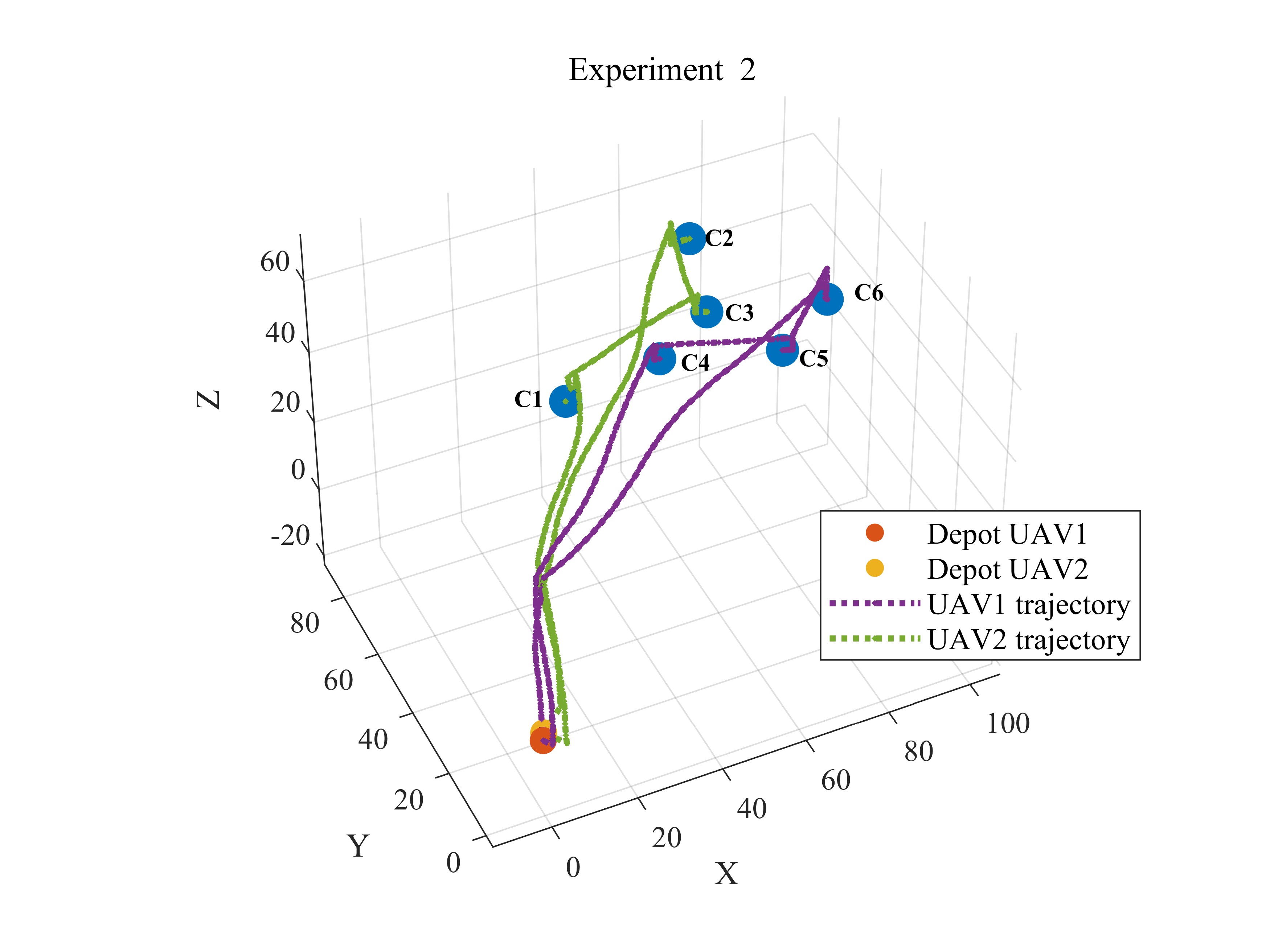}
    \subcaption{}
    \label{fig:Exp2}
  \end{subfigure}
  \begin{subfigure}[t]{0.48\linewidth}
    \centering
    \includegraphics[width=\linewidth]{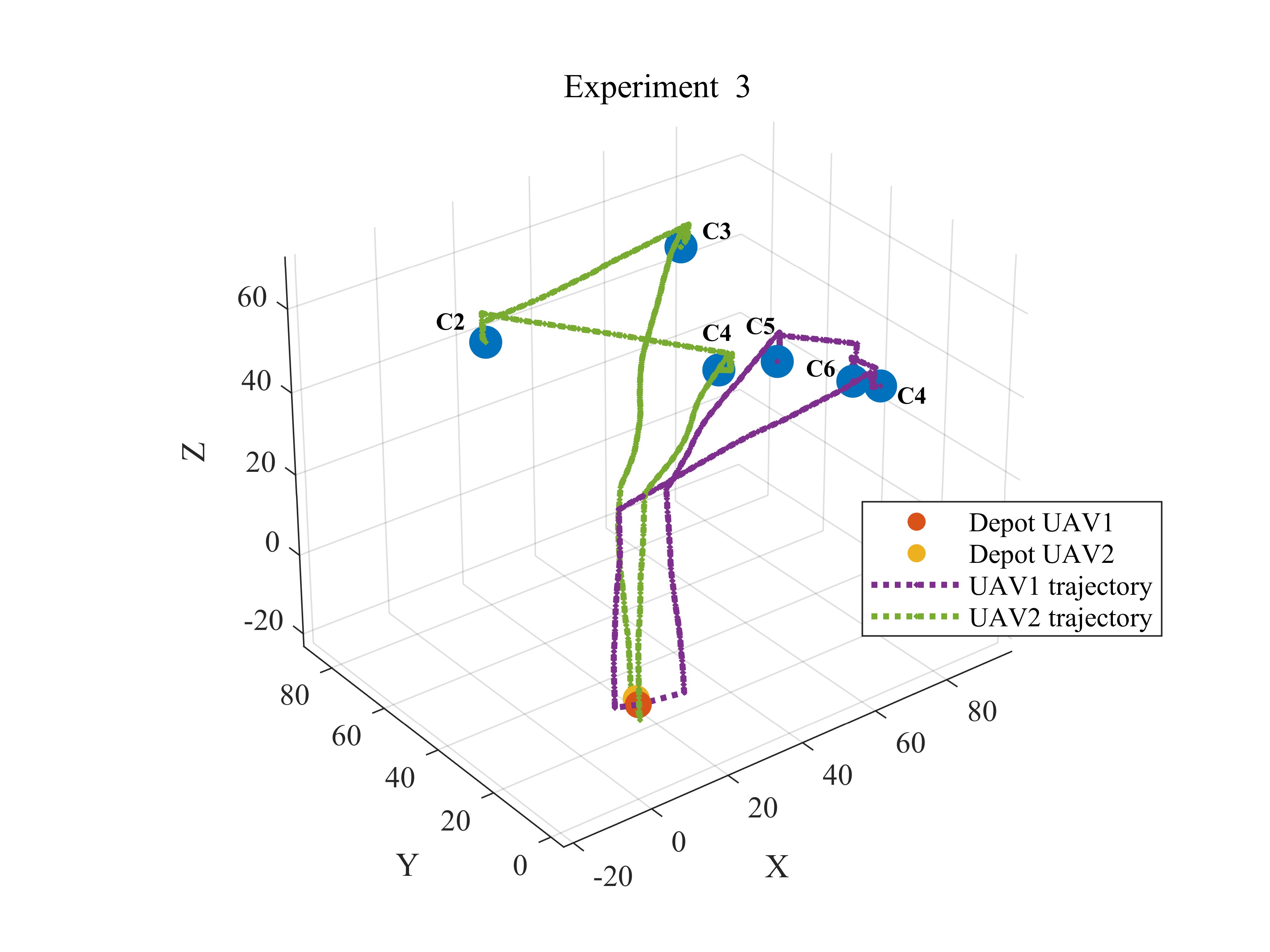}
    \subcaption{}
    \label{fig:Exp3}
  \end{subfigure}
  \caption{Representative real-flight experiments used for validation.
The trajectories show human-operated UAV flights between target points and illustrate variability in route structure and local motion behavior.}
  \label{fig:Experiments}
\end{figure}

Fig.~\ref{fig:Experiments} reports three representative real-flight experiments. In each experiment, six cities, denoted by \(C_n\), are visited by two human-operated UAVs. The pilots start from the central depot, visit the assigned cities, and return to the depot after completing the mission. The three experiments exhibit different route structures and local motion patterns, reflecting human-pilot variability, different city distributions, and practical differences in maneuver execution.

These trajectories are useful for validation because they are not perfectly smooth and do not exactly follow the simulated expert trajectories. They therefore provide a more challenging input for testing whether the proposed belief-update mechanism can correct symbolic predictions under non-ideal observations.

\begin{figure}[ht!]
  \centering
  \begin{subfigure}[t]{0.48\linewidth}
    \centering
    \includegraphics[width=\linewidth]{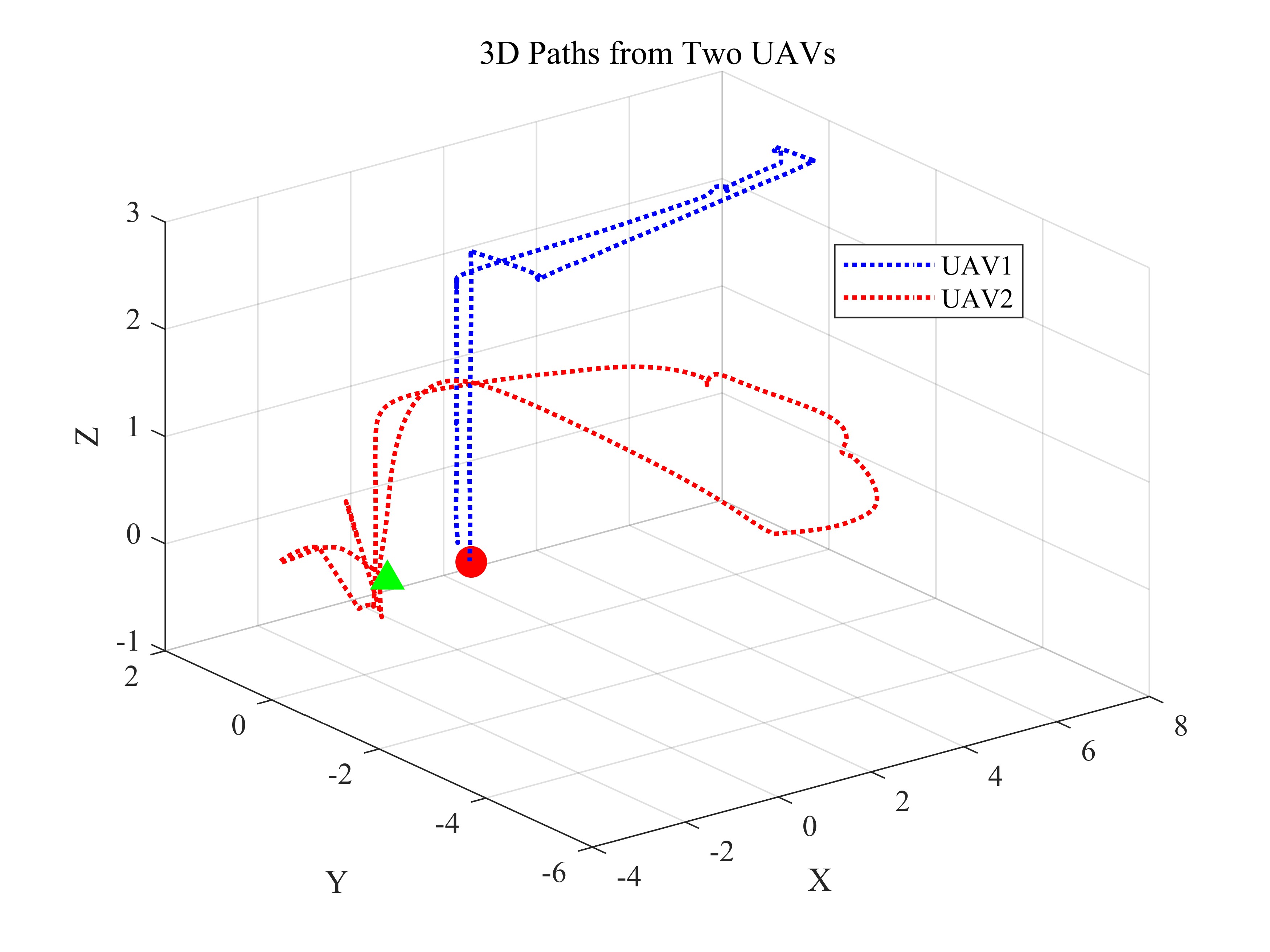}
    \subcaption{}
    \label{fig:trajectory-a}
  \end{subfigure}
  \hfill
  \begin{subfigure}[t]{0.48\linewidth}
    \centering
    \includegraphics[width=\linewidth]{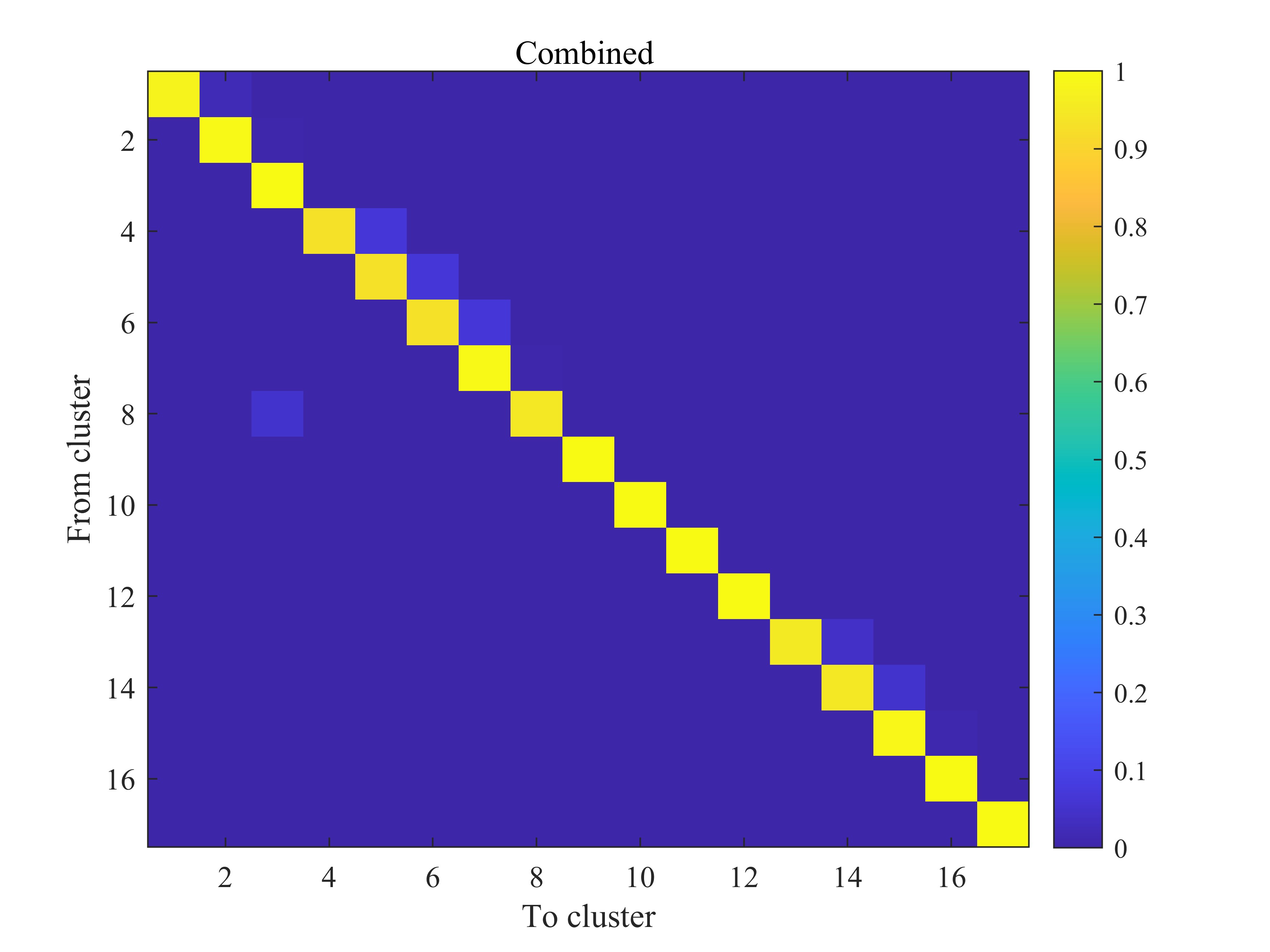}
    \subcaption{}
    \label{fig:matrix-b}
  \end{subfigure}
  \caption{Integration of real-flight data into the learned world model.
(a) Example UAV trajectories obtained from human-operated flights.
(b) Combined transition matrix derived from velocity clusters using the Growing Neural Gas algorithm.}
  \label{fig:trajectories-matrix}
\end{figure}

Fig.~\ref{fig:trajectories-matrix}(a) shows representative trajectories obtained from human-operated flights. Fig.~\ref{fig:trajectories-matrix}(b) shows the combined transition matrix derived from velocity-cluster sequences using the Growing Neural Gas algorithm. This step converts continuous real-flight trajectories into symbolic motion observations, making them compatible with the Motion Dictionary and the probabilistic world model.

%

\subsection{Belief Correction on Real-Flight Trajectory Clusters}
\label{subsec:real_flight_belief_correction}

\begin{figure}[ht]
\centering
\includegraphics[width=0.30\textwidth]{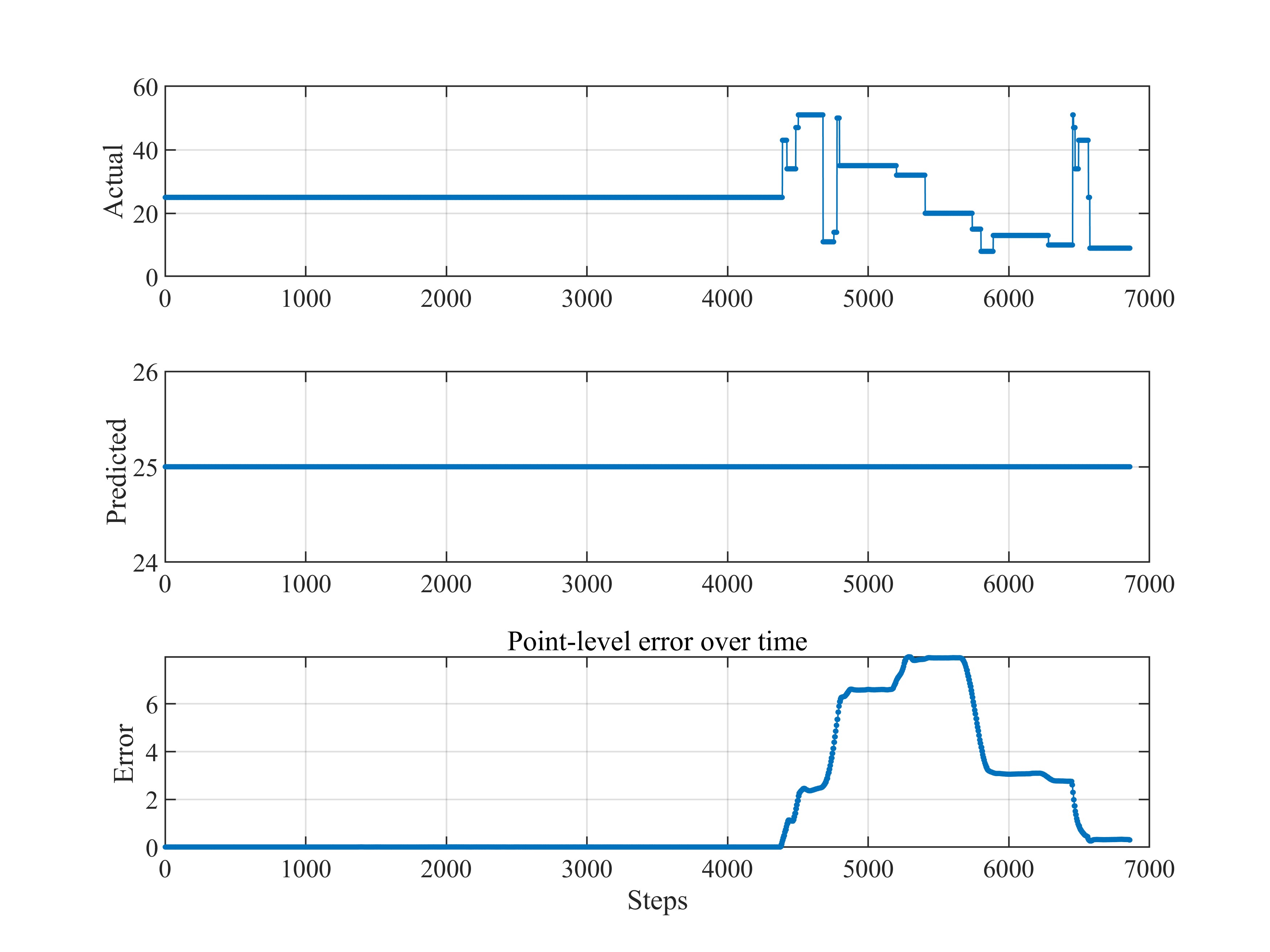} 
\caption{Trajectory prediction before Bayesian convergence.
Top: observed cluster labels.
Middle: predicted labels.
Bottom: prediction errors, highlighting large discrepancies before belief correction.}
\label{fig:prediction1} 
\end{figure}

\begin{figure}[h!]
\centering
\includegraphics[width=0.4\textwidth]{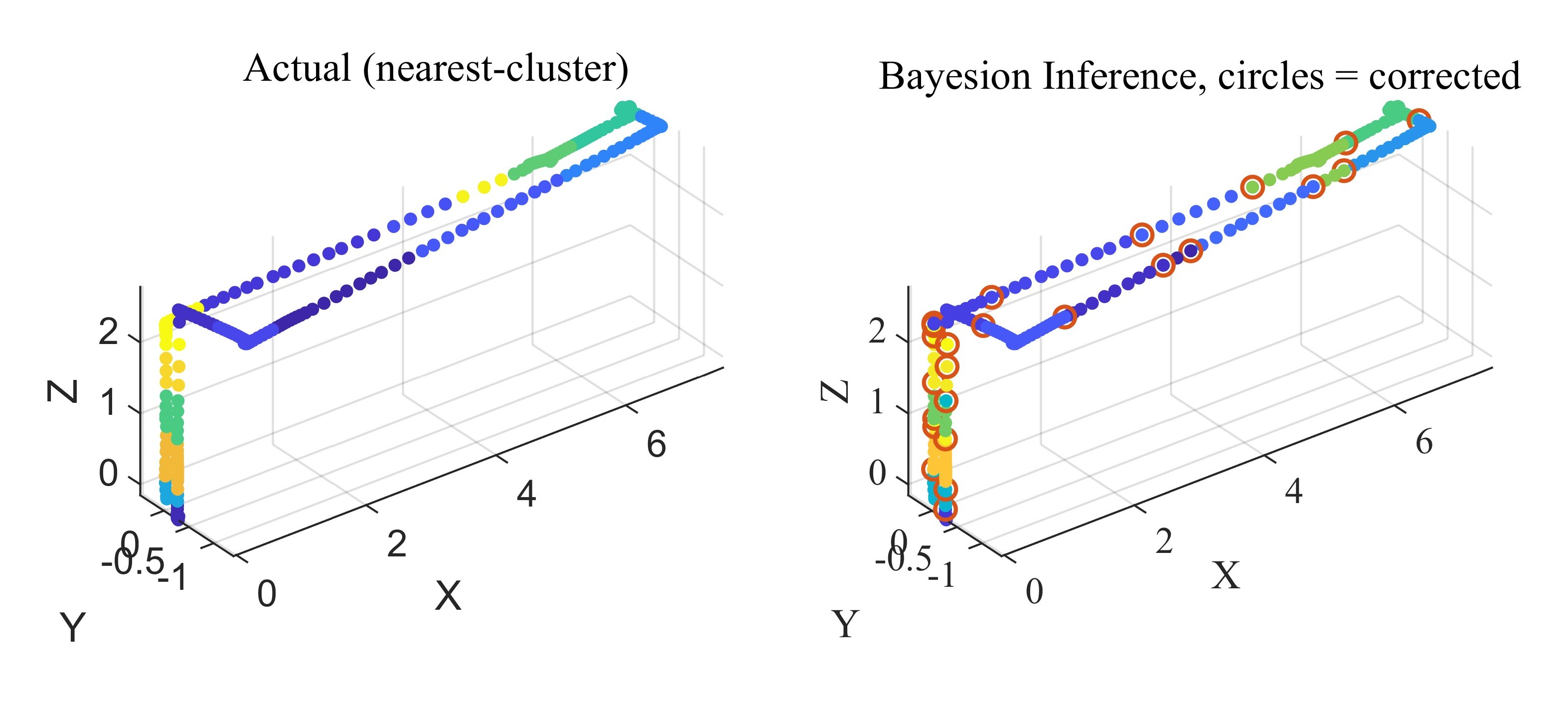}
\includegraphics[width=0.4\textwidth]{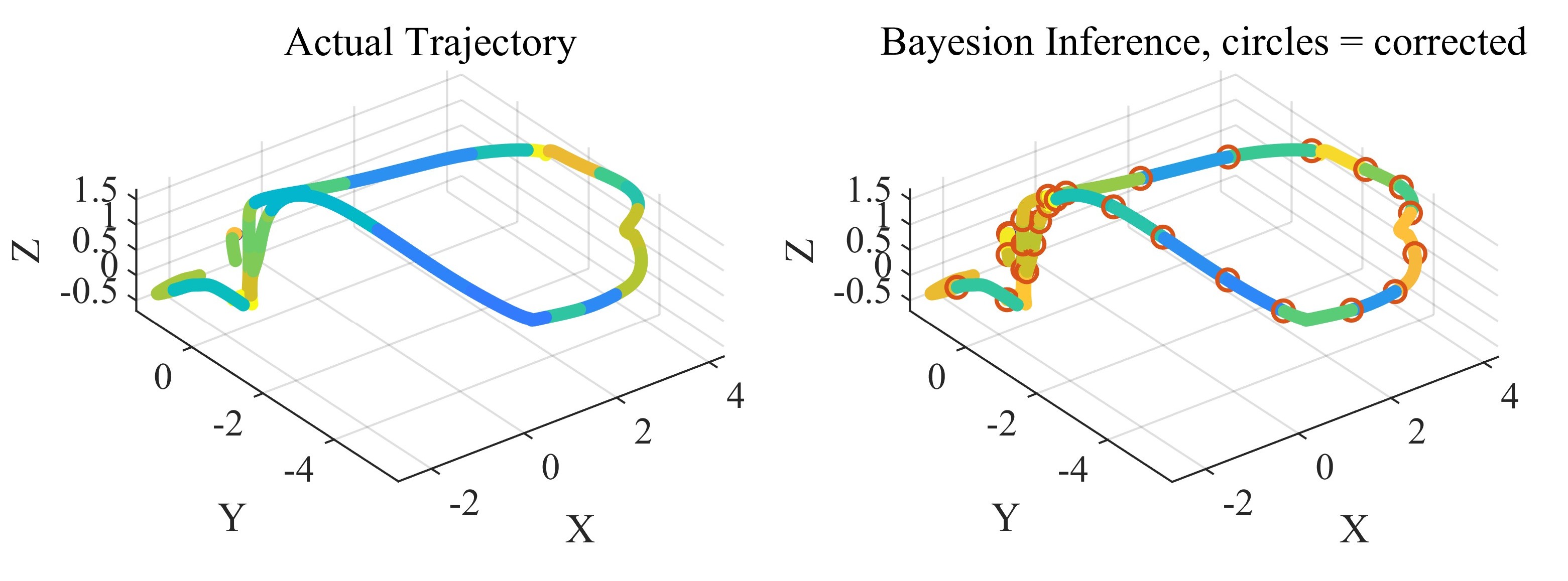}
\caption{Prediction results after Bayesian belief updating.
Left: observed 3D trajectory with cluster labels.
Right: predicted trajectory, where red circles indicate inference-based corrections.}
\label{fig:predictionandcorrection}
\end{figure}

\begin{figure}[ht]
\centering
\includegraphics[width=0.30\textwidth]{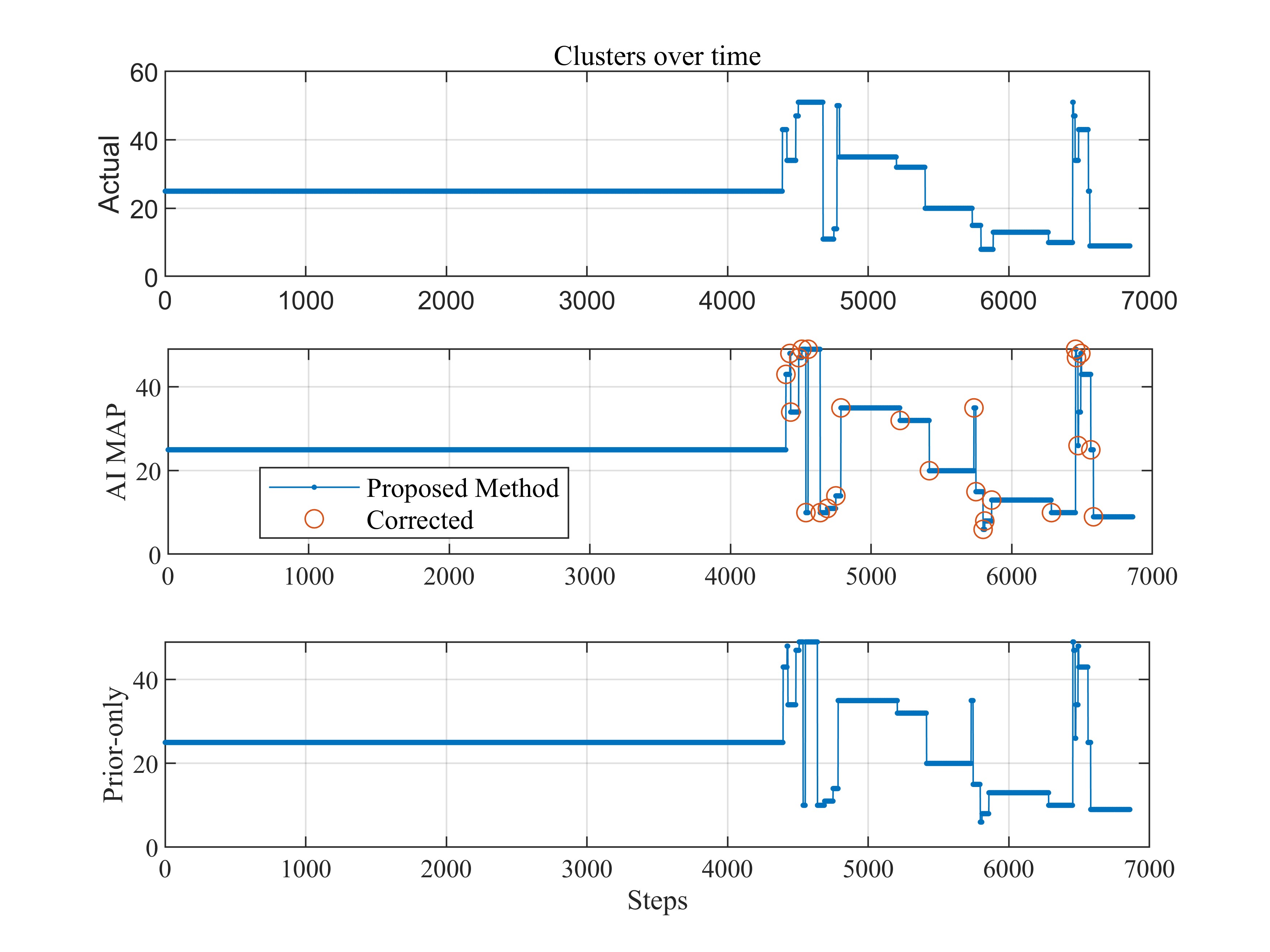} 
\caption{Comparison of original, predicted, and prior cluster labels.
Bayesian correction reduces prediction error and improves alignment with the observed real-flight trajectory labels.}
\label{fig:predictioncorrection} 
\end{figure}

Figs.~\ref{fig:prediction1}--\ref{fig:predictioncorrection} evaluate the belief-correction mechanism using real-flight trajectory clusters. Before Bayesian correction, the predicted symbolic labels deviate from the observed cluster sequence because the real-flight trajectories contain human-pilot variability and sensing noise. This produces large prediction errors, as shown in Fig.~\ref{fig:prediction1}.

After belief updating, the predicted symbolic trajectory becomes more consistent with the observed cluster sequence. Fig.~\ref{fig:predictionandcorrection} shows the corrected trajectory, with red circles indicating inference-based corrections. Fig.~\ref{fig:predictioncorrection} further compares the original, predicted, and prior cluster labels, showing that Bayesian correction reduces the prediction mismatch.

These results confirm that the proposed framework can adapt its symbolic predictions online when the observation statistics deviate from the prior world model. Importantly, this correction is achieved through belief updating and does not require retraining the dictionaries or transition matrices.

%

\subsection{Testing Under Unseen Target Layouts}
\label{subsec:unseen_conditions}

\begin{figure}[ht!]
  \centering
  \begin{subfigure}[t]{0.48\linewidth}
    \centering
    \includegraphics[width=\linewidth]{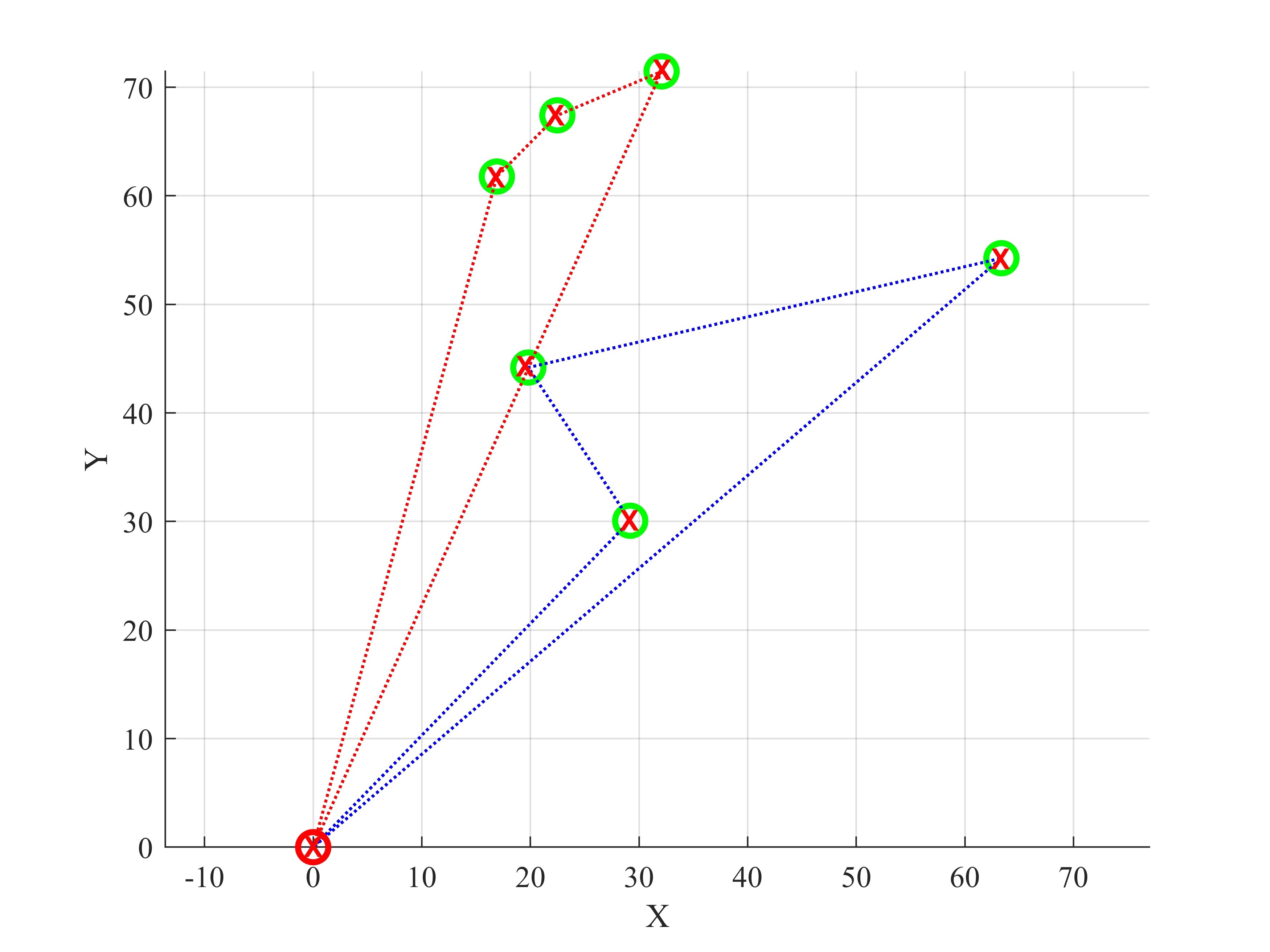}
    \subcaption{}
    \label{fig:Exp1Surp}
  \end{subfigure}
  \hfill
  \begin{subfigure}[t]{0.48\linewidth}
    \centering
\includegraphics[width=\linewidth]{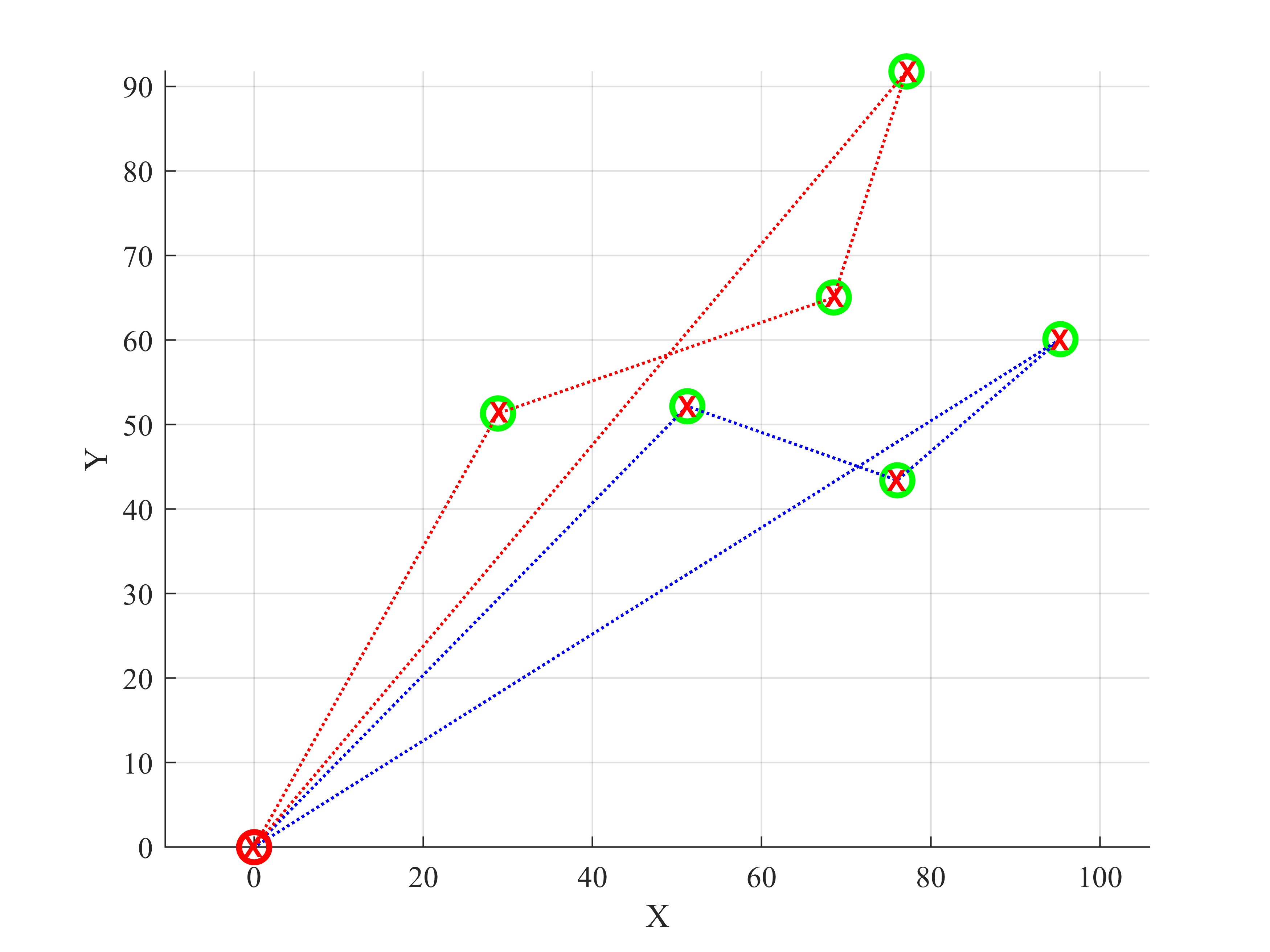}
    \subcaption{}
    \label{fig:Exp2Surp}
  \end{subfigure}
  \begin{subfigure}[t]{0.48\linewidth}
    \centering
    \includegraphics[width=\linewidth]{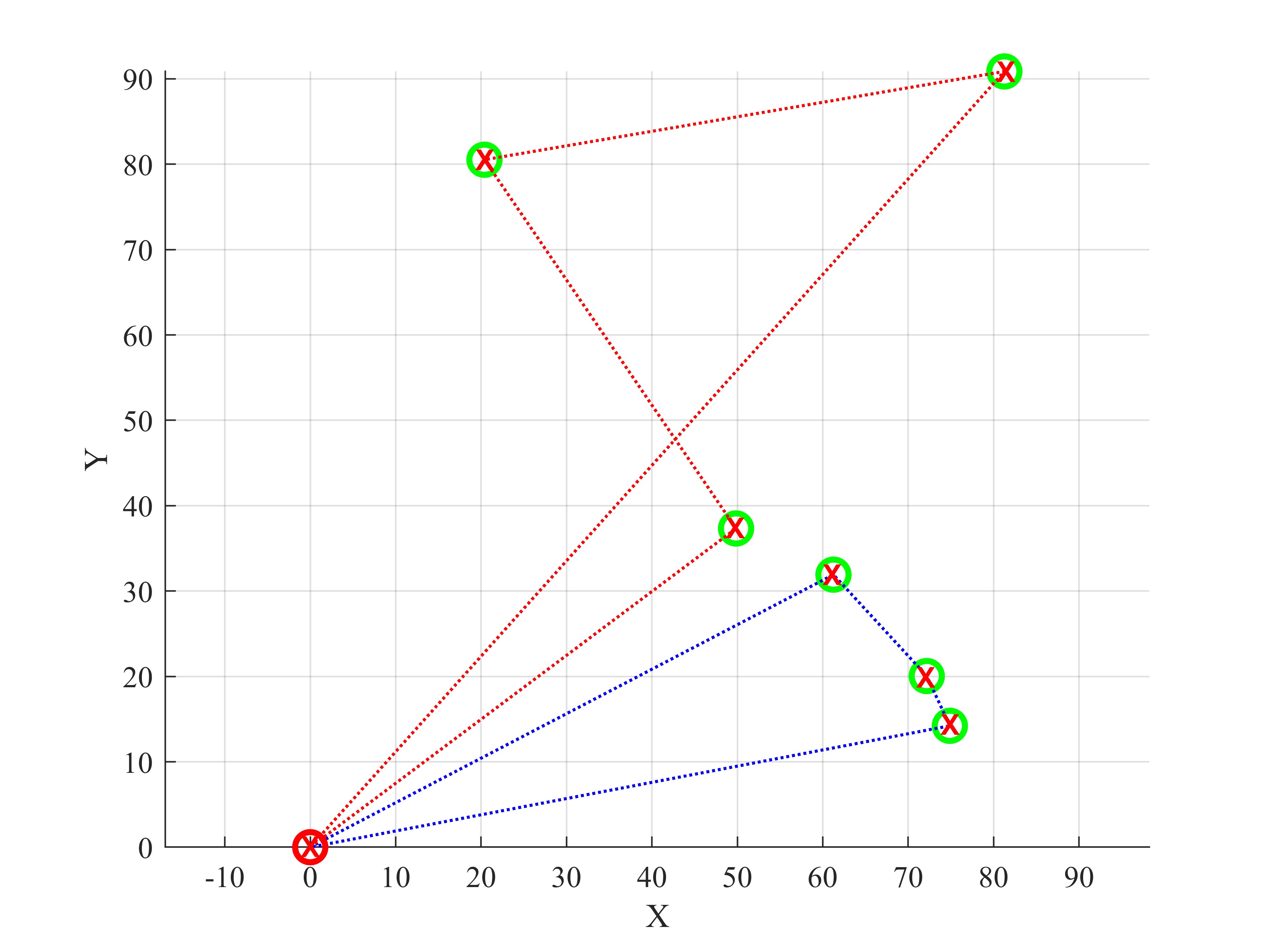}
    \subcaption{}
    \label{fig:Exp3Surp}
  \end{subfigure}
  \caption{Testing under unseen target layouts.
The proposed world model updates its symbolic beliefs and infers coherent mission and route decisions when the spatial configuration differs from the training demonstrations.}
  \label{fig:ExpSurp}
\end{figure}

Fig.~\ref{fig:ExpSurp} evaluates the proposed framework under unseen target layouts using the real-flight validation setting. In these experiments, the spatial configurations of the target cities differ from those used during training. The target positions are provided as observations; therefore, the task is not to predict the physical coordinates themselves. Rather, the objective is to infer a coherent mission division, route order, and local motion behavior for a previously unseen layout.

Fig.~\ref{fig:Exp1Surp} shows the result for the first unseen-layout experiment, where the red marker indicates an unexpected change in the environment. The learned world model updates the symbolic belief and produces a consistent mission and route interpretation for the new configuration. Similar behavior is observed in Figs.~\ref{fig:Exp2Surp} and \ref{fig:Exp3Surp}, where the framework adapts to different target arrangements.

These results show that the proposed world model is not restricted to memorized city configurations. Instead, it uses the learned probabilistic structure to infer expert-consistent decisions for new spatial layouts.

\subsection{Local Path Prediction from the Learned World Model}
\label{subsec:local_path_prediction}

\begin{figure}[ht!]
  \centering
  \begin{subfigure}[t]{0.48\linewidth}
    \centering
    \includegraphics[width=\linewidth]{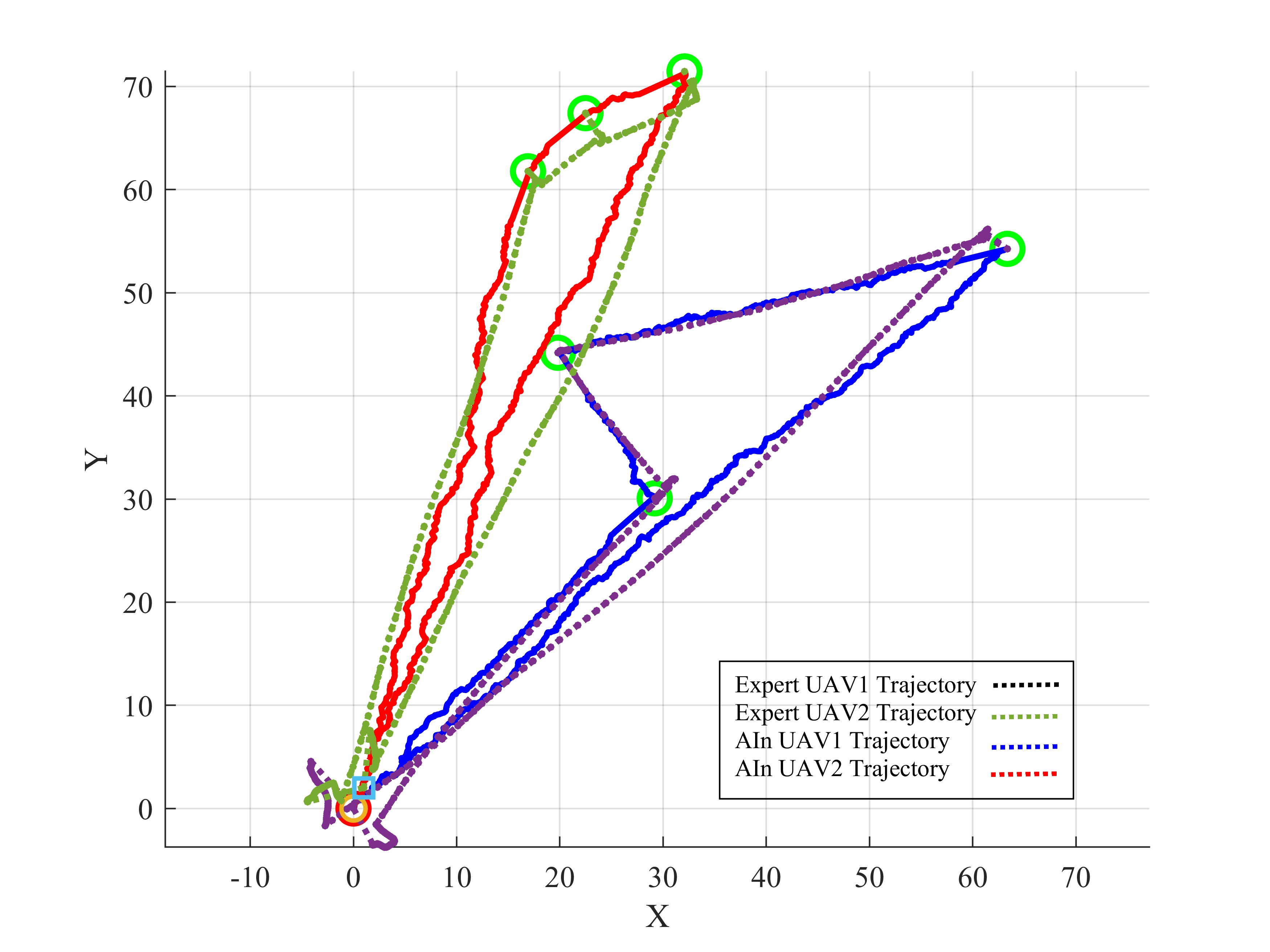}
    \subcaption{}
    \label{fig:Exp1Com}
  \end{subfigure}
  \hfill
  \begin{subfigure}[t]{0.48\linewidth}
    \centering
\includegraphics[width=\linewidth]{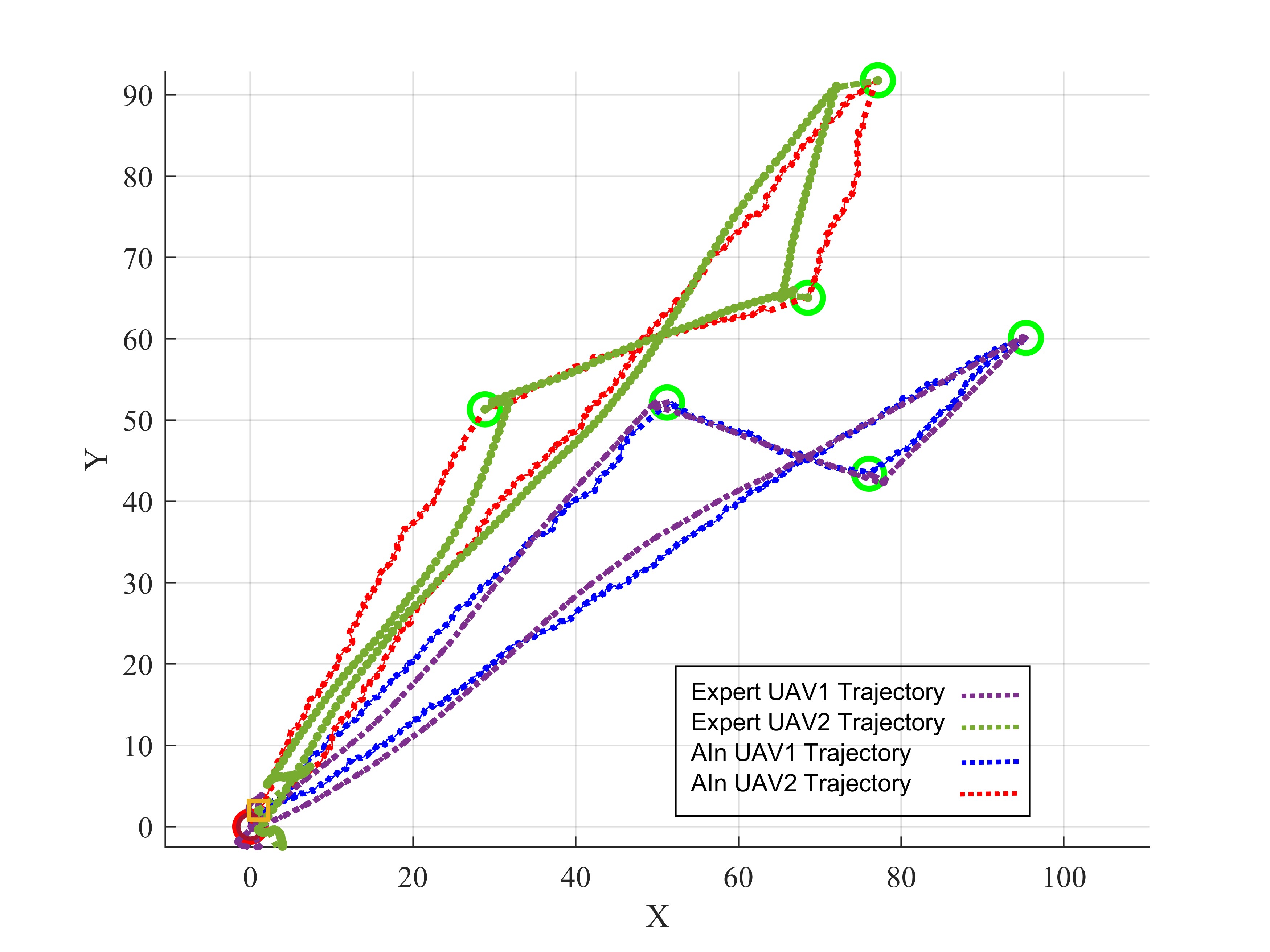}
    \subcaption{}
    \label{fig:Exp2Com}
  \end{subfigure}
  \begin{subfigure}[t]{0.48\linewidth}
    \centering
    \includegraphics[width=\linewidth]{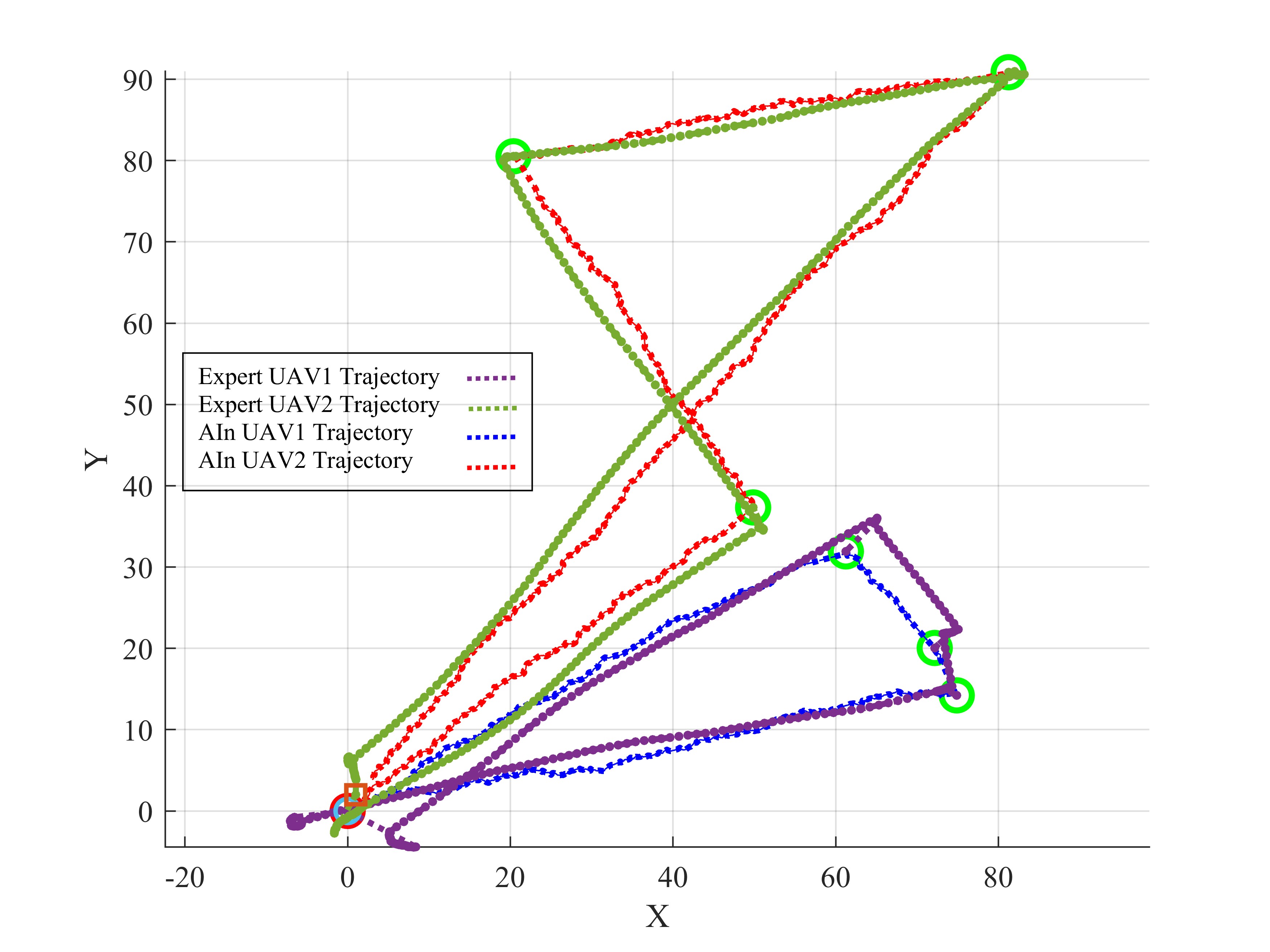}
    \subcaption{}
    \label{fig:Exp3Com}
  \end{subfigure}
  \caption{Comparison between human-expert real-flight trajectories and trajectories predicted by the proposed world model.
The predicted paths preserve the dominant local route structure while tolerating human-pilot variability and observation noise.}
  \label{fig:ExpCom}
\end{figure}

Fig.~\ref{fig:ExpCom} compares human-expert flight paths with trajectories predicted by the proposed active-inference-based world model for three validation experiments. In these experiments, the high-level task division and route ordering have already been inferred for the unseen layouts shown in Fig.~\ref{fig:ExpSurp}. The focus of Fig.~\ref{fig:ExpCom} is therefore the local path-prediction capability of the motion-level world model.

In Fig.~\ref{fig:Exp1Com}, the predicted trajectory follows the main structure of the human-operated trajectory, indicating that the learned motion words capture the dominant local motion behavior. In Fig.~\ref{fig:Exp2Com}, the model again produces a path consistent with the observed expert trajectory while adapting its prediction through belief updating. In Fig.~\ref{fig:Exp3Com}, the geometry is more complex, but the predicted path remains stable and preserves the main route structure.

The purpose of this comparison is not to reproduce every small fluctuation of the human pilot trajectory. Human-operated flights include local deviations, non-smooth motion, and pilot-specific maneuvers. Instead, the objective is to verify whether the proposed model predicts coherent local paths that remain consistent with the inferred mission and route decisions. The results show that the motion-level world model captures the dominant structure of local flight behavior while tolerating observation noise and trajectory variability.

%

\subsection{Success-Rate Analysis in Online Testing}
\label{subsec:success_rate}

\begin{figure}[ht]
\centering
\includegraphics[width=0.40\textwidth]{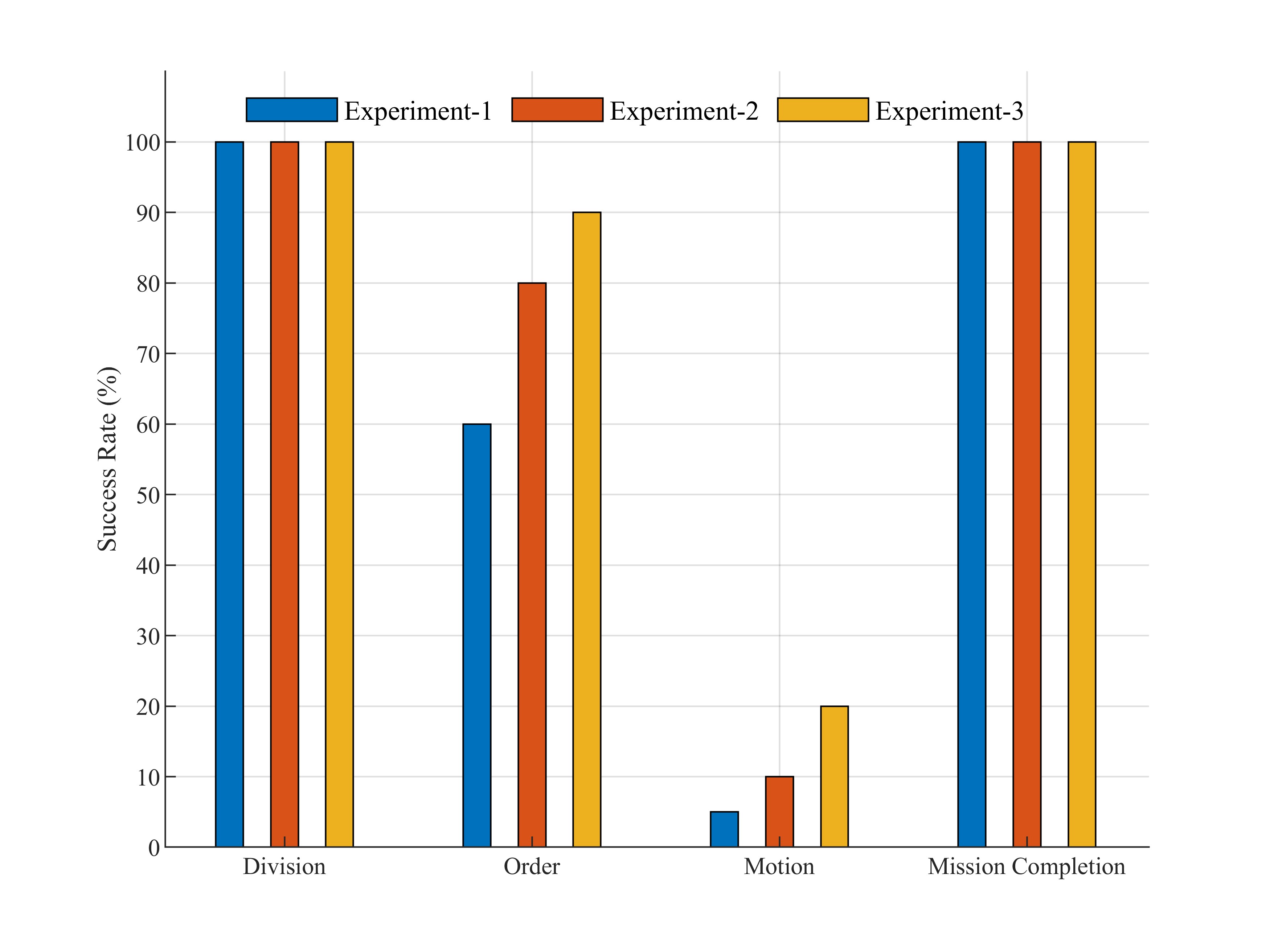} 
\caption{Success rates at different decision levels during online testing.
Mission division and mission completion remain successful across all experiments, while route ordering and motion prediction reflect the increasing difficulty of lower-level inference under real-flight variability.}
\label{fig:ExpSucc} 
\end{figure}

Fig.~\ref{fig:ExpSucc} reports the success rate of the proposed world model at different decision levels under the three real-flight validation experiments. The evaluated levels are mission division, route ordering, motion prediction, and overall mission completion.

The mission-division level achieves 100\% success in all experiments. This indicates that the high-level symbolic belief is robust under unseen target layouts and that the framework can consistently assign cities to UAVs in a way that preserves the global mission objective.

The ordering-level success rate improves across the three experiments. This behavior reflects the effect of online belief updating: as observations are incorporated, the posterior route belief becomes more consistent with the observed structure. This should be interpreted as online posterior correction rather than parameter retraining, since the learned dictionaries and transition matrices remain fixed during testing.

The motion-level success rate is lower than the mission and route levels. This is expected because motion-level prediction is evaluated against noisy human-operated trajectories, where local deviations, non-smooth maneuvers, and pilot-specific behavior increase the mismatch between predicted and observed motion labels. Nevertheless, the motion-level success improves across the experiments, indicating that the belief-correction mechanism can reduce local prediction mismatch.

Finally, the mission-completion success rate remains 100\% in all experiments. This is an important result: although local motion prediction is more sensitive to noise, the overall hierarchical framework preserves the global mission objective. The swarm can still complete the task successfully because high-level mission and route decisions remain coherent even when local motion observations are imperfect.

Overall, Fig.~\ref{fig:ExpSucc} confirms that the proposed framework provides robust mission-oriented behavior under unseen layouts and real-flight trajectory variability. The results support the central claim that hierarchical probabilistic inference can maintain high-level task consistency while correcting low-level motion predictions online.

\subsection{Discussion}
\label{subsec:results_discussion}

The results demonstrate that the proposed framework successfully combines expert demonstrations, probabilistic symbolic modeling, and online belief updating for multi-UAV trajectory design. The learned transition matrices show that expert behavior can be compressed into a hierarchical world model. The abnormality plots verify that online decisions follow the proposed KL-based selection mechanism. The trajectory execution and replanning results show that the selected symbolic actions can be translated into feasible UAV trajectories and updated when new targets appear.

The comparison with modified Q-learning indicates that the proposed method benefits from the expert-derived hierarchical prior, producing smoother and more stable behavior. The comparison with GA--RF should be interpreted differently: GA--RF is the offline expert used to generate demonstrations, while the proposed framework is an online inference mechanism that reuses the learned expert structure. Therefore, the main advantage of the proposed method is not replacing the expert optimizer, but enabling expert-consistent online adaptation without repeated global optimization.

The real-flight validation further shows that the learned world model can handle noisy and non-smooth trajectory observations. Belief updating reduces symbolic prediction errors, and high-level mission completion remains successful even when motion-level prediction is affected by human-pilot variability. These results support the applicability of the proposed framework to adaptive UAV swarm planning, while also indicating that future work should further validate the method in outdoor environments, with larger heterogeneous swarms, communication constraints, and stronger aerodynamic disturbances.

\section{Conclusion}

This paper presented an expert-guided active-inference-inspired framework for adaptive UAV swarm trajectory planning. The proposed approach learns a hierarchical probabilistic world model from GA--RF expert demonstrations and uses it online to support mission allocation, route ordering, and motion-level adaptation. By representing expert behavior through Mission, Route, and Motion dictionaries, and by learning the probabilistic transitions among these levels, the framework converts multi-UAV planning from repeated global optimization into structured online inference.

Online decision-making was formulated through KL-divergence-based abnormality minimization. Candidate mission, route, and motion actions were evaluated by comparing their posterior symbolic beliefs with expert-derived reference distributions. This enabled the swarm to adapt to new target configurations and environmental changes without rerunning the offline GA--RF optimizer. At the motion level, EKF- and PF-assisted state estimation provided continuous-state correction and improved collision-aware trajectory execution under uncertainty.

Simulation results showed that the learned transition matrices capture meaningful expert structure, that online candidate selection follows the proposed abnormality-minimization rule, and that the resulting trajectories remain coherent across mission, route, and motion levels. Compared with modified Q-learning, the proposed framework produced smoother and more stable behavior while preserving expert-consistent mission efficiency. Real-flight trajectory validation further demonstrated that the learned world model can correct symbolic prediction errors under noisy, non-smooth, human-operated UAV trajectories, while maintaining high mission-completion success.
The present results suggest that expert-derived probabilistic world models can provide an interpretable and computationally efficient basis for adaptive UAV swarm autonomy. 
%

\bibliographystyle{IEEEtran}
\bibliography{references}

\begin{thebibliography}{10}
\providecommand{\url}[1]{#1}
\csname url@samestyle\endcsname
\providecommand{\newblock}{\relax}
\providecommand{\bibinfo}[2]{#2}
\providecommand{\BIBentrySTDinterwordspacing}{\spaceskip=0pt\relax}
\providecommand{\BIBentryALTinterwordstretchfactor}{4}
\providecommand{\BIBentryALTinterwordspacing}{\spaceskip=\fontdimen2\font plus
\BIBentryALTinterwordstretchfactor\fontdimen3\font minus \fontdimen4\font\relax}
\providecommand{\BIBforeignlanguage}[2]{{%
\expandafter\ifx\csname l@#1\endcsname\relax
\typeout{** WARNING: IEEEtran.bst: No hyphenation pattern has been}%
\typeout{** loaded for the language `#1'. Using the pattern for}%
\typeout{** the default language instead.}%
\else
\language=\csname l@#1\endcsname
\fi
#2}}
\providecommand{\BIBdecl}{\relax}
\BIBdecl

\bibitem{11465094}
K.~Arshid, A.~Krayani, L.~Marcenaro, D.~M. Gomez, and C.~Regazzoni, ``{Active Inference-Driven World Modeling for Adaptive UAV Swarm Trajectory Design},'' in \emph{ICASSP 2026 - 2026 IEEE International Conference on Acoustics, Speech and Signal Processing (ICASSP)}, May 2026, pp. 22\,147--22\,151.

\bibitem{Ahmad2025}
F.~Ahmad, M.~Y. Mirza, I.~Hussain, and K.~Arshid, ``{A Multi‐Ray Channel Modelling Approach to Enhance UAV Communications in Networked Airspace},'' \emph{Inventions}, vol.~10, no.~4, p.~51, July 2025.

\bibitem{Arshid2025UAVSwarm}
K.~Arshid, A.~Krayani, L.~Marcenaro, D.~M. Gomez, and C.~Regazzoni, ``Toward autonomous uav swarm navigation: A review of trajectory design paradigms,'' \emph{Sensors}, vol.~25, no.~18, p. 5877, Sep. 2025.

\bibitem{liu2021car}
J.~Liu, Y.~Wang, P.-Q. Huang, and S.~Jiang, ``Car: A cutting and repulsion-based evolutionary framework for mixed-integer programming problems,'' \emph{IEEE Transactions on Cybernetics}, vol.~52, no.~12, pp. 13\,129--13\,141, 2021.

\bibitem{chai2025trajectory}
Y.~Chai, Z.~Zhang, H.~Yu, J.~Han, Y.~Fang, and X.~Liang, ``A trajectory planning scheme for collaborative aerial transportation systems by graph-based searching and cable tension optimization,'' \emph{IEEE/ASME Transactions on Mechatronics}, 2025.

\bibitem{wang2023improved}
F.~Wang, G.~Xu, and M.~Wang, ``An improved genetic algorithm for constrained optimization problems,'' \emph{IEEE Access}, vol.~11, pp. 10\,032--10\,044, 2023.

\bibitem{gad2022particle}
A.~G. Gad, ``Particle swarm optimization algorithm and its applications: a systematic review,'' \emph{Archives of computational methods in engineering}, vol.~29, no.~5, pp. 2531--2561, 2022.

\bibitem{manullang2023optimum}
M.~J.~C. Manullang, K.~Priandana, and M.~K.~D. Hardhienata, ``Optimum trajectory of multi-uav for fertilization of paddy fields using ant colony optimization (aco) and 2-opt algorithms,'' in \emph{AIP conference proceedings}, vol. 2482, no.~1.\hskip 1em plus 0.5em minus 0.4em\relax AIP Publishing, 2023.

\bibitem{shi2024deep}
B.~Shi, Z.~Chen, and Z.~Xu, ``A deep reinforcement learning based approach for optimizing trajectory and frequency in energy constrained multi-uav assisted mec system,'' \emph{IEEE Transactions on Network and Service Management}, 2024.

\bibitem{li2023computation}
X.~Li, Y.~Qin, J.~Huo, and W.~Huangfu, ``Computation offloading and trajectory planning of multi-uav-enabled mec: A knowledge-assisted multiagent reinforcement learning approach,'' \emph{IEEE Transactions on Vehicular Technology}, vol.~73, no.~5, pp. 7077--7088, 2023.

\bibitem{pezzulo2024active}
G.~Pezzulo, T.~Parr, and K.~Friston, ``Active inference as a theory of sentient behavior,'' \emph{Biological Psychology}, vol. 186, p. 108741, 2024.

\bibitem{nozari2022incremental}
S.~Nozari, A.~Krayani, L.~Marcenaro, D.~Martin, and C.~Regazzoni, ``Incremental learning through probabilistic behavior prediction,'' in \emph{2022 30th European Signal Processing Conference (EUSIPCO)}.\hskip 1em plus 0.5em minus 0.4em\relax IEEE, 2022, pp. 1502--1506.

\bibitem{Arshid2025}
K.~Arshid, A.~Krayani, L.~Marcenaro, D.~M. Gomez, and C.~Regazzoni, ``Uav swarm trajectory design for wireless networks using genetic algorithm‐driven repulsion forces,'' \emph{IEEE Access}, Sep. 2025.

\bibitem{Ergezer2013TAES}
H.~Ergezer and K.~Leblebicioglu, ``Path planning for uavs for maximum information collection,'' \emph{IEEE Transactions on Aerospace and Electronic Systems}, vol.~49, no.~1, pp. 502--520, Jan 2013.

\bibitem{Li2021TAES}
B.~Li, J.~Zhang, L.~Dai, K.~L. Teo, and S.~Wang, ``A hybrid offline optimization method for reconfiguration of multi-uav formations,'' \emph{IEEE Transactions on Aerospace and Electronic Systems}, vol.~57, no.~1, pp. 506--520, Feb 2021.

\bibitem{Chen2018TAES}
Y.~Chen, D.~Yang, and J.~Yu, ``Multi-uav task assignment with parameter and time-sensitive uncertainties using modified two-part wolf pack search algorithm,'' \emph{IEEE Transactions on Aerospace and Electronic Systems}, vol.~54, no.~6, pp. 2853--2872, Dec 2018.

\bibitem{Zhang2022TAES}
J.~Zhang, Y.~Cui, and J.~Ren, ``Dynamic mission planning algorithm for uav formation in battlefield environment,'' \emph{IEEE Transactions on Aerospace and Electronic Systems}, vol.~59, no.~4, pp. 3750--3765, Aug 2023.

\bibitem{Wu2025SciRep}
W.~Wu, L.~Zhang, J.~Le, and Z.~Lu, ``Integrated method for multi-uav task assignment and trajectory planning with deadlock based on three-dimensional dubins path,'' \emph{Scientific Reports}, vol.~15, pp. 1--13, 2025.

\bibitem{Roberge2018TAES}
V.~Roberge, M.~Tarbouchi, and G.~Labonté, ``Fast genetic algorithm path planner for fixed-wing military uav using gpu,'' \emph{IEEE Transactions on Aerospace and Electronic Systems}, vol.~54, no.~5, pp. 2105--2117, Oct 2018.

\bibitem{Ekechi2025DronesMARL}
C.~C. Ekechi, T.~Elfouly, A.~Alouani, and T.~Khattab, ``A survey on uav control with multi-agent reinforcement learning,'' \emph{Drones}, vol.~9, no.~7, p. 484, 2025.

\bibitem{Chen2025CSURRLUAV}
H.~Chen, Y.~Lin, M.~Fu, L.~Yao, and M.~Sheng, ``A survey on reinforcement learning methods for uav systems,'' \emph{ACM Computing Surveys}, 2025.

\bibitem{10938048}
Z.~Peng, G.~Wu, B.~Luo, and L.~Wang, ``Multi-uav cooperative pursuit strategy with limited visual field in urban airspace: A multi-agent reinforcement learning approach,'' \emph{IEEE/CAA Journal of Automatica Sinica}, vol.~12, no.~7, pp. 1350--1367, July 2025.

\bibitem{Xu2025Drones}
Q.~Xu, F.~Wu, and Z.~Chen, ``Neural network-based path planning for fixed-wing uavs with constraints on terminal roll angle,'' \emph{Drones}, vol.~9, no.~5, p. 378, 2025.

\bibitem{Horn2012JGCD}
J.~F. Horn, E.~M. Schmidt, B.~R. Geiger, and M.~P. DeAngelo, ``Neural network-based trajectory optimization for unmanned aerial vehicles,'' \emph{Journal of Guidance, Control, and Dynamics}, vol.~35, no.~2, pp. 548--562, 2012.

\bibitem{Obite2023AIUAV}
F.~Obite, A.~Krayani, A.~S. Alam, L.~Marcenaro, A.~Nallanathan, and C.~Regazzoni, ``Intelligent resource allocation for uav-based cognitive noma networks: An active inference approach,'' in \emph{2023 IEEE Future Networks World Forum (FNWF)}, Nov 2023, pp. 1--7.

\bibitem{Krayani2023Sensors}
\BIBentryALTinterwordspacing
A.~Krayani, K.~Khan, L.~Marcenaro, M.~Marchese, and C.~Regazzoni, ``A goal-directed trajectory planning using active inference in uav-assisted wireless networks,'' \emph{Sensors}, vol.~23, no.~15, 2023. [Online]. Available: \url{https://www.mdpi.com/1424-8220/23/15/6873}
\BIBentrySTDinterwordspacing

\end{thebibliography}
\begin{IEEEbiography}[{\includegraphics[width=1in,height=1.40in,clip,keepaspectratio]{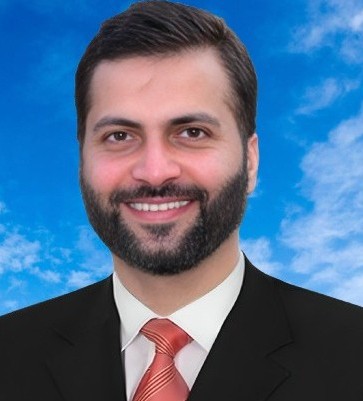}}]{Kaleem Arshid}  (Member, IEEE)  currently a Ph.D student in a joint doctoral program at Department of Electrical, Electronic, Telecommunications Engineering, and Naval Architecture (DITEN), University of Genova, Italy and Department of Systems Engineering and Automation, Carlos III University, Madrid, Spain. He did his MPhil in Computer Science from Preston University in Kohat, Pakistan, in 2017, and his Bachelor of Science in Information Technology from the University of Azad Jammu and 
Kashmir in Muzaffarabad, Pakistan, in 2013. From 2017 to 2019, he served as a Lecturer in the Department of Computer Science at the University of Sargodha, Pakistan. His research interests span across Cognitive Radio Networks, Active Inference, Gen.AI, Machine Learning, Artificial Intelligence and UAV-Assisted Wireless Networks. 
\end{IEEEbiography}
\begin{IEEEbiography}[{\includegraphics[width=1in,height=1.25in,clip,keepaspectratio]{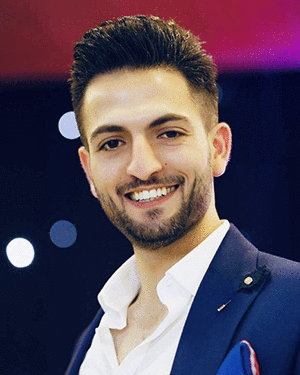}}]{Ali Krayani}(Member, IEEE) is an Assistant Professor within the Department of Electrical, Electronic, Telecommunications Engineering, and Naval Architecture (DITEN) at the University of Genoa, Italy. He earned his Bachelor’s degree in Telecommunication Engineering from the Politecnico di Torino, Italy, in 2014, followed by a Master’s degree in Telecommunication Engineering from the University of Florence, Italy, in 2017. In April 2022, he was awarded a joint Ph.D. degree from the University of Genoa, Italy, and Queen Mary University of London, London, United Kingdom. Dr. Krayani has worked as a Software Engineer in various companies and was a Postdoctoral Research Fellow at DITEN from 2021 to 2023. In 2023, he received the Best Paper Award at the IEEE Wireless Communications and Networking Conference (WCNC). He serves as a Guest Editor and reviewer for several academic journals. His current research interests include integrated sensing and communication, cognitive radios, AI-enabled radios, wireless communications (5G, 6G), UAV communications, physical layer security, IoT, semantic communications, UAV swarms, NOMA, federated learning, and artificial intelligence.
\end{IEEEbiography}
\begin{IEEEbiography}[{\includegraphics[width=1in,height=1.25in,clip,keepaspectratio]{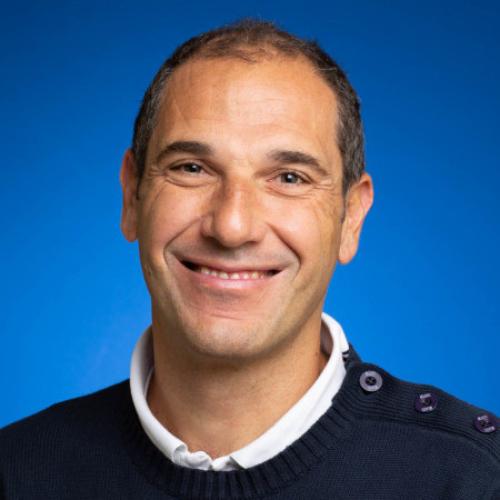}}]{Lucio Marcenaro} ((Senior Member, IEEE) received
 the degree in electronic engineering in 1999 and
 the Ph.D. degree in computer science and electronic
 engineering in 2003. He is currently an Associate
 Professor of telecommunications with the Polytech
nic School of Engineering, University of Genoa.
 He has over 20 years of experience in signal pro
cessing and image sequence analysis. He is the
 author of about 160 scientific papers related to signal
 processing for computer vision and cognitive radio.
 His main current research interests include video
 processing for event recognition, detection, and localization of objects in
 complex scenes, distributed heterogeneous sensors, and bio-inspired cognitive
 autonomous systems.
\end{IEEEbiography}
\begin{IEEEbiography}[{\includegraphics[width=1in,height=1.8in,clip,keepaspectratio]{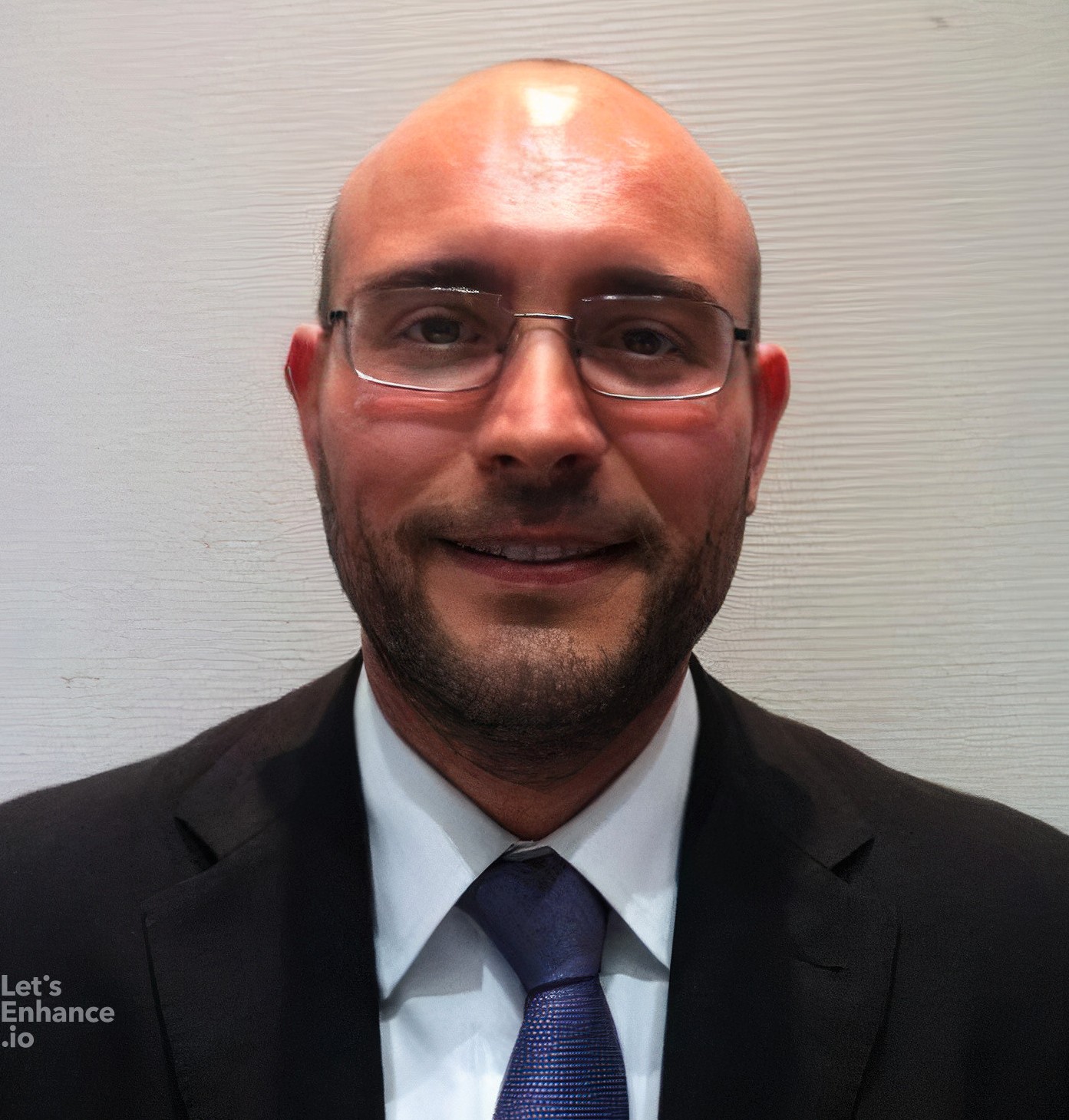}}]{David Martín Gómez} (Member, IEEE) received the degree in industrial physics (automation) from the National University of Distance Education (UNED)
in 2002 and the Ph.D. degree in computer science from the Spanish Council for Scientific Research (CSIC) and UNED, Spain, in 2008. He has been a member of the Intelligent Systems Laboratory
since 2011. He is currently a Professor at the Carlos III University of Madrid. In 2014, he was awarded with the VII
Barreiros Foundation Award to the best research in the automotive field. In 2015, the IEEE Society has awarded him as the Best Reviewer of the 18th IEEE International Conference on Intelligent
Transportation Systems.
\end{IEEEbiography}
\begin{IEEEbiography}[{\includegraphics[width=1in,height=1.25in,clip,keepaspectratio]{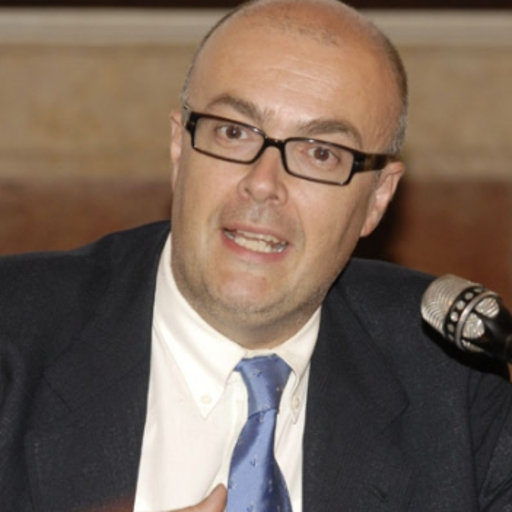}}]{Carlo Regazzoni}(Senior Member, IEEE) is cur
rently a Full Professor of cognitive telecommu
nications systems with the DITEN, University of
 Genoa, Italy. He has been responsible of several
 national and EU funded research projects. He is
 also the coordinator of international Ph.D. courses
 on interactive and cognitive environments involving
 several European universities. He served as a general
 chair for several conferences and an associate/guest
 editor for several international technical journals.
 He has served in many roles in governance bodies
 for IEEE SPS and he is serving as the Vice President-Conferences for the
 IEEE Signal Processing Society from 2015 to 2017.
\end{IEEEbiography}
\end{document}